# Action–outcome coupling: a pre-investment feasibility audit for offline reinforcement learning in supply chain operations

Tung Dang[a,*], The Hung Phung[b], Son Lam Nguyen[c] and Tu Nguyen[d,*]

[a]Graduate School of Science, The University of Tokyo, 7-3-1 Hongo Bunkyo-ku, Tokyo 113-0033, Japan
[b]Faculty of Human Resource Management, Trade Union University, 169 Tay Son, Kim Lien, Hanoi Vietnam
[c]Institute for Business Management Development, Academy Of Finance, 58 Le Van Hien Street, Dong Ngac Ward, Hanoi Vietnam
[d]Ministry of Industry and Trade, 23 Ngo Quyen, Hoan Kiem, Hanoi Vietnam

ABSTRACT

Prescriptive analytics initiatives in logistics, such as offline reinforcement learning and next-best-action systems, are often funded before anyone verifies that the available historical logs can support an improved decision policy. We formalise this pre-investment feasibility question via action–outcome coupling: a total-effect variance index measuring how much expected outcomes change with the logged action. Coupling provides an algorithm-independent upper bound on any policy's achievable value, expressed in monetary units per decision for both single-step and discounted sequential objectives. We also analyse the asymptotic false-licensing risk of an invalid workflow: fitting a value model and evaluating the model's greedy policy on the same data. The false-licence rate is driven mainly by the number of state cells over which selection is performed, rising from the nominal test level with one cell to near certainty with a few dozen, while overlap has a smaller effect. We then propose an audit that assigns logs to four tiers of decision support, separately for fixed and contextual policies, and returns an explicit stopping rationale. On a randomised e-commerce log, the audit endorses a fixed recommendation policy, certifying gains equal to 30% of the theoretical ceiling; the remaining gap reflects bound looseness under an 80-action design. Two public logistics logs fail at different checkpoints: one because the target is an accounting identity of the action, and another because, with 27 clusters, estimated coupling stays below the design's detectability threshold even though a conservative offline learner certifies an improvement in 29 of 30 seeds.

## 1. Introduction

A fulfilment organisation contemplating a next-best-action programme may possess demand forecasts and a pipeline plan, yet lack a principled method to determine whether its historical training log is informative enough to support the learning of any decision policy. Given the substantial organisational commitment, this question is consequential: reinforcement learning (RL) is increasingly applied in logistics and supply chains, including inventory replenishment [1, 2], dynamic pricing [3], multi-echelon control [4, 5], and demand fulfilment [6]. In commercial settings, offline RL is often the most relevant paradigm because it seeks to learn from existing transactional records rather than through costly or risky online experimentation [7, 8]. Nevertheless, a foundational pre-investment question is frequently left implicit: can the available log support a policy? This paper formalises that question and provides a computable procedure that a decision-maker can apply ex ante.

The literature's own empirical record motivates this focus. Reviews of deep RL for inventory control describe a persistent gap between methodological optimism and operational deployment: the design and tuning effort each algorithm demands may itself hamper application in practice [2], practical application so far has produced mixed results, and the demonstrations at industrial scale that do exist are single-firm deployments [9]. The principal technical obstacles are well documented: offline RL methods may extrapolate unreliably outside the action distributions represented in the log [10, 11], and off-policy evaluation (OPE) can fail in ways that are not apparent from point estimates alone [12, 13]. Current diagnostics primarily assess the feasibility of estimation (such as overlap, propensity model calibration, and effective sample size [14, 15, 16]), but do not directly address whether the policy-relevant estimand is substantively nonzero. Consequently, a dataset can satisfy these diagnostics and still be unusable for policy learning if outcomes exhibit negligible or no causal responsiveness to decisions. This pathology is not merely theoretical: in a widely used benchmark dataset [17], the service target is an accounting identity of the action in 97.55% of observations, a defect that cannot be remedied by increased sample size, improved overlap, or alternative modelling choices.

We formalise the latent quantity of interest as the action–outcome coupling, which quantifies, averaged over states, the extent to which the expected outcome varies as the decision (action) is perturbed under the logging distribution. This quantity corresponds to a total-effect sensitivity index [18] and therefore preserves interaction effects that would be obscured by one-factor analyses based solely on marginal action means. The definition yields three practically relevant implications that are computable ex ante, i.e.,

*Corresponding author
dangthanhtung91@g.ecc.u-tokyo.ac.jp (T. Dang); tunn@moit.gov.vn (T. Nguyen)
ORCID(s): 0000-0002-4974-9632 (T. Dang); 0009-0001-8682-5235 (T.H. Phung); 0009-0001-5302-5047 (S.L. Nguyen); 0009-0001-1781-2617 (T. Nguyen)

prior to constructing any downstream learning or evaluation pipeline. First, the attainable value of any policy is upper-bounded by the coupling present in the data, an algorithm-independent performance ceiling that we express in dollars per order (Theorem 1). The same argument, run through the performance difference lemma, bounds the sequential gap by the discounted coupling $\mathcal{D}_{\text{seq}}$ (Theorem 1(c)), and the continuation coupling $\mathcal{D}_{\text{cont}}$ that this decomposition isolates gives a testable condition for downstream responsiveness (Proposition 1). We do not exercise that condition here, because all three corpora record single-step decisions. Second, assessing whether a log contains sufficient coupling reduces to a power analysis with two distinct thresholds: the minimum coupling that is detectable, and, separately, the minimum coupling that is learnable by a specified learner (Proposition 3); we emphasise that the learnability frontier is a calibrated conjecture, not a theorem. Third, coupling should be assessed prior to overlap diagnostics. Under fold reuse (and only under reuse), near-random logging can drive standard off-policy estimators to a narrow interval centred on a spurious improvement; the rate depends on the deployed learner, and Theorem 2 gives its limit, separating overlap, which bounds the statistic's drift within a factor of $\sqrt{2}$ and the rate within roughly 2.3-fold, from the number of state cells selected over, which drives the rate towards one. This failure mode occurs in the regime usually regarded as the most benign. Cross-fitting eliminates the relevant bias [19], implying that the pathology is attributable to reuse; our contribution is to characterise its rate and to make the optimizer's curse [20] operational in a form that decision-makers can value.

This theory motivates an instrument intended to be run before engineering effort is committed to any operational pipeline. We present the instrument as a design-science artefact [21, 22] that produces a graded recommendation instead of a binary accept/reject decision. It routes a log to the class of decision support that the log can sustain (refuse, predict, shadow, or deploy) and attaches a reason code indicating the audit stage at which it stopped. Unlike a checklist, the instrument puts a price on a refusal. By inverting the power calculation, it returns an exploration budget: randomise $x\%$ of orders for $y$ weeks to achieve power sufficient to detect coupling valued at $\$z$ per order. The instrument therefore tells managers not only when current data are insufficient to justify a policy, but also what it would cost to make the data adequate.

We audited three public logs, and the instrument licensed one of them. On a randomised e-commerce log [15] the coupling clears detectability by a wide margin and a learner certifies a fixed rule, so the audit licenses fixed rules; the increment that personalisation would add is detectable but does not certify, so contextual rules are refused. The two logistics logs are refused, at different stages. The DataCo benchmark [17] fails the definitional-target gate and, on its margin reward, is underpowered at the clusters available. The Olist e-commerce fulfilment log [23] passes the estimability gates, and its conservative learner [11] appears to improve in 29/30 honest-split seeds against a calibrated permutation null, yet that certificate is inadmissible by the audit's own coherence rule, because at the available 27 clusters the coupling does not clear its detectability threshold. A pre-specified saturation ladder, read on the absolute coupling, does not decline on either window, so the refusal makes no claim about confounding [24, 25].

The paper makes four contributions: (i) a formal definition of action–outcome coupling and a monetary upper bound on policy value, for single-step objectives (Theorem 1) and, through a performance-difference argument, for discounted sequential ones (Theorem 1(c)), which also yields a testable condition for downstream responsiveness, not exercised here because all three corpora record single-step decisions; (ii) a coupling-before-overlap sequence with the limiting rate of false-positive bias under fold reuse, verified against the finite procedure for three fitted learners and separating the role of overlap from that of the number of state cells selected over (Theorem 2); (iii) a four-level routing tool with reason codes, evaluated end-to-end against the procedures it replaces and decomposed into the part each change contributes: of the 18.7% false-licence rate that fold splitting alone leaves on contextual policies, 18.6 points are removed by asking what personalisation adds instead of whether a contextual policy beats the logging policy, and the audit's own machinery removes the remainder and answers the logs it cannot decide with an indeterminate verdict and the pilot that would settle it, at the cost of abstaining on 45% of contextual logs; and (iv) a validation on three public logs: a randomised log licensed at Tier 3 for non-contextual rules and refused at certification for contextual ones (1 (L)), and two logistics logs rejected at different stages, with the refusal recurring on a later window. We emphasise that the learnability frontier remains a calibrated conjecture, and none of the three logs tests it: the two logistics logs are refused before certification binds, and the randomised log certifies far above the band.

## 2. Related work

*Policy learning from observational data.* Regret bounds for policies learned from logged data describe how quickly a learned rule approaches an optimal policy $\pi^\star$ [26, 27, 28]. They do not, however, address whether the value gap $V(\pi^\star) - V(\pi_b)$ is bounded away from zero. If $Q(s, \cdot)$ is action-invariant on the support of the reference design, $\pi^\star$ is no better than any other policy, so the regret bound is correct but vacuous. We test the premise those bounds assume; the ceiling in Theorem 1 bounds the numerator of the value gap that the regret literature takes as given.

*Off-policy evaluation and feasibility diagnostics.* OPE work studies when value estimates are reliable, proposing doubly robust and data-efficient estimators, self-normalised weights, and diagnostics based on overlap and effective sample size [13, 14, 19, 29, 30]. These diagnostics assess estimator quality, not whether the estimand is non-degenerate. Related work on deficient support concerns $\pi_b$ when some actions have zero logging probability [16, 31, 32]. Our coupling is a property of $Q$: support may be full and overlap and calibration perfect while outcomes do not respond to actions. Our condition therefore precedes OPE diagnostics; the Olist audit shows that all standard diagnostics can pass while the log should still be rejected.

*Testing causal effects and heterogeneity.* Closest to our coupling test C3 are propensity-score and potential-outcomes methods [33, 34], heterogeneity prediction and sensitivity analysis [19, 25], and uplift/predict-then-optimize models [35, 36]. These aim to infer effects. Here, coupling supports a power calculation and routing rule, is benchmarked against an economic break-even, and is combined with a saturation ladder to separate real effects from the logging policy's private information. Our index $\mathcal{D}$ is a total-effect (Sobol-type) sensitivity index, not a main effect [18], because fulfilment coupling is interactive: effects can offset across lanes, yielding zero main effect but non-zero $\mathcal{D}$, which one-way tests miss.

*Post-selection inference and the optimizer's curse.* Selecting and evaluating on the same data inflates false positives; cross-fitting mitigates this [20, 37, 38, 39, 40, 41]. We characterise the limiting rate and separate the two parameters that drive it: with fold reuse, the false-positive rate rises with effective overlap and peaks under uniform logging, but the drift is bounded (bias factor $\leq \sqrt{2}$, rate $\approx 2.3\times$). In contrast, selecting over many state cells drives the rate from $\alpha$ to near certainty. A learner making one global comparison stays at $\alpha$ for any overlap (Theorem 2(i)). This is a reuse effect: cross-fitting removes the bias, so the coupling test is run first, cheaply, on disjoint folds.

*Design-science evaluation.* We treat the instrument as a design-science artefact and evaluate it accordingly [21, 22]. It reports false-licence and false-kill rates as well as successes, and the analysis is pre-specified except where Section 6 states otherwise. Prior work thus asks whether estimates, learning, or inference remain valid given an underlying effect. The earlier question, whether logged outcomes respond to logged decisions enough to justify building, is typically assumed. We test and quantify that question, and show that on real logistics data the answer can be no.

## 3. The feasibility conditions and the coupling object

### 3.1. Setting and notation

A log contains $n$ rows $(s_i, a_i, y_i)$. In fulfilment a row is an order line: the state $s$ is recorded before the decision, the action $a \in \{1, \dots, K\}$ is the mode committed to and is drawn from an unknown logging policy $\pi_b(a \mid s)$, and $y_i = Y_i(a_i)$ is the realised outcome. Rows are grouped into $G$ clusters (lanes, dispatch windows, customer regions) within which outcomes may be dependent.

Coupling is a property of the response surface, not of the way the incumbent happened to sample actions. All quantities are therefore defined relative to a reference design $\rho$, a distribution over the actions among which a candidate policy may choose:

$$Q(s,a) = \mathbb{E}[Y(a) \mid s], \qquad \bar{Q}_\rho(s) = \mathbb{E}_{a\sim\rho}[Q(s,a)], \quad (1)$$

with $V(\pi) = \mathbb{E}_s \mathbb{E}_{a\sim\pi}[Q(s,a)]$, $\Delta(\pi) = V(\pi) - V(\pi_b)$ and $\pi^\star = \arg\max_\pi V(\pi)$. A prescriptive programme [8, 26, 27, 28, 42] is an investment on the premise $\Delta(\pi^\star) > 0$. The audit asks whether a log supports that premise, and for which class of policies. We use a uniform $\rho$ by default. It maximises $\rho_{\min}$ and so minimises the ceiling constant $c_\rho$ of Theorem 1 (Remark 7). This is a reporting convention, not a recommended policy. A firm with a different feasible design reruns the audit under its own $\rho$, and a refusal under the uniform design does not carry over automatically.

*Notation.* The overlap floor $\epsilon_0$ requires $\pi_b(a \mid s) \geq \epsilon_0$ for every $a$ with $\rho(a) > 0$. The importance weight is $w = \rho(a)/\pi_b(a \mid s)$ and the Kish effective sample size is $n_{\text{eff}} = (\sum_i w_i)^2 / \sum_i w_i^2$ [43, 44]. The ceiling constant is $c_\rho = \sqrt{(1-\rho_{\min})/\rho_{\min}}$, equal to $\sqrt{K-1}$ for uniform $\rho$. The information count $n_{\text{info}}$ (§3.4) and the detectability threshold $\eta^2_{\min}$ (Proposition 3) complete the list. Three standing scope conditions (row clustering, the dependence of every verdict on the chosen reward, and the two-sided effect of binning) are collected in Remark 6.

*Policy classes.* We distinguish the class $\Pi$ of all state-dependent policies from the class $\Pi_{\text{const}}$ of non-contextual rules that play one action at every state. The distinction is not cosmetic. A log can support a single best action while being far too sparse to support personalisation, and the audit then licenses the first class and not the second (§5.1). Each verdict therefore names its policy class.

*The sequential case.* Both field corpora record single-step decisions, but the settings that motivate offline RL (inventory, multi-echelon control) are sequential, so the ceiling is also stated for discounted objectives. A log of trajectories is generated by a Markov decision process $(\mathcal{S}, \mathcal{A}, P, R, \gamma)$ with $\gamma \in [0,1)$ and initial distribution $\mu$. The reference design becomes a stationary policy $\rho$ with action-value function $Q^\rho(s,a) = \mathbb{E}[\sum_{t\geq 0} \gamma^t R_t \mid s_0 = s, a_0 = a, \rho]$ and value $V^\rho(s) = \mathbb{E}_{a\sim\rho} Q^\rho(s,a)$, and $d^\pi_\mu(s) = (1-\gamma)\sum_{t\geq 0} \gamma^t \Pr(s_t = s \mid \mu, \pi)$ is the normalised discounted occupancy of policy $\pi$. Setting $\gamma = 0$ recovers the single-step objects: $Q^\rho = R = Q$, $V^\rho = \bar{Q}_\rho$ and $d^\pi_\mu = \mu$ for every $\pi$.

### 3.2. Assumptions

Table 1 records which result needs which assumption. One point carries weight: Theorem 1 needs A3 and A4 and neither A1 nor A2. Without SUTVA the function $Q(s,a)$ is not defined; ignorability and overlap are needed only to estimate $\mathcal{D}$ from a log. Theorem 2 needs none of A1, A2 or A5 beyond finite outcome variance, because it is a statement about a decoupled log analysed with a known learner.

**Table 1**
Assumptions, the results that require them, and the consequence of failure.

| # | Assumption | Required for | If it fails | Detail |
|---|---|---|---|---|
| A1 | Ignorability $\{Y(a)\} \perp a \mid s$ | estimating $\mathcal{D}$; ladder | a positive verdict may be an artefact (the ladder screens this); a negative verdict may be masked | Appendix A.1 |
| A2 | Overlap under $\rho$ (floor $\epsilon_0 > 0$) | identification of $\mathcal{D}$; detectability | $\mathcal{D}$ not identified | Appendix A.2 |
| A3 | $\mathbb{E}[Y^2 \mid s, a] < \infty$ | Theorem 1 | ceiling not defined | — |
| A4 | SUTVA, no interference | Theorem 1 (defines $Q$) | ceiling is partial-equilibrium; direction of the error undetermined | Appendix A.3 |
| A5 | $\mathbb{E}[w^2 Y^4] < \infty$ | Prop. 3; interval estimation | central-limit approximations fail; use self-normalised or trimmed weights | Appendix A.4 |

## 3.3. Action–outcome coupling

The law of total variance under $\rho$ gives

$$\operatorname{Var}_\rho(Y) = \operatorname{Var}_s(\bar{Q}_\rho(s)) + \underbrace{\mathbb{E}_s \operatorname{Var}_{a\sim\rho}[Q(s,a)]}_{\mathcal{D}} + \mathbb{E}_{s,a\sim\rho}[\operatorname{Var}(Y \mid s,a)], \tag{2}$$

whose left side is the outcome variance the reference design would produce, not the observed $\operatorname{Var}(Y)$.

**Definition 1** (Action–outcome coupling)**.** *The coupling is $\mathcal{D} = \mathbb{E}_s \operatorname{Var}_{a\sim\rho}[Q(s,a)]$. Its normalised forms are $\eta_a^2 = \mathcal{D}/\operatorname{Var}_\rho(Y)$ and the partial coupling $\eta_{a|s}^2 = \mathcal{D}/(\mathcal{D} + \mathbb{E}_{s,a\sim\rho}[\operatorname{Var}(Y \mid s,a)])$; the two coincide exactly when $\operatorname{Var}_s(\bar{Q}_\rho(s)) = 0$.*

$\mathcal{D} = 0$ means that, at every state carrying mass, the expected outcome is the same whichever action in the support of $\rho$ is taken. $\mathcal{D}$ is a total-effect sensitivity index [18]: it averages the action-induced variability within each state, so it keeps interactions that marginal action means would cancel. To make this explicit, let $M(a) = \mathbb{E}_s[Q(s,a)]$, $\bar{M} = \mathbb{E}_{a\sim\rho} M(a)$ and $I(s,a) = Q(s,a) - \bar{Q}_\rho(s) - M(a) + \bar{M}$. Because $\mathbb{E}_s I(s,a) = 0$ for every $a$, the cross term vanishes and

$$\mathcal{D} = \underbrace{\operatorname{Var}_{a\sim\rho}\left(M(a)\right)}_{\mathcal{D}_{\text{main}}} + \underbrace{\mathbb{E}_s \mathbb{E}_{a\sim\rho}\left[I(s,a)^2\right]}_{\mathcal{D}_{\text{int}}}. \tag{3}$$

The two components answer different managerial questions. $\mathcal{D}_{\text{main}}$ bounds what any non-contextual rule can gain (Corollary 2); $\mathcal{D}$ bounds what any policy can gain. A one-way $F$-test on the $K$ action means has power only against $\mathcal{D}_{\text{main}}$: under a purely interactive alternative, in which a mode helps some lanes and hurts others, its power equals its size although $\mathcal{D} > 0$.

The test therefore targets the same object as the estimand. Two nested models are fixed before the target data are analysed:

$$\begin{aligned} \mathcal{M}_0 &: \mathbb{E}[Y \mid s] = f_0(s), \\ \mathcal{M}_1 &: \mathbb{E}[Y \mid s, a] = f_0(s) + \textstyle\sum_{k\geq 2} \alpha_k \mathbf{1}[a{=}k] \\ &\qquad + \sum_{k\geq 2} g_k(s)\mathbf{1}[a{=}k], \end{aligned} \tag{4}$$

with $df_1 = \dim(\mathcal{M}_1) - \dim(\mathcal{M}_0)$. Dropping the $g_k$ terms gives the main-effect contrast $\mathcal{M}_1^{\text{main}}$ with $df_1 = K - 1$, which tests $\mathcal{D}_{\text{main}}$ and is the relevant contrast for $\Pi_{\text{const}}$. Two rules apply to both contrasts. The action main effect is always retained, since omitting it removes the alternative a one-way test would detect. No covariate enters $\mathcal{M}_1$ without also entering $\mathcal{M}_0$; otherwise the extra degrees of freedom would measure additional state dependence, not coupling. The estimator $\hat{\omega}^2$ of the partial coupling is computed from the same contrast as the test (Appendix A.5), and the audit prints the covariate groups and their degrees of freedom.

## 3.4. Information accounting

The verdict of §5.1 compares a coupling interval with a detectability threshold. The comparison is meaningful only if both sides count information in the same way. A threshold computed as if rows were independent, set against an interval from a cluster bootstrap, compares two different sample sizes, and the error can run in either direction. We define the information count $n_{\text{info}}$ of the tested action block as the number of independent, equally weighted rows that would give its test statistic the same sampling variance, and obtain it by one of three routes. The route follows from the design and is fixed before the target data are analysed.

(i) *Randomised logs with known propensities.* No information count is needed. The threshold is calibrated on the randomisation distribution itself: actions are permuted within strata, a known coupling is injected, and the threshold is the smallest coupling the test detects with probability $1 - \beta$ (Algorithm 2). The randomisation distribution is fixed by the design, so this calibration remains valid whatever the dependence among outcomes [34].

(ii) *Observational logs with independent rows.* $n_{\text{info}} = n_{\text{eff}}$.

(iii) *Observational logs with clustered rows.* The information is deflated by the generalised design effect of the action block [45]:

$$\begin{aligned} n_{\text{info}} &= n_{\text{eff}}^{\text{clu}} = \frac{n_{\text{eff}}}{\bar{\delta}}, \\ \bar{\delta} &= \frac{1}{df_1} \operatorname{tr}\left(V_0^{-1} V_{\text{CR}}\right), \end{aligned} \tag{5}$$

where $V_0$ and $V_{\mathrm{CR}}$ are the model-based and the cluster-robust covariance matrices of the $df_1$ action coefficients of $\mathcal{M}_1$ under the importance weights.

Two remarks qualify route (iii). First, the familiar factor $1+(\bar{m}-1)\hat{\rho}_{\mathrm{ICC}}$ is the design effect of a mean. For a regressor that varies within clusters, such as the action, it overstates $\bar{\delta}$ by roughly the inverse of the within-cluster correlation of the action [46]. A larger design effect raises the threshold and makes a refusal easier, so the mean-based factor is not the conservative choice for a refusal. Second, with few clusters $\bar{\delta}$ is itself uncertain. The audit reports a leave-one-cluster-out jackknife interval $[\bar{\delta}_{\mathrm{lo}}, \bar{\delta}_{\mathrm{hi}}]$ and carries it into a threshold interval. A cluster-robust Wald test of $df_1$ restrictions also requires $df_1 \le G-1$; a specification that violates this rank condition is replaced by the largest admissible one, under a fallback rule fixed in advance.

### 3.5. Overlap: two statistics, two roles

Overlap enters the analysis twice. Identification of $\mathcal{D}$ requires every action with $\rho(a) > 0$ to have positive logging probability at every state with mass, a pointwise condition. Estimation precision depends on an average.

**Definition 2** (Effective overlap)**.** *For a candidate policy $\pi$, the effective overlap functional is*

$$H(\pi) := \mathbb{E}_s\, \mathbb{E}_{a\sim\pi}\big[\pi_b(a \mid s)^{-1}\big].$$

*For the uniform policy we write*

$$H = \mathbb{E}_s\Big[K^{-1}\sum_a \pi_b(a \mid s)^{-1}\Big], \qquad o_{\mathrm{eff}} := K/H,$$

*so that $K \le H \le \epsilon_0^{-1}$, with $H = K$ if and only if the logging policy is uniform.*

Two facts are used repeatedly. Under uniform $\rho$, $n_{\mathrm{eff}} \to n\, o_{\mathrm{eff}}$ (Appendix A.6), so the effective sample size of the coupling estimator and the effective overlap are one number. By Jensen's inequality, $o_{\mathrm{eff}} \le K^2/\sum_a p_a^{-1}$ with $p_a$ the marginal action shares, with equality only when $\pi_b$ does not depend on the state (Remark 5). The floor $\epsilon_0$ is neither necessary nor sufficient for a given value of $H$ (Appendix B.2), which is why precision is reported through $H$ and identification is screened pointwise.

### 3.6. The conditions

Table 2 lists the conditions. Three kinds of priority should be kept apart. Logically, C1 and C2 come first: if the target is defined by the action or the state has leaked, any measured $\mathcal{D}$ is an artefact. For identification, overlap and calibration must be checked before estimation, because $\mathcal{D}$ is recovered by reweighting with $\hat{\pi}_b$. For the decision, C3 binds, because it alone determines the ceiling. Testing coupling early is an efficiency rule: it can stop a programme before any off-policy evaluation pipeline is built. Validity comes from fold discipline, not from the order of the tests (§4.4). C6 applies only when the proposed objective is sequential (Proposition 1). Two qualifications attach to every verdict and are derived in Appendix A.7 and Appendix A.3: a negative verdict carries a robustness value, and the ceiling is a partial-equilibrium statement.

## 4. Theory

### 4.1. The prescriptive ceiling

Theorem 1 is a statement about the population response surface and needs only finite second moments. Ignorability and overlap do not enter it; they are needed to estimate $\mathcal{D}$ from a log, not to bound what a policy can gain.

**Theorem 1** (Prescriptive ceiling)**.** *Under Assumptions A3 and A4, let $c_\rho := \sqrt{(1-\rho_{\min})/\rho_{\min}}$. Then:*

(a) $V(\pi^\star) - V(\rho) = \mathbb{E}_s[\max_a Q(s,a) - \bar{Q}_\rho(s)] \le c_\rho\sqrt{\mathcal{D}}$, *and the constant is sharp; for uniform $\rho$ the bound is $\sqrt{(K-1)\mathcal{D}}$.*

(b) *Against any baseline policy, in particular the incumbent, $\Delta(\pi^\star) \le 2c_\rho\sqrt{\mathcal{D}}$.*

(c) Sequential form. *Let $\mathcal{D}_{\mathrm{seq}} := \mathbb{E}_{s\sim d_\mu^\rho} \mathrm{Var}_{a\sim\rho}\, Q^\rho(s,a)$ and $C := \max_s d_\mu^{\pi^\star}(s)/d_\mu^\rho(s)$. For every $\gamma \in [0,1)$,*

$$V^{\pi^\star}(\mu) - V^\rho(\mu) \le \frac{c_\rho}{1-\gamma}\sqrt{\mathbb{E}_{s\sim d_\mu^{\pi^\star}} \mathrm{Var}_{a\sim\rho}\, Q^\rho(s,a)} \le \frac{c_\rho\sqrt{C}}{1-\gamma}\sqrt{\mathcal{D}_{\mathrm{seq}}},$$

*and at $\gamma = 0$ both bounds reduce to (a).*

*Proof.* At each state the $\rho$-weighted Laguerre–Samuelson inequality bounds $\max_a Q(s,a) - \bar{Q}_\rho(s)$ by $c_\rho$ times the $\rho$-standard deviation of $Q(s,\cdot)$; Jensen's inequality then moves the expectation inside the square root. Part (b) applies the same step to $-Q$. Part (c) runs the step inside the performance difference lemma [47]. Details are in Appendix B.1. □

*Corollary* 1 (Coupling and action-invariance). Under Assumption A4, $\mathcal{D} = 0$ if and only if $Q(s,\cdot)$ is constant on $\mathrm{supp}(\rho)$ for almost every $s$. In that case $\Delta(\pi) = 0$ for every policy with $\mathrm{supp}(\pi) \subseteq \mathrm{supp}(\rho)$.

*Corollary* 2 (Ceiling by policy class). For every non-contextual rule $\pi_a \equiv a$, $V(\pi_a) - V(\rho) = M(a) - \bar{M} \le c_\rho\sqrt{\mathcal{D}_{\mathrm{main}}}$, with $\mathcal{D}_{\mathrm{main}} \le \mathcal{D}$ from (3). The ceiling for $\Pi_{\mathrm{const}}$ is therefore computed from the main-effect contrast, and the ceiling for $\Pi$ from the full contrast.

*Remark* 1 (How the ceiling may be used). Theorem 1 is an upper bound and can only be used in one direction. A ceiling below the cost of building and running a programme is a sufficient reason to stop; a ceiling above that cost licenses nothing. The bound is attained only when a single action carries the whole advantage at every state, and it loosens as $K$ grows ($c_\rho = \sqrt{K-1}$ for uniform $\rho$). The audit therefore reports, wherever a policy can be evaluated, the tightness ratio of an achieved value to the ceiling, and when the log allows it a direct cross-fitted estimate of $\mathbb{E}_s[\max_a Q -$

**Table 2**
Feasibility conditions. Stage numbers refer to §5.1.

| Code | Condition | Type | Statistic | Threshold | Stage |
|---|---|---|---|---|---|
| C1 | Non-definitional target | domain + screen | out-of-sample $R^2$ of $y$ on $(s, a)$; functional recovery | flag as $R^2 \to 1$; human review | gate |
| C2 | No post-decision leakage | domain + screen | timestamp rule + training-time assertion | binary | gate |
| C3 | Action–outcome coupling | statistical | $\hat{\omega}^2$ for the pre-specified contrast(s), with interval and threshold on one information footing (§3.4) | three-branch rule (12) | 1 |
| C4a | Identification (local overlap) | statistical | state mass with $\hat{\pi}_b(a \mid s) < \delta_{\text{trim}}$ | $\delta_{\text{trim}}$ pre-specified | 0 |
| C4b | Precision (effective overlap) | statistical | $\hat{H}$, $o_{\text{eff}}$, $n_{\text{info}}$ | reported | 0 / 2 |
| C5a | Propensity calibration | statistical | ECE, calibration slope | ECE $< 0.05$ | 0 |
| C5b | Behaviour informativeness | statistical | held-out log-loss against marginal shares | $0.01$ nats/row | 0 |
| C6 | Downstream responsiveness (sequential objectives only) | statistical | $\hat{\mathcal{D}}_{\text{cont}}$ by fitted $Q$-evaluation of $\rho$; $\hat{\chi}_{\text{seq}}$ reported | lower limit of $\hat{\mathcal{D}}_{\text{cont}}$ above its calibrated threshold | 1 |

$\bar{Q}_\rho$] with an upper confidence limit valid under non-unique optima [48].

The constant matters. The classical factor $\sqrt{K-1}$ belongs to an unweighted mean and does not carry over to a weighted design: for $K = 2$ and $\rho = (0.1, 0.9)$ the sharp constant is $c_\rho = 3$, so using $\sqrt{K-1} = 1$ would understate the attainable improvement threefold (Remark 7).

*Where the sequential coupling comes from.* In part (c), $\mathcal{D}_{\text{seq}}$ is the dispersion statistic of Definition 1 applied to the reference policy's own action-value function, so it is estimable by fitted $Q$-evaluation of $\rho$ [49] and the coupling test extends to sequential objectives without a new estimator. The concentrability coefficient $C$ [50] is where overlap re-enters, now as a bound on the ratio of occupancies rather than on the per-state action distribution. The next result separates the immediate and the downstream channels.

**Proposition 1** (Immediate and downstream coupling)**.** *Write $PV^\rho(s,a) = \mathbb{E}[V^\rho(s') \mid s, a]$ and define, under the same state distribution $d_\mu^\rho$,*

$$\begin{aligned}\mathcal{D}_1 &:= \mathbb{E}\,\mathrm{Var}_{a\sim\rho}\, R(s,a),\\ \mathcal{D}_{\text{cont}} &:= \mathbb{E}\,\mathrm{Var}_{a\sim\rho}\left[Q^\rho(s,a) - R(s,a)\right] \\ &= \gamma^2\,\mathbb{E}\,\mathrm{Var}_{a\sim\rho}\, PV^\rho(s,a).\end{aligned} \tag{6}$$

*Then*

$$\left|\sqrt{\mathcal{D}_1} - \sqrt{\mathcal{D}_{\text{cont}}}\right| \le \sqrt{\mathcal{D}_{\text{seq}}} \le \sqrt{\mathcal{D}_1} + \sqrt{\mathcal{D}_{\text{cont}}},$$
$$\mathcal{D}_{\text{seq}} - \mathcal{D}_1 = \mathcal{D}_{\text{cont}} + 2\gamma\,\mathbb{E}\,\mathrm{Cov}_{a\sim\rho}\left[R(s,a), PV^\rho(s,a)\right],$$

*and $\mathcal{D}_{\text{cont}} = 0$ if and only if, at $d_\mu^\rho$-almost every state, the action does not change the expected value of $\rho$ from the next state onward.*

Proposition 1 fixes what condition C6 should test. Downstream responsiveness is $\mathcal{D}_{\text{cont}} > 0$, and C6 is decided on $\hat{\mathcal{D}}_{\text{cont}}$. The comparison of the sequential with the one-step coupling answers a different question, whether a single-step audit misstates the coupling, and is reported as the diagnostic

$$\chi_{\text{seq}} := \frac{\mathcal{D}_{\text{seq}} - \mathcal{D}_1}{\mathcal{D}_{\text{seq}}} \in (-\infty, 1]. \tag{7}$$

Its sign is set by the covariance term as much as by $\mathcal{D}_{\text{cont}}$. When an order costs money now and pays back later, the immediate and downstream effects offset, $\chi_{\text{seq}} < 0$, and the one-step audit overstates the coupling that a sequential policy can exploit, even though the action changes the future. In 3,000 random MDPs (2–8 states, 2–4 actions, $\gamma \in \{0.5, 0.8, 0.9, 0.95\}$, Dirichlet(0.5) transitions) this occurred in 1,164 cases (39%), and both inequalities of the proposition held in every one. A test of C6 based on the sign of $\chi_{\text{seq}}$ would have missed downstream responsiveness in all of them.

## 4.2. Degeneracy under decoupling

**Proposition 2** (Confident degeneracy)**.** *Assume $\mathcal{D} = 0$, so $Q(s,a) = g(s)$, and let $\hat{Q}(s,a) = g(s) + \epsilon(s,a)$ with $\epsilon(s,a) \sim \mathcal{N}(0, \sigma_\epsilon^2)$ independent across actions. The greedy policy selects the action with the largest fitting error, and the direct-method estimate of its improvement satisfies $\mathbb{E}[\hat{\Delta}_{\text{DM}}] = \sigma_\epsilon m_K > 0$, where $m_K$ is the expected maximum of $K$ independent standard normal variables, while the true improvement is zero.*

Proposition 2 is the optimizer's curse [20] and the maximisation bias of value-based RL [39, 40] for $K$ Gaussian errors; we state it because the audit uses its constant. The exact $m_K$, not its asymptotic form $\sqrt{2\log K}$, is the relevant quantity: the approximation overstates the spurious gain by a factor of 2.09 at $K = 2$. Two departures from the assumptions act in opposite directions. Correlated fitting errors lower $\mathbb{E}[\max_a \epsilon]$ when variances are equal (Remark 10). Unequal action frequencies, the normal case in fulfilment, give rare actions noisier estimates, and the greedy rule then favours them: for $K = 2$ the expected spurious gain is $\sqrt{(\sigma_1^2 + \sigma_2^2)/(2\pi)}$, driven by the noisier, rarer action. The same mechanism offers a second, non-decoupling explanation of the single-mode policy in §6.3.

### 4.3. Detectability and learnability

**Proposition 3** (Detectability frontier)**.** *Let* $df_1 = \dim(\mathcal{M}_1) - \dim(\mathcal{M}_0)$*, let* $n_{\text{info}}$ *be the information count of §3.4 and* $\nu = n_{\text{info}} - \dim(\mathcal{M}_1)$*. For the F-test of* $\mathcal{M}_0$ *against* $\mathcal{M}_1$ *in a homoscedastic Gaussian linear model with* $n_{\text{info}}$ *independent rows, at level* $\alpha$ *and power* $1 - \beta$*, the smallest detectable partial coupling is*

$$\eta^2_{\min} = \frac{\lambda_{\alpha,\beta,df_1}}{n_{\text{info}} + \lambda_{\alpha,\beta,df_1}}, \tag{8}$$

*where* $\lambda_{\alpha,\beta,df_1}$ *is the exact noncentrality at which the noncentral* $F_{df_1,\nu}$ *test has power* $1 - \beta$*.*

This is Cohen's minimum detectable effect [51], written for a nested contrast. The noncentrality is computed exactly: the large-$df_1$ approximation understates $\lambda$ by 25% at $df_1 = 21$ and lowers the threshold, the unsafe direction for an instrument that licenses spending (Appendix B.3). Because $\eta^2_{\min}$ grows with $df_1$, a richer state specification can make a genuinely coupled log undetectable, so the specification is fixed in advance and reported with $(df_1, \nu)$.

Equation (8) is exact only under its model. Importance weights, clustering and the estimated propensity all move the sampling distribution of the test as actually run. The verdict therefore uses the calibrated threshold

$$\eta^2_{\min,\text{cal}} := \inf\left\{\eta^2 : \Pr_{\eta^2}(\text{test rejects}) \geq 1 - \beta\right\}, \tag{9}$$

estimated by injecting known coupling into data resampled on the footing of §3.4 (Algorithm 2). Equation (8) remains the planning formula: it sizes pilots and is checked against (9) in every audit. Detectable coupling is necessary but not sufficient for a licence; the gap between detecting an effect and estimating it well enough to act on it is the policy-learning analogue of the detection–estimation gap in minimax testing [52, 53, 54].

*Conjecture* (Learnability frontier, a working hypothesis). For a fixed learner there is a threshold $\eta^2_{\text{learn}}(n_{\text{eff}}) \approx \kappa'/n_{\text{eff}}$ above which the learner certifies an improvement on an honest split, with $\kappa'$ depending on the learner and the estimator.

The $1/n_{\text{eff}}$ rate is suggested by regret bounds for policy learning [26, 27, 28]; it is not proved here, and no verdict in this paper rests on it. If it holds,

$$\frac{\eta^2_{\text{learn}}}{\eta^2_{\min}} \approx \frac{\kappa'}{\lambda_{\alpha,\beta,df_1}}, \tag{10}$$

so the band between the two frontiers is non-empty if and only if $\kappa' > \lambda_{\alpha,\beta,df_1}$. Its one testable implication is that $\eta^2_{\text{cert}}\, n_{\text{eff}}$ is constant across sample sizes (Appendix B.5). Whenever the deployed learner can be run on the log, direct certification (§5.1) replaces the conjecture.

### 4.4. Reusing a fold: the false-positive rate of a certified improvement

A common shortcut fits a value model, selects the greedy policy and evaluates it with a doubly robust estimator [29] on the same data. On a decoupled log the true improvement is zero, so every interval that excludes zero is a false licence. This subsection gives the rate at which that happens for learners that select actions state by state.

Consider $S$ states with $n_s$ rows each ($n = Sn_s$), $K = 2$ actions logged with $\pi_b(1 \mid s) = p \in (0, 1)$, and outcomes $Y = g(s) + e$ with $\mathbb{E}e = 0$, $\operatorname{Var} e = \sigma^2 < \infty$, independent of the action, so $D = 0$. The learner estimates $\hat{Q}(s, a)$ by the cell mean, plays $\hat{\pi}(s) = \arg\max_a \hat{Q}(s, a)$, and reports the in-sample doubly robust estimate $\hat{\Delta} = n^{-1} \sum_i \phi_i$ with

$$\phi_i = \hat{Q}(s_i, \hat{\pi}(s_i)) + \frac{\mathbf{1}[a_i = \hat{\pi}(s_i)]}{\pi_b(a_i \mid s_i)}\big(y_i - \hat{Q}(s_i, a_i)\big) - y_i,$$

its influence-function standard error $\widehat{\text{SE}} = \text{sd}(\phi)/\sqrt{n}$, and the Wald decision $|\hat{\Delta}| > z\widehat{\text{SE}}$ with $z = z_{1-\alpha/2}$. The overlap of this design is $H = \frac{1}{2}\big(p^{-1} + (1-p)^{-1}\big)$, with $H = 2$ under uniform logging.

**Theorem 2** (Reuse error of state-wise greedy selection)**.** *In the setting above, let* $n_s \to \infty$ *with* $S$ *and* $p$ *fixed. Then*

$$\frac{\hat{\Delta}}{\widehat{\text{SE}}} \Rightarrow T_S := \frac{\sum_{s=1}^S c_s |Z_s|}{\left(\sum_{s=1}^S c_s^2\right)^{1/2}}, \qquad c_s = \begin{cases} \sqrt{(1-p)/p}, & Z_s > 0, \\ \sqrt{p/(1-p)}, & Z_s \leq 0, \end{cases} \tag{11}$$

*with* $Z_1, \ldots, Z_S$ *independent standard normal, and the false-licence rate converges to* $\Pr(T_S > z)$*. Moreover:*

- (i) One selection. $T_1 = |Z_1|$*, so with a single state the rate converges to* $\alpha$ *for every* $p$*. The same holds for any learner whose greedy rule is one global comparison, such as a pooled model with a single action effect.*
- (ii) Many selections. *With* $\mu_S = \sqrt{SH/(\pi(H-1))}$*,* $(T_S - \mu_S)/\varsigma(H) \Rightarrow \mathcal{N}(0, 1)$ *as* $S \to \infty$*, where* $\varsigma(H)$ *is given in (27) and* $\varsigma(2) = \sqrt{1 - 2/\pi} \approx 0.60$*. The rate therefore tends to one, and* $\Phi((\mu_S - z)/\varsigma)$ *approximates it.*
- (iii) Overlap. $\mathbb{E}[T_S \mid c] = \sqrt{2/\pi} \sum_s c_s / \|c\|$*, so* $\mathbb{E}[T_S] \leq \sqrt{2S/\pi}$*, with equality for* $S \geq 2$ *if and only if* $p = \frac{1}{2}$*. The centring* $\mu_S$ *decreases in* $H$*, from* $\sqrt{2S/\pi}$ *under uniform logging to* $\sqrt{S/\pi}$ *as* $p \to 0$*. Overlap moves* $\mu_S$ *by at most a factor* $\sqrt{2}$*; the number of selections moves it without bound.*

**Table 3**
Limit false-licence rate $\Pr(T_S > z)$ of Theorem 2 at $\alpha = 0.05$, by number of states and logging probability, from $4\times10^5$ draws of (11). Simulations of the full procedure with $n_s = 5{,}000$ ($S \le 10$) and $n_s = 500$ ($S = 30$) agree within Monte-Carlo error.

| | $p$ ($o_{\text{eff}}$) | | |
|---|---|---|---|
| $S$ | 0.50 (1.00) | 0.20 (0.64) | 0.05 (0.19) |
| 1 | 0.050 | 0.050 | 0.050 |
| 3 | 0.168 | 0.129 | 0.104 |
| 10 | 0.823 | 0.611 | 0.418 |
| 30 | 1.000 | 1.000 | 0.982 |

(iv) Split. *If $\hat{Q}$ and $\hat{\pi}$ are fitted on a fold independent of the one used to evaluate them, $\mathbb{E}[\hat{\Delta} \mid \hat{\pi}] = 0$ and the rate converges to $\alpha$ for every $S$ and $p$.*

The proof is in Appendix B.6. Its key step is that, with cell means, the in-sample residuals of each cell sum to zero, so the inverse-propensity term contributes nothing to $\hat{\Delta}$ while contributing fully to $\widehat{\text{SE}}$: in each state $\hat{\Delta}$ reduces to the direct-method gain of a two-armed argmax, and the standard error to $\sigma c_s$.

Table 3 shows the two parameters at work. The number of local selections governs the size of the error: with thirty states the shortcut certifies a non-existent improvement with limiting probability at least 0.98 at every overlap level in the table. Overlap governs its direction: for every $S \ge 2$ in our grid the rate falls as logging moves away from uniform ($p \in \{0.5, 0.35, 0.2, 0.1, 0.05, 0.02\}$, $S \le 20$), so near-random logging is the most exposed regime, but only within the factor of part (iii). Overlap does not by itself drive the error: at $S = 1$ it has no effect.

*Remark* 2 (A correct standard error makes the error larger). Under reuse the influence-function standard error overstates the sampling variability of $\hat{\Delta}$: $n\,\mathbb{E}[\widehat{\text{SE}}^2] \to \sigma^2(H-1)$ while $n\,\text{Var}(\hat{\Delta}) \to \sigma^2(H - 1 - H/\pi)$ (Lemma 1). Replacing it by a consistent estimate of $\text{sd}(\hat{\Delta})$, for example a bootstrap of the whole fit-and-evaluate pipeline, raises the false-licence rate to approximately $\Phi\big(\sqrt{S}\,h(o_{\text{eff}}) - z\big)$ for large $S$, with $h(o) = [\pi(1 - o/2) - 1]^{-1/2}$; at $S = 10$ this is 0.99 under uniform logging, against 0.82 with the influence-function standard error. A better interval does not repair reuse; only a split does.

*Remark* 3 (Other learners). Theorem 2 covers the two ends of a spectrum. A pooled learner makes one selection and sits at $\alpha$ (part (i)); a saturated learner makes $S$ independent selections. Flexible learners lie between, with an effective number of selections that depends on how much they share across states, and their rate is measured by simulation on decoupled data with the audit's own design (§6.5). For such learners the doubly robust correction no longer vanishes, and its effect is summarised by the optimism constant of Definition 3.

**Definition 3** (Optimism constant)**.** *For a learner with greedy policy $\hat{\pi}$ and in-sample doubly robust estimate $\hat{\Delta}$ on decoupled data, $\tau := 1 - \mathbb{E}[\hat{\Delta}]/\mathbb{E}[\hat{\Delta}_{\text{DM}}]$ is the signed fraction of the direct-method optimism removed by the correction. For the saturated learner $\tau = 0$ exactly; $\tau < 0$ means the correction amplifies the optimism (Remarks 12 and 13).*

The practical rule follows. Folds are kept disjoint in every audit, because that is what guarantees validity (part (iv)), whatever the learner and whatever the overlap. The coupling test is run first because it is cheap and can end the decision before a pipeline is built; that ordering is an efficiency rule and is the sense in which this paper puts coupling before overlap.

# 5. The instrument: a design-science artefact for decision feasibility

## 5.1. The audit and the gate

The audit runs in three stages (Figure 1). **Stage 0** applies the gates C1 and C2, screens identification with the local statistic C4a, and reports the diagnostics $\widehat{H}$, $o_{\text{eff}}$, C5a and C5b together with the information route of §3.4. **Stage 1** runs the coupling test C3 for each policy class (the main-effect contrast for $\Pi_{\text{const}}$, the full contrast (4) for $\Pi$), the saturation ladder, and, when the objective is sequential, condition C6. **Stage 2** certifies a fixed learner on disjoint folds [19]. Every choice that could move a verdict (contrast, ladder, binning, segmentation, learner, thresholds) is fixed in a written plan before the target data are analysed (Appendix C.4).

*The detectability gate.* The verdict compares a coupling interval $[\hat{L}, \hat{U}]$ with a threshold interval $[\eta^2_{\text{min,lo}}, \eta^2_{\text{min,hi}}]$, both computed on the information footing of §3.4:

$$\text{verdict} = \begin{cases} \text{below detectability}, & \hat{U} < \eta^2_{\text{min,lo}}, \\ \text{robustly detectable}, & \hat{L} > \eta^2_{\text{min,hi}}, \\ \text{indeterminate}, & \text{otherwise}. \end{cases} \qquad (12)$$

The threshold interval comes from the uncertainty in the information count (the jackknife interval of $\bar{\delta}$ in (5)) or, for the calibrated threshold (9), from its Monte-Carlo error. The branches are exhaustive and mutually exclusive. The rule asserts neither side when the data cannot separate the coupling from its threshold, and narrowing either interval can only move a verdict out of indeterminate, never into it, so an abstention is resolved by more data or a pilot. Both intervals refer to the same support, contrast and weighting; comparing a coupling from the full log with a threshold from a trimmed log, or from another ladder rung, is an error of bookkeeping, not of judgement.

With few clusters the percentile bootstrap is not a safe input to the second branch. For an $R^2$-type statistic the bootstrap world treats the fitted action block, noise included, as the truth, so the bootstrap distribution sits above the sampling distribution and the percentile lower limit is biased upward, the permissive direction. The robustly detectable branch is therefore decided by a test at the boundary: the fitted action block is rescaled so that its coupling equals $\eta^2_{\mathrm{min,hi}}$, data are resampled around that null on the same footing, and the branch is taken if the observed $\hat{\omega}^2$ falls in the upper $\alpha$ tail (Appendix A.5). The below branch uses the upper limit, which is conservative for a refusal, and is confirmed by the mirror test at $\eta^2_{\mathrm{min,lo}}$.

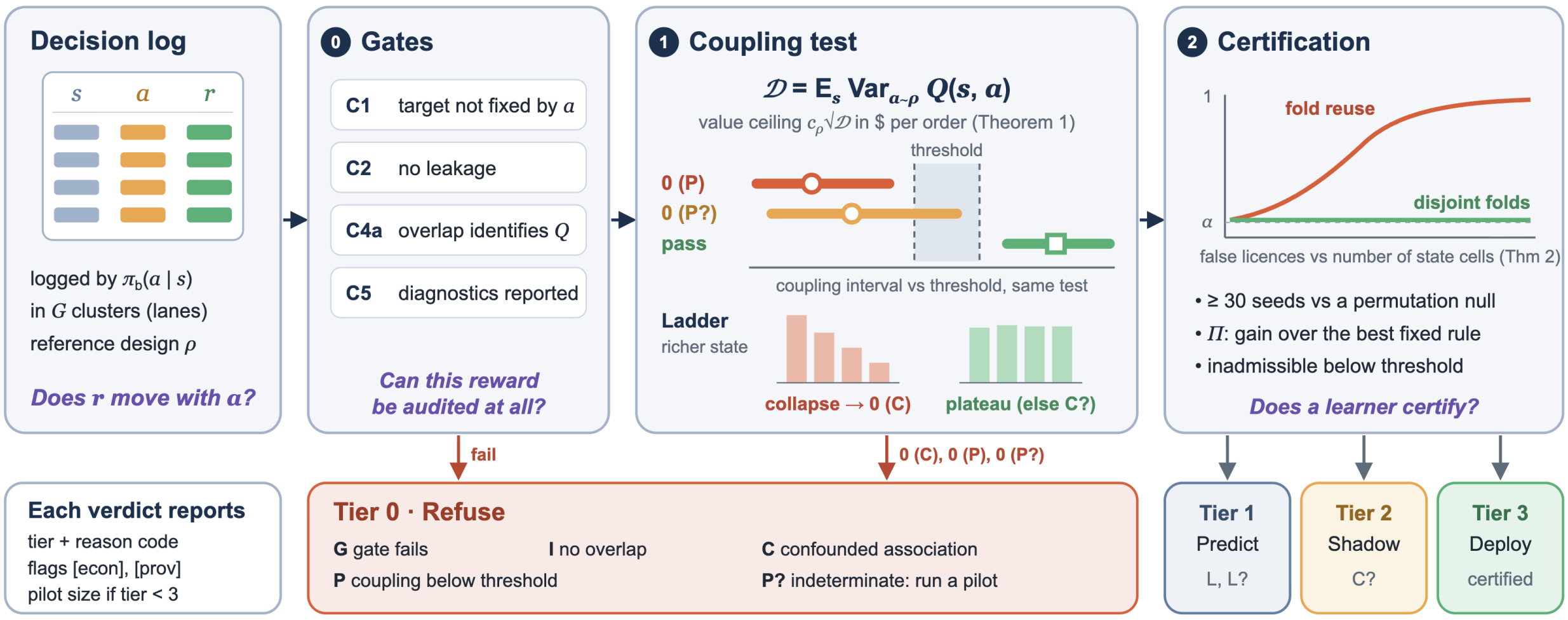

**Figure 1:** Overview of the audit. A decision log passes three stages, each of which can stop it. Stage 0 checks that the reward can be audited at all. Stage 1 estimates the action–outcome coupling $\mathcal{D}$, converts it into a value ceiling in dollars per order (Theorem 1), and compares its interval with a detectability threshold calibrated on the same test (12); a saturation ladder checks whether the coupling survives richer state models. Stage 2 certifies a fixed learner on disjoint folds, because under fold reuse the false-licence rate grows with the number of state cells (Theorem 2). The routing rule of §5.2 assigns one of four tiers to each policy class, with a reason code for the stage at which the audit stopped.

The first branch of (12) is an equivalence test with margin $\eta^2_{\mathrm{min}}$: it concludes that the coupling is smaller than what the design can detect. The same construction with the break-even coupling of the economic gate as margin gives the decision-relevant version, two one-sided tests in the sense of [55] and [56]: if the upper limit lies below break-even, the log is refused on economic grounds whatever its detectability. The detectability margin answers whether the data can see the effect; the break-even margin answers whether seeing it could pay.

*Certification.* When the deployed learner can be run on the log, its improvement is certified directly rather than inferred from Conjecture 4.3. The protocol (Appendix C.2) fixes the learner, uses disjoint folds (a rolling origin for time-ordered logs), repeats over $S \geq 30$ seeds, and declares certification when the proportion of seeds whose doubly robust interval excludes zero exceeds $q$. The seed runs share one dataset and are strongly dependent, so $q$ is calibrated on a permutation null that re-runs the whole pipeline with actions permuted within strata; with $B$ permutations the smallest attainable $p$-value is $1/(B+1)$, and $B \geq 999$ is used. One constraint links the stages: certification at a coupling below the detectability threshold is inadmissible. Theorem 2 and Proposition 2 show that a greedy learner produces confident improvements on decoupled data, so a certificate that the coupling test cannot support is more plausibly selection on fitting error than signal.

*The economic gate.* Detectability is not managerial relevance. Theorem 1 turns a coupling into an upper bound on value per decision, so the audit reports the break-even coupling $\mathcal{D}_{\mathrm{be}}$ at which the ceiling equals the fully loaded cost per decision, on the absolute scale and on both normalised scales (Appendix C.1). A self-check verifies the conversion: because the ceiling scales with $\sqrt{\mathcal{D}}$, the coupling-scale ratio must equal the square of the money-scale ratio,

$$\frac{\eta_a^2}{\eta_{a,\mathrm{be}}^2} = \left(\frac{c_\rho\sqrt{\mathcal{D}}}{c^{\mathrm{be}}}\right)^2, \tag{13}$$

and both sides are printed with each economic statement. Because the ceiling is an upper bound, the economic gate is stated as an upper bound on net present value. With $N_t$ decisions in year $t$, operating cost $o_t$, build cost $b$, discount rate $i$, horizon $T$ and $\mathcal{D}^+$ the upper confidence limit of $\mathcal{D}$ in money units,

$$\mathrm{NPV}_{\max} = \sum_{t=1}^{T} \frac{N_t\, c_\rho \sqrt{\mathcal{D}^+} - o_t}{(1+i)^t} - b. \tag{14}$$

The flag [econ] is raised when $\mathrm{NPV}_{\max} < 0$: even capturing the whole ceiling would not repay the programme. Realised value is a fraction $r \in (0, 1]$ of the ceiling; $r$ is reported as a sensitivity map over $(b, r)$ and is never transferred from one log to another, since it depends on the learner, the reward and the state resolution.

**Table 4**
The four classes of decision support. A distinction earns a tier only when it changes what the data may be used for. Each verdict names its policy class; the reason the audit stops is carried by the codes of Table 5, and two flags qualify any tier. Every tier below 3 carries an exploration-pilot branch.

| Tier | Name | What the data support | Reasons |
|---|---|---|---|
| 0 | Refuse | nothing for this (dataset, reward, reference design, policy class) | G, I, C, P, P? |
| 1 | Predict | calibrated prediction and delivery-date quoting with coverage; no policy verdict | L, L? |
| 2 | Shadow | policy optimisation on the logged distribution; measured shadow deployment, not rollout | C? |
| 3 | Deploy | offline policy optimisation within the named policy class | — |

## 5.2. The routing rule

The gate routes a log to the highest tier it can justify for each policy class, and records separately why it stopped. Tiers and reasons are orthogonal: four tiers (Table 4) and eight reason codes (Table 5) describe every outcome of the rule. The rule is lexicographic and stops at the first condition met:

(1) **Gates.** If C1 or C2 fails, 0 (G): no coupling verdict for this reward, since $Q(s, a)$ is not meaningful. If C4a fails, 0 (I): the data are silent, not negative.

(2) **Ladder.** The ladder re-estimates $\hat{\omega}^2$ over a pre-specified sequence of richer state models (Appendix C.2). It has three outcomes, mirroring (12). *Collapse*, 0 (C), requires two things: the upper confidence limit of the noise-adjusted ratio of the richest to the base rung lies below $\delta_{\text{col}}$, and the enrichment covariates significantly improve a held-out propensity model for the action ($\Delta_{\text{prop}} > 0$). A decline without the second condition is as consistent with variance absorption as with confounding and licenses no causal reading; on a randomised log the second condition cannot hold, so collapse cannot fire. *Plateau* requires the upper confidence limits of two consecutive relative changes to lie below $\delta_{\text{pl}}$. Anything else is *indeterminate*, adds C? and caps the verdict at Tier 2. Collapse is checked before detectability and overrides it: a confounded association is not rescued by being precisely measured.

(3) **Detectability.** Apply (12): below gives 0 (P), indeterminate gives 0 (P?), robustly detectable continues.

(4) **Certification.** Not certified gives 1 (L), indeterminate gives 1 (L?), certified continues to Tier 3, or to Tier 2 if the ladder added C?.

(5) **Flags.** Add [econ] if $\text{NPV}_{\text{max}} < 0$ in (14). If the objective is sequential, run C6: if $\hat{\mathcal{D}}_{\text{cont}}$ is detectable, or C6 cannot be run, the single-step verdict is marked [prov], because the one-step audit then understates or misstates the coupling (Proposition 1); a negative $\hat{\chi}_{\text{seq}}$ is reported with the warning that the one-step coupling overstates what a sequential policy can exploit.

The rule is applied separately to $\Pi_{\text{const}}$ and $\Pi$, and the two verdicts are reported side by side. Because $\Pi_{\text{const}} \subset \Pi$, a $\Pi$ verdict stated on the same quantities as the $\Pi_{\text{const}}$ verdict would be implied by it and would carry no information. The $\Pi$ verdict is therefore stated on the *increment* that personalisation adds. Its detectability test is the interaction coupling $\mathcal{D}_{\text{int}}$ of (3), measured at a pre-specified estimable contrast, and its certification is of the paired difference between the contextual policy and the certified non-contextual rule on the same evaluation fold. A log can then be licensed for fixed rules and refused for personalisation, which is the case we meet in §6.2, and a reader cannot mistake a $\Pi$ licence for a restatement of the $\Pi_{\text{const}}$ one. Aggregate and segment verdicts are reconciled by reporting a surviving segment as a scoped verdict for that segment only, never as an upgrade of the dataset (Appendix C.3).

*From refusal to plan.* A refusal without a plan has limited managerial value. Inverting (8) gives the information needed to detect a target coupling at the chosen power; Theorem 1 maps a target gain per decision to that coupling; and Definition 2 shows that uniform randomisation maximises overlap. Together they give an exploration budget: for this reward and action set, randomise $x$% of orders for $y$ weeks to reach power $1-\beta$ against a coupling worth \$$z$ per order. The budget is a price, not a promise, and every tier below 3 reports it, so each refusal comes with the experiment that could overturn it.

## 5.3. Operating characteristics

An artefact that issues verdicts must report its own error rates, and the two directions are not symmetric. A false kill demotes a genuinely prescriptive log below Tier 3 and destroys value. A false licence certifies a decoupled log and permits the failure this paper is about. Because (12) may abstain, three reporting rules apply: directional error rates are reported conditional on a verdict, alongside the abstention rate broken out by the abstaining branch; the two directions are never pooled; and adjacent-tier misroutings are reported by pair. On a finite grid a zero count is reported as $0/m$ with its one-sided 95% bound of about $3/m$, not as a measured zero (Appendix C.4).

# 6. Empirical evaluation

## 6.1. Design

The evaluation has two legs, kept apart: no simulator constant is re-fitted on log data, and no field result tunes a threshold. Three public logs test whether the audit stops where it should (§6.2–§6.4); controlled studies with known truth measure its error rates (§6.5). Every verdict is the three-branch rule (12), comparing a coupling interval on the partial scale $\eta^2_{a|s}$ with a threshold calibrated on the test that produced

**Table 5**
Reason codes. A trailing "?" has one meaning throughout: the three-branch rule abstained, and more data, more seeds or a larger split would resolve it. Codes combine; a Tier-1 verdict may carry L and C?.

| Code | Fires when | Indicated response |
|---|---|---|
| G | C1 or C2 fails | redefine the outcome or audit another reward, then re-enter at Stage 0; distinct from I, since overlap is not what failed |
| I | C4a fails | randomise; exploration is most clearly indicated |
| C | ladder collapses (observational logs only) | do not treat the association as coupling; randomise to sever the masking mechanism (Appendix A.1), or add the pre-decision covariate the collapse implicates, then re-audit |
| C? | ladder has not plateaued | observational log: ignorability not yet corroborated, so stabilise the state specification or confirm by randomisation before scale; randomised log: the coupling still grows with the state (effect modification), so the state specification, not the causal reading, is unsettled |
| P | coupling below detectability | use as a negative control; pilot sized by the reported RV and by (8) |
| P? | detectability indeterminate | pilot sized to resolve the frontier; report which interval binds |
| L | deployed learner does not certify | more data or wider exploration: the effect is detectable but underpowered for policy |
| L? | certification indeterminate | more seeds or a larger honest split before deciding |

**Table 6**
Every verdict in the evaluation. Couplings and thresholds are $\eta^2_{a|s} \times 10^{-3}$. Thresholds are calibrated on the test that produces the interval (Algorithm 2): the randomisation for OBD, the wild cluster bootstrap for the two logistics logs. Full audits are in Appendix D.1 (OBD) and Appendix D.3 (DataCo, Olist).

| Corpus (reward) | Class | coupling $[\hat{L}, \hat{U}]$ | threshold | branch | ladder | certificate | **Verdict** |
|---|---|---|---|---|---|---|---|
| OBD (click) | $\Pi_{\text{const}}$ | $[0.513, 1.596]$ | $[0.025, 0.029]$ | above (18×) | plateau | 26/30, $p = 0.003$ | **Tier 3** [econ] |
| | $\Pi$ (increment) | $0.601^a$ | — | detectable | C? | 0/30 | **1 (L)** |
| DataCo (on-time) | both | C1 fails: target is an identity of the action in 97.55% of rows | | | | | **0 (G)** |
| DataCo (margin) | $\Pi_{\text{const}}$ | $[0.000, 0.295]$ | $[0.235, 0.354]$ | overlaps | — | — | **0 (P?)** |
| | $\Pi$ | $[0.066, 1.05]$ | $[1.09, 1.73]$ | below | — | — | **0 (P)** |
| Olist (review), dev. | $\Pi_{\text{const}}$ | $[0.00, 0.96]$ | $[1.55, 2.33]$ | below | not read[b] | — | **0 (P)** |
| | $\Pi$ | $[0.30, 4.08]$ | $[9.40, 16.9]$ | below | C? | 29/30, inadmissible | **0 (P)** |
| Olist, held-out | $\Pi_{\text{const}}$ | $[0.46, 4.67]$ | $[2.66, 4.46]$ | overlaps | C? | — | **0 (P?)** |
| | $\Pi$ | $[1.46, 7.22]$ | $[9.03, 17.1]$ | below | C? | — | **0 (P)** |

*Notes.* [a] interaction coupling at item×`user_feature_0` ($df_1 = 237$), permutation $p = 0.005$; the certified quantity is the paired uplift over the fixed rule, $-0.00145$ $[-0.00357, +0.00066]$ click per impression. [b] the base rung is indistinguishable from zero, so the ratio is unstable (Table S9). Contrasts are $df_1 = 79$ for OBD, 3 and 21 for DataCo, 3 and 9 for Olist.

the interval (Algorithm 2); only the saturation ladder is read on the absolute $\hat{D}$ (step 2). Each corpus carries one verdict per policy class: the main-effect contrast bounds $\Pi_{\text{const}}$, and $\Pi$ is judged on what personalisation adds over the best fixed rule (§5.2). Table 6 lists every verdict, Figure 2 places the three logs on the coupling axis, and Appendix D gives each audit in full.

The outcome is one licence and two refusals, each stopping at a different stage. A randomised e-commerce log is licensed for fixed rules and refused for personalisation. A logistics benchmark fails the definitional-target gate. A second logistics log, on which a conservative learner certifies in 29 of 30 seeds, is refused because its coupling does not clear what 27 clusters can detect.

## 6.2. A randomised log the instrument licenses

The Open Bandit Dataset [15] logs fashion recommendations under a uniform policy, $\pi_b \equiv 1/K$, so overlap is perfect ($o_{\text{eff}} = 1$) and A1 holds by design. It is not a logistics log; it serves as the positive control. The reward is the click, the action the recommended item ($K = 80$), the state the four user-feature strata, and the cluster unit the 404 strata they form. Gates C1, C2 and C4a pass (Appendix D.1).

*Fixed rules reach Tier 3.* Only the main-effect contrast is estimable ($df_1 = 79$; the interacted contrast would need 32,074 degrees of freedom against 404 clusters), so the audit measures a non-contextual coupling. Items are randomised independently of the cluster unit, so detectability is decided against the randomisation itself: Algorithm 2 on a within-cluster permutation null gives $\eta^2_{\text{min,cal}} = 0.027 \times 10^{-3}$ $[0.025, 0.029]$, and the coupling's lower limit, $0.513 \times 10^{-3}$, exceeds the upper end by a factor of 18. The main-effect ladder is flat to within 0.5% over four enrichments (Appendix D.2). The deployed shrunk tabular learner certifies in 26/30 seeds against a permutation null with $B = 999$ ($p = 0.003$), and the routing rule returns Tier 3 for non-contextual rules.

*Personalisation is refused at certification.* The increment personalisation would add is detectable: the interaction coupling at item×`user_feature_0` is $0.601 \times 10^{-3}$ ($df_1 = 237$, permutation $p = 0.005$). It does not certify. The paired uplift of the contextual policy over the certified fixed rule is −0.00145 [−0.00357, +0.00066] click per impression and no seed certifies it, so Π returns 1 (L): the effect exists, and these data cannot turn it into value. The interval contains zero, so the finding is that the log rules out a gain above +0.00066, not that personalisation loses. The covariate ladder rises from 0.70 to $10.31 \times 10^{-3}$; on a uniform log $\Delta_{\text{prop}} = 0$ by construction, so the rise is effect modification and is recorded as C?. A per-rung permutation null excludes a degrees-of-freedom artefact (Appendix D.2).

*The ceiling is loose by a measurable amount.* The ceiling $c_\rho\sqrt{D_{\text{main}}}$ is 0.0139 click per impression, and the certified policy reaches +0.00411, or 29.7% [15.7%, 43.7%] of it. The best single item, chosen on the training days, reaches 28.4%, so the shortfall is the looseness of the bound at $K = 80$, not of the learner (Remark 1; Table S4). Three further checks leave the verdict unchanged: the doubly robust estimator recovers a concurrent on-policy value to within 1.6%, neither the contextual rule nor the platform's Thompson policy outperforms the fixed rule, and pooled rolling-origin splits reproduce the uplift. Every choice for this corpus was made with the data in hand, so the audit is exploratory: it shows that the instrument *can* license a log, not that it does so under a plan fixed in advance (Appendix D.1).

## 6.3. A benchmark that fails a gate

DataCo [17] is the de facto public benchmark for learning supply-chain decisions; the action is the committed shipping mode. Its main service target fails C1 before any estimation. The late-delivery flag equals $\mathbf{1}[\text{realised} > \text{scheduled}]$ in 97.55% of rows, and the scheduled promise is constant within each mode, so the target is an accounting identity of the action and the verdict is 0 (G). No sample size, overlap restriction or model can repair a definitional target. A conservative offline agent (discrete CQL) trained on the log shows the consequence: it certifies only under reward weightings that emphasise the on-time component, and the certified policy raises the on-time metric by 23.3 points by lengthening the promise by 1.07 days, with no change in realised delivery time (Figure S1; Table S6). Realised transit time also shows zero within-mode variance for one mode, so we treat DataCo as a benchmark of uncertain provenance rather than field evidence, and report the C1 failure because of the benchmark's prominence (Appendix D.4).

On the margin reward the audit is underpowered, and the two policy classes stop at different branches. Logging is close to random with respect to recorded state (C5b = −0.009 nats), and the log has only 23 lanes. The design effect of the action block is $\bar{\delta} = 3.35$ at $df_1 = 3$ and 3.60 at $df_1 = 21$, not the value of one that a zero intraclass correlation would suggest (Appendix D.7). On the calibrated footing the fixed-rule coupling $[0.000, 0.295] \times 10^{-3}$ overlaps its threshold [0.235, 0.354], giving 0 (P?), and the contextual coupling [0.066, 1.05] lies below [1.09, 1.73], giving 0 (P). The two codes lead to different actions: P licenses use of the log as a negative control, while P? licenses only a pilot sized to resolve the frontier (Appendix C.3). Solving (8) at 80% power, verifying the measured coupling would take about 95,000 randomised orders for fixed rules and 187,000 for contextual ones, or 1.4 and 2.8 years at the log's volume; a pilot powered at break-even would take about 21 years. The economic gate points the same way: the ceiling is \$1.92 per order, and the programme repays its \$90,000 build cost only at a realisation rate above 0.26, which nothing on this log supports (Appendix D.4).

## 6.4. A corpus where a clean certificate is inadmissible

In the Olist marketplace log [23] the action is the delivery promise ($K = 4$ bins) and the reward is the customer review. The log passes the gates and Stage 0, but only after trimming: lane geography predicts the promise (C5b = +0.27 nats), so some propensities are genuinely very small, and the 0.16% of rows below the pre-specified trim $\pi_b \geq 0.01$ would cut $n_{\text{eff}}$ from 17,133 to 2,905. This is why C4a and C4b are reported separately (Appendix D.5).

*The certificate.* The deployed conservative learner (CQL) certifies in 29/30 honest-split seeds, with mean uplift +0.023 review points, against a permutation null whose 95th percentile is 0.25 ($p < 0.01$). By the usual standard of offline evaluation, the log is ready for deployment.

*Why the audit refuses it.* The coherence constraint (Appendix C.2) forbids certification at a coupling that does not clear its detectability threshold, and at 27 clusters this coupling does not. At the headline main-effect contrast the development coupling is $[0.00, 0.96] \times 10^{-3}$ against a calibrated threshold of [1.55, 2.33]. At the richest interacted contrast the clusters can carry ($df_1 = 9$; the planned $df_1 = 36$ is rank-deficient at $G = 27$) it is [0.30, 4.08] against [9.40, 16.9]. Both cells are 0 (P). By Proposition 2, a certificate at such a coupling is more plausibly selection on fitting error than signal, so the audit records it as unsupported. In one corpus, this is the central claim of the paper: a conservative learner produced a clean certificate on a coupling the design cannot resolve. No admissible trim changes the outcome (Table S10).

*No confounding claim.* The saturation ladder, recomputed on the audit's own trim, weights and clusters and read on $\hat{D}$, does not decline: $L4/L0$ is 1.17 [0.73, 2.57] on the development window and 1.29 [0.88, 2.03] on the held-out one, and the noise control stays within 4% of one (Table S9). The enrichment covariates also fail to predict the action on either window ($\Delta_{\text{prop}} = -0.191$ nats, [−0.358, +0.033] on development; −0.135, [−0.281, +0.067] on held-out). Neither condition for 0 (C) holds, so the ladder is C? and the verdict rests on detectability. Read on the normalised scale, a ladder can decline even when the absolute coupling does not (Appendix D.11); Appendix D.8 gives the ladder rung by rung together with the trim sweep.

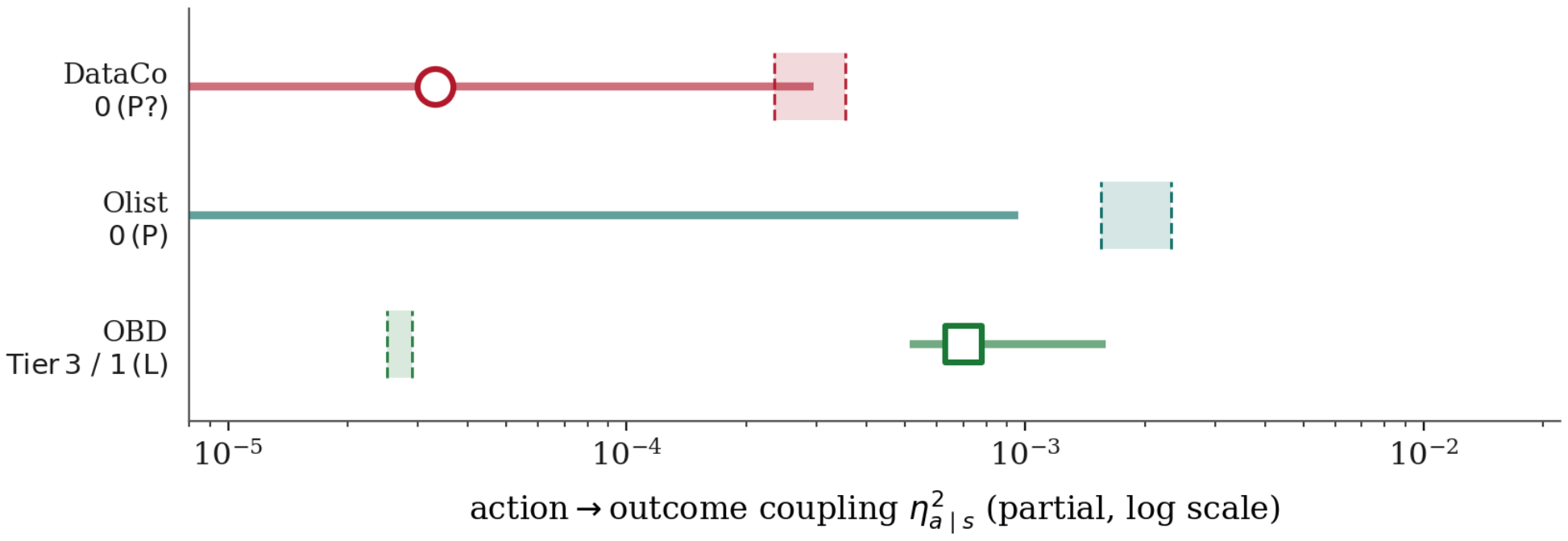


**Figure 2:** All three logs on the coupling axis, each against its own detectability threshold band (shaded), calibrated by Algorithm 2 as in Table 6. The three stop at three different places. Both logistics logs are drawn at the headline fixed-rule contrast $df_1 = 3$, where Olist's development interval lies below its band (0 (P)) while DataCo's overlaps it (0 (P?)); OBD's lower limit clears its band by 18×.

*A second window.* The final six months ($n$ = 38,923) form a second observation window rather than a confirmatory test, because the pipeline was corrected after their first results were seen (Appendix D.6). The refusal recurs there but its code does not: the interacted cell remains 0 (P), while at the main-effect contrast the coupling $[0.46, 4.67]$ overlaps its threshold $[2.66, 4.46]$, giving 0 (P?). On neither window can 27 clusters resolve the coupling well enough to license a policy. The refusal does not dismiss the data: under hierarchical Benjamini–Hochberg control, 6 of 37 lanes keep a significant association against 1.9 expected, and Table 4 advises calibrated quoting below Tier 3.

### 6.5. Controlled validation

Five studies with known truth measure what the field corpora cannot. Each is summarised here; designs, tables and further results are in Appendix D.

*The reuse limit holds for the finite procedure and on real outcomes.* Run as an analyst would run it (fit cell means, take the greedy policy, evaluate it doubly robustly on the same rows), the saturated learner matches the limit $\Pr(T_S > z)$ of Theorem 2 within three Monte-Carlo standard errors at four of five logging probabilities. It departs upward, the conservative direction, only where the rare action has 25 rows per state (Appendix D.9). The two parameters separate as part (ii) predicts (Figure 3): sweeping overlap from uniform to $p = 0.05$ changes the rate by a factor of 1.7, whereas sweeping the number of state cells from 1 to 32 takes it from 0.057 to 1.000. A pooled learner stays at $\alpha$ throughout, and a disjoint split returns every learner to $[0.043, 0.064]$, as part (iv) states. The CQL learner deployed on Olist behaves like a pooled one under reuse ($S_{\text{eff}} = 1.4$), because its value function is flat enough for the doubly robust correction to cancel the direct-method gain; its certificate is still computed on disjoint folds, since part (iv) is what guarantees validity for any learner. On real outcomes with the coupling removed by permutation, splitting keeps the rate at 0.030–0.065. Reuse rises with the number of strata and tracks the limit evaluated at the empirical cell sizes, with ratios of 0.84–1.18 (Table S12); this generalisation is an empirical finding and is not covered by the proof.

*At 23–27 clusters the test has to be chosen.* Interval and threshold must come from the same test. Of four candidates on designs matched to each corpus, only the constrained wild cluster bootstrap holds its size at every contrast, within $[0.056, 0.068]$ (Table S14). The cluster-robust Wald test over-rejects, increasingly with $df_1$. CR2 [57] holds its size at $df_1 = 3$ but at $df_1 \geq 9$ almost never rejects, and its own reference distribution implies thresholds two orders of magnitude above any attainable coupling. Thresholds are therefore calibrated on the wild bootstrap itself (Table S8). The same grid states what such logs can support: at each design's own threshold, the bootstrap resolves a main-effect contrast with probability 0.55–0.65 at 1× and 0.82–0.93 at 2×, but only 0.19–0.21 and 0.38–0.43 at $df_1 = 18$. A log with a few tens of clusters can license a fixed rule but not a contextual one, however many rows it has, which is why §5.2 routes the two classes separately (Appendix D.10).

*The audit licenses logistics designs when it should.* Both field logistics logs are refused, so a plasmode study tests licensing on their real designs. Weights, clusters, overlap, contrast and row count are kept; only the outcome is replaced, with a known action effect implanted at multiples of each design's calibrated threshold (Appendix D.12). At four times the threshold, fixed rules reach Tier 3 in 39/40 replications on the Olist design and 37/40 on the DataCo design, so the field refusals are not an artefact of designs that cannot yield a licence. With no implanted coupling, no verdict above Tier 1 was issued in 80 replications (rule-of-three bound 3.7%), nor in 160 replications in which the only action–outcome association runs through a confounder. With a proxy for the confounder in the ladder, the Olist design returns 0 (C) in 30/40. The DataCo design returns none, which is correct, because no recorded covariate predicts its action. At the threshold itself, 90% of cells return 0 (P?): the rule abstains where the data cannot decide.

*The ladder's collapse branch is conservative, at a stated cost.* On four scenarios with known structure (Appendix D.11),

**Table 7**
The audit against three baselines, on 1,800 simulated logs per policy class. B1 checks overlap and certifies on reused folds. B2 splits the folds and certifies the contextual policy against the logging policy. B2′ splits the folds and certifies the *increment* over the best fixed rule, but runs no coupling test and no ladder. B3 is the full rule. False-licence and false-kill rates are conditional on issuing a verdict, so B3's rates are computed on fewer verdicts than the baselines'; abstentions, rejection counts and recall are over the whole grid.

| | $\Pi_{\text{const}}$ (fixed rules) | | | | $\Pi$ (what personalisation adds) | | | |
|---|---|---|---|---|---|---|---|---|
| | B1 | B2 | B2′ | **B3** | B1 | B2 | B2′ | **B3** |
| False licence | 0.017 | 0.003 | 0.003 | **0.000** | 0.328 | 0.187 | **0.001** | **0.000** |
| False kill | 0.017 | 0.048 | 0.048 | 0.009 | 0.018 | 0.127 | 0.227 | 0.061 |
| Abstains, of 1,800 | 0 | 0 | 0 | 533 | 0 | 0 | 0 | 809 |
| Usable logs rejected, of 1,800 | 30 | 86 | 86 | **12** | 32 | 228 | 408 | **60** |
| True Tier-3 recall, of 480 | 450 | 394 | 394 | 285 | 448 | 252 | 72 | 32 |

two results matter. First, a ratio of normalised couplings can move in the wrong direction: a confounder that also predicts the outcome shrinks the denominator of $\hat{\omega}^2$ faster than its numerator. Reading the ladder on the absolute $\hat{D}$ removes the problem, and step 2 of the rule reads it there. Second, with that correction and a one-sided test of $\Delta_{\text{prop}}$, the branch collapses in 94% of replications when the whole coupling is spurious and never in the six scenarios where collapse would be wrong: variance absorption, a post-decision collider, and effect modification on a randomised log. Its failure mode is confounding mixed with a true effect, where it collapses in 40% of replications although 58% of the coupling is genuine.

*Against the procedures it replaces.* The full rule (B3) is run end-to-end on 1,800 simulated logs per policy class, with three baselines on the same data (Table 7). B1 checks overlap and certifies on reused folds. B2 splits the folds and certifies against the logging policy. B2′ splits the folds and certifies the increment over the best fixed rule, but runs no coupling test and no ladder. For fixed rules all four are defensible, with false-licence rates of 1.7%, 0.3%, 0.3% and 0.0%. For contextual rules they separate. B1 licenses 32.8% of decoupled logs and B2 18.7%, because splitting repairs the inference but not the question. B2′ brings the rate to 0.1%, so nearly the whole reduction comes from asking what personalisation adds, a change any analyst can adopt without this audit. The coupling test and the ladder change how the remaining logs are answered. Counted over the whole grid, B3 rejects 60 usable contextual logs outright against 408 for B2′, but it also abstains on 809 of the 1,800 logs and licenses 32 of the 480 true Tier-3 cells, against 72 for B2′: the reduction in outright rejections is absorbed by abstention, not converted into licences. What B3 adds is a third answer: an indeterminate verdict that comes with the pilot that would settle it, where B2′ can only license or reject. At these cluster counts the audit trades recall for protection, and it reports the abstention rate (Appendix D.13). The instrument's self-checks are also reported there: the bias of $\hat{\omega}^2$ under each corpus's own weights, ceiling tightness, null calibration and robustness values (Table S17).

## 7. Discussion

Three public logs show that the instrument locates why a log fails and that stages and policy classes are not interchangeable. The randomised log supports only fixed rules (Tier 3). For contextual rules, the audit asks what personalisation adds over the best fixed rule; since $\Pi_{\text{const}} \subset \Pi$, other comparisons are vacuous. Here the increment is detectable but not certifiable: its paired uplift over the certified fixed rule is $-0.0015$ $[-0.0036, +0.0007]$ click per impression, and no seed certifies it. Thus a log may suffice for fixed recommendations but not for personalisation, and the instrument identifies which.

Both logistics logs are refused, for different reasons. The public benchmark fails the definitional-target gate because the target is an accounting identity, so no sample size, overlap, or degrees-of-freedom choice can fix it (0 (G)). Its margin-reward audit is also underpowered at the available clusters (0 (P?) for fixed rules, 0 (P) for contextual ones), and the payoff ceiling repays the build cost only at a realisation rate that nothing on the log supports, so the statistical and economic gates agree. The second corpus passes estimability but is still refused: it has an importance-weight pathology that trimming controls and that a design-level overlap claim would have hidden. A conservative learner certifies in 29 of 30 seeds under a calibrated permutation null, but that certificate is inadmissible under the audit's coherence rule: with 27 clusters the coupling is below the detectability threshold, yielding 0 (P) in three window-by-class cells and 0 (P?) in the fourth, for fixed rules on the held-out window. Expanding the pre-specified state does not reduce the coupling on either window (Table S9), and the enrichment covariates do not predict the action, so the refusal is for power and no confounding claim is made.

Relative to the procedures it replaces, the audit's value is mainly in personalisation, and narrower than it appears. For fixed rules, all four procedures are defensible. For contextual policies, certifying on reused folds licenses about a third of decoupled logs; splitting folds still licenses about a fifth, because it fixes inference without changing the question. Asking the increment question, what personalisation adds

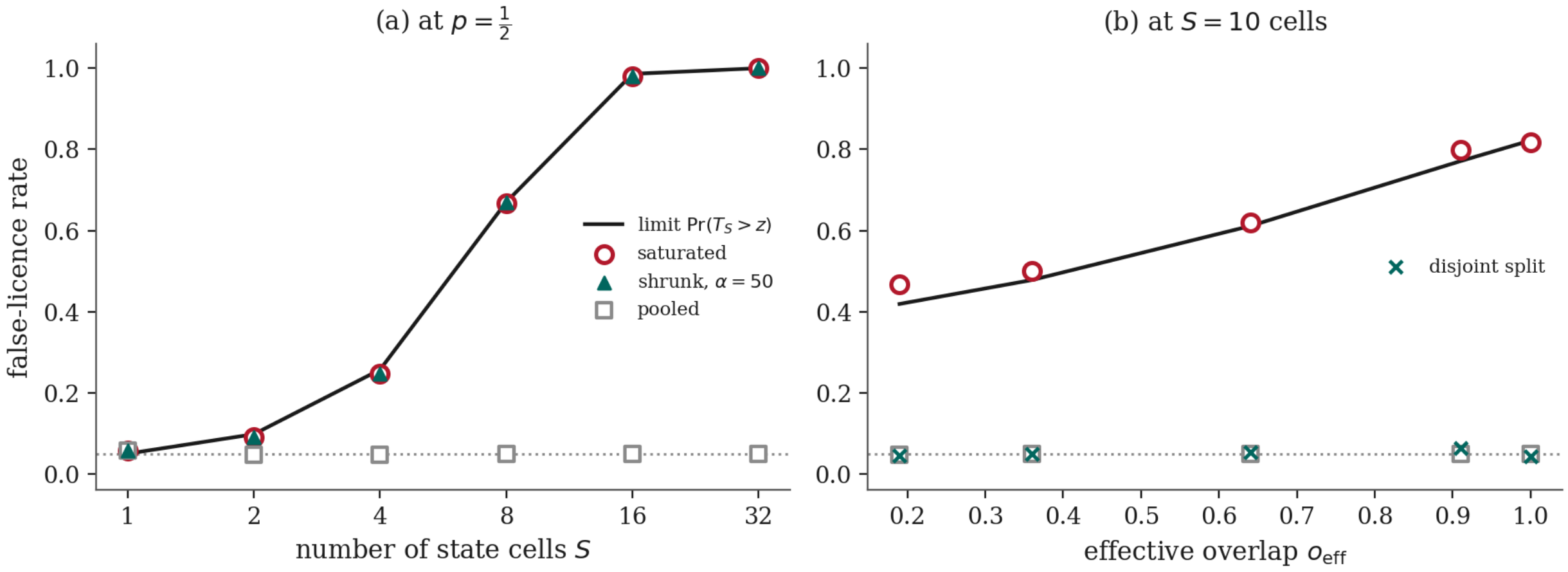


**Figure 3:** Theorem 2 against the finite procedure on decoupled data (true improvement = 0); the line is the exact limit $\Pr(T_S > z)$ and the points are fitted learners. Panel (a) sweeps the number of state cells and panel (b) the overlap, on the same vertical scale: the first takes the rate from $\alpha$ to one, the second by a factor of at most 2.33 over $p \in [0.02, 0.5]$ (the horizontal axis of panel (b) is the effective overlap $o_{\text{eff}}$). The pooled learner sits at $\alpha$ throughout both, and a disjoint split returns every learner there (Table S11).

over the best fixed rule, cuts this to 0.1% even before coupling tests. The audit's own machinery then adds a third answer: where the increment rule can only license or reject, the audit returns an indeterminate verdict together with the pilot that would settle it. It abstains on 809 of 1,800 contextual logs and licenses 32 rather than 72 of the 480 true Tier-3 cells (Table 7), so at these cluster counts it trades recall for protection. Two practical conclusions follow: ask the increment question first, then test whether the log can support it.

This paper offers four practical lessons for decision-makers. First, a certificate is not a licence: existence, magnitude, and causality differ, and an interval excluding zero shows existence only. Second, a saturation ladder flags confounding only if its enrichment covariates predict the action. On Olist the ladder, read on the absolute coupling, does not decline on either window, and the enrichment covariates do not predict the action, so the audit records C?, not 0 (C); read on the normalised scale, a confounder can move the ladder in the wrong direction (Appendix D.11). With limited power and without the propensity condition, calling confounding would be the overreach the tool is meant to prevent. Third, with 27 clusters, cluster-level inference is binding; row-level $n$ cannot fix it. The audit reports the action block's design effect $\bar{\delta}$ with a jackknife band per contrast, alongside $(\bar{m}, G)$. The choice of deflator matters: the mean-based Kish factor differs from $\bar{\delta}$ by a factor of three to four on these logs, in opposite directions on the two corpora. So does the reference distribution: with 27 uneven clusters, the action block has 3–9 degrees of freedom, not the thousands implied by independent rows, and calibrating the thresholds on the wild bootstrap itself changes the reason code, never the tier, in three of six logistics cells (Table S8). Fourth, shift the key check to budget approval: before funding a prescriptive programme, pre-commit to measuring whether the reward responds to the decision, and by enough to justify the build cost. The test is cheap, runs before any pipeline is built, returns a support class and an estimated response, and, where needed, specifies the randomised experiment that could change the verdict.

The results sharpen three themes. They make the optimizer's curse [20] measurable pre-investment via a spurious-gain floor $\sigma_\epsilon m_K$ and ceiling $c_\rho \sqrt{D}$, so that decision-makers can put a price on a spurious gain. They recast the minimax testing detection–estimation gap [52] as distinct detectability and learnability frontiers in policy learning, and show the gap can invert: when $\hat{\kappa}' < \lambda$, the learnable threshold can fall below the detectable one, leaving no band between. Finally, they operationalise design-science evaluation [21] by requiring artefacts to report operating characteristics as part of the contribution.

These claims have several limitations. (i) Neither logistics log has a true positive verdict; the only positive result is from a randomised recommendation log. For logistics, licensing is assessed via the plasmode in Table S16, which preserves weights, clusters, and contrasts and implants known coupling, but we do not audit any logistics log with a real positive verdict. The randomised audit is exploratory (state, clustering unit, split, learner chosen post hoc); confirmatory evidence would require a registered protocol and a second randomised audit, which we have not carried out. Both Olist windows are exploratory, because the pipeline was revised after the held-out results had been seen. (ii) The learnability frontier is a calibrated, learner- and data-dependent conjecture and does not transfer across designs; certification requires evaluating the deployed learner on the stated design. Verdicts are specific to a (dataset, reward, reference-design) triple (e.g., on-time vs. margin). (iii) The

dollar ceiling is partial-equilibrium. With congestion, capacity constraints, or tier repricing, the bias direction is unclear; refusals assume the action mix remains near the logged distribution, and with positive interference a policy can add value even if order-level coupling is zero. (iv) Downstream responsiveness (C6) is determined by the continuation coupling $\mathcal{D}_{\text{cont}}$ of Proposition 1, not by the sequential share $\chi_{\text{seq}}$ (which can be negative when $\mathcal{D}_{\text{cont}} > 0$ and is only a two-sided diagnostic). Neither estimator nor diagnostic is field-tested because all corpora are single-step; the [prov] flag is never exercised, so C6's reliability rests on inherited fitted-$Q$ evaluation. (v) The negative optimism constant is measured for one learner family without derivation; it is descriptive and conclusions do not depend on its sign. (vi) Field evidence uses three public logs; controlled claims rely on simulation and plasmodes from two real designs, so we cannot infer prevalence in practice. (vii) The logistics corpora have few clusters (23 and 27), which limits cluster-robust Wald tests to $df_1 \le G - 1$, makes pairs cluster-bootstrap intervals unreliable [58], and biases the percentile lower bound for $\hat{\omega}^2$ upward; detectability therefore uses a boundary test (Appendix A.5(v)). In practice, about 25 clusters may support a non-contextual rule but not a contextual one, regardless of row count. (viii) Two non-modelling conventions move thresholds more than any estimate here: using the action block's $\bar{\delta}$ instead of the mean-based Kish factor lowers Olist's threshold about 4× and raises DataCo's about 3×; and the reference distribution matters as much, since the CR2 correction implies thresholds far above any attainable coupling, which is why the thresholds are calibrated on the wild bootstrap itself (Appendix D.10). The calibrated and independent-rows footings agree on three of six logistics cells and differ only in the reason code of the other three (Table S8). At these cluster counts the coupling–threshold gap is uncertain to a factor of a few, and $\bar{\delta}$ is itself estimated from 23–27 clusters.

## Data and code availability

The code is available at `https://github.com/tungtokyo1108/LogWorthy`.

## Declaration of generative AI and AI-assisted technologies in the writing process

During the preparation of this work the authors used Claude Opus to improve the readability and language of the manuscript. After using this tool, the authors reviewed and edited the content as needed and take full responsibility for the content of the publication.

## Declaration of competing interest

The authors declare that they have no known competing financial interests or personal relationships that could have appeared to influence the work reported in this paper.

## CRediT authorship contribution statement

**Tung Dang:** Conceptualization, Methodology, Software, Validation, Formal analysis, Visualization, Writing – original draft, Writing – review & editing. **The Hung Phung:** Conceptualization, Methodology, Investigation, Supervision, Writing – review & editing. **Son Lam Nguyen:** Conceptualization, Methodology, Investigation, Supervision, Writing – review & editing. **Tu Nguyen:** Conceptualization, Methodology, Investigation, Supervision, Writing – review & editing.

# Appendix A. Supporting material for Section 3

This appendix contains the derivations, conventions and algorithms behind Section 3. Throughout, $w = \rho(a)/\pi_b(a \mid s)$, $\mathcal{D} = \mathbb{E}_s \operatorname{Var}_{a\sim\rho}[Q(s,a)]$, and $\epsilon_0$ is the overlap floor.

## Appendix A.1. Ignorability

Assumption A1 is used at one point only: to identify $Q(s,a)$ with the observational regression $Q_{\text{obs}}(s,a) = \mathbb{E}[Y \mid s,a]$, and hence to identify $\mathcal{D}$. It plays no role in Theorem 1.

What matters operationally is the direction in which a violation acts. Let $U$ be an unrecorded variable and suppose latent ignorability, $\{Y(a)\} \perp a \mid (s,U)$. This is the standard sensitivity device, an assumption rather than a consequence of A1 failing; without it the displays below are inequalities of unknown sign. Write $Q(s,u,a) = \mathbb{E}[Y \mid s, U = u, a]$ and $r_a(u \mid s) = p(u \mid s,a)/p(u \mid s)$, a density ratio with unit mean under $p(u \mid s)$. Then $Q_{\text{obs}}(s,a) = \mathbb{E}_{u|s}[r_a(u \mid s)Q(s,u,a)]$ while $Q(s,a) = \mathbb{E}_{u|s}[Q(s,u,a)]$: the observational regression averages $U$ over its law given the action taken, the causal one over its law given the state alone. Let $\beta(s,a) = Q_{\text{obs}}(s,a) - Q(s,a)$.

If $\beta(s,\cdot)$ varies with $a$ in the same direction as $Q(s,\cdot)$, the measured coupling is inflated. This is the case the saturation ladder screens, because enriching the state moves $r_a$ toward one. If $\beta(s,\cdot)$ varies against $Q(s,\cdot)$, the measured coupling is suppressed, and at $\beta(s,a) = -Q(s,a) + c(s)$ it vanishes. No ladder detects the second case, so every negative verdict carries the sensitivity analysis of Appendix A.7.

## Appendix A.2. Overlap and identification of $\mathcal{D}$

For any $\rho$-integrable $g$, Assumption A2 gives the Horvitz–Thompson identity [59]

$$\mathbb{E}_\rho[g(Y)] = \mathbb{E}\left[\frac{\rho(a)}{\pi_b(a \mid s)}\, g(Y)\right] = \mathbb{E}[w\, g(Y)], \tag{15}$$

so $\operatorname{Var}_\rho(Y) = \mathbb{E}[wY^2] - (\mathbb{E}[wY])^2$, and every component of (2) is a reweighted moment. Identification of $\mathcal{D}$ therefore requires $\pi_b(a \mid s) > 0$ for every $a$ with $\rho(a) > 0$ at every state with mass. The requirement is pointwise, which is why the audit screens it with the local statistic C4a rather than with the mean functional $H$ (§3.5). Where it fails on a region of positive mass, the reference variance on that region is not a functional of the observed distribution and the audit returns 0 (I).

## Appendix A.3. SUTVA and partial equilibrium

Assumption A4 makes $Y_i(a_i)$, and thus $Q$, well defined. With interference, the outcome of unit $i$ depends on the actions of others, $\mathbf{a}_{-i}$, so the ceiling must be indexed by the action profile, $Q^{\mathbf{a}_{-i}}(s,a)$, and Theorem 1 then bounds the improvement only with the aggregate action mix held at its logged value.

Theory does not fix the sign of the resulting error. With negative interference (capacity, congestion, tier breaches), shifting volume onto a mode worsens outcomes for orders already using it, and the ceiling is optimistic. With positive interference (consolidation economies, volume discounts that reprice the whole book), realised improvement can exceed the ceiling and can be positive even when $\mathcal{D} = 0$, because value then comes from changing the aggregate flow rather than individual choices. The instrument measures neither effect, so a refusal holds for policies that leave the aggregate action mix approximately unchanged, and Corollary 1 is stated under A4 wherever it is used to justify a refusal.

*Remark* 4 (Coarsening weakly reduces measured coupling). Let $B$ coarsen the action set and $Q_B(s,b) = \mathbb{E}_{a\sim\rho}[Q(s,a) \mid a \in b]$. Then $\mathcal{D}_B \le \mathcal{D}$ (Appendix A.8).

Binning acts in two ways, and only one is one-sided. The ceiling is monotone: bins are unions of actions, so $\rho_{B,\min} \ge \rho_{\min}$ and $c_{\rho_B} \le c_\rho$, and with $\mathcal{D}_B \le \mathcal{D}$ this gives $c_{\rho_B}\sqrt{\mathcal{D}_B} \le c_\rho\sqrt{\mathcal{D}}$; coarsening cannot inflate the ceiling. The verdict is not monotone: coarsening also lowers the threshold, through a smaller $df_1$ and a weakly larger $n_{\text{eff}}$ (Appendix A.8). Binning too coarsely or too finely can each push a log below detectability, so the binning rule is part of the written plan.

## Appendix A.4. The moment condition of Assumption A5

The coupling statistic uses reweighted second moments $\hat{m}_2 = n^{-1}\sum_i w_i y_i^2$. A central-limit approximation needs only a finite summand variance,

$$\operatorname{Var}(wY^2) < \infty \iff \mathbb{E}\big[w^2Y^4\big] < \infty, \tag{16}$$

which is weaker than $\mathbb{E}[w^4Y^4] < \infty$ and matters because fulfilment margins and near-floor weights are heavy-tailed. If $Y$ is bounded and $\epsilon_0 > 0$, both conditions hold since $w \le \rho_{\max}/\epsilon_0$.

Two further requirements are invoked where needed. The ratio $\hat{\eta}^2$ needs a joint central-limit theorem for numerator and denominator, which follows from (16) applied to each term and a denominator bounded away from zero (Appendix A.5(iii)). If $\hat{\pi}_b$ is estimated, the weights are random; this uncertainty is propagated by refitting $\hat{\pi}_b$ within resamples (Appendix A.5(iv)). When weights are extremely heavy-tailed, the audit reports $\max_i w_i$, $\widehat{\mathbb{E}}[w^2]$ and the weight share of the top one per cent of rows, and switches to self-normalised or trimmed weights [14] at the trimming level fixed in the plan.

## Appendix A.5. Estimation conventions

The following conventions apply to every reported coupling.

*(i) The estimator and its degrees of freedom.* Let $\mathrm{SS}_\Delta = \widehat{\mathrm{RSS}}_{\mathcal{M}_0} - \widehat{\mathrm{RSS}}_{\mathcal{M}_1}$, $\mathrm{MSE} = \widehat{\mathrm{RSS}}_{\mathcal{M}_1}/\nu$ with $\nu = n_{\text{info}} - \dim(\mathcal{M}_1)$, and $\mathrm{SS}_T$ the weighted total sum of squares. Then

$$\hat{\omega}^2 = \frac{\mathrm{SS}_\Delta - df_1 \cdot \mathrm{MSE}}{\mathrm{SS}_T + \mathrm{MSE}}, \qquad \widehat{\eta^2_{a|s}} = \frac{\mathrm{SS}_\Delta - df_1 \cdot \mathrm{MSE}}{\mathrm{SS}_\Delta + \nu \cdot \mathrm{MSE}},$$

(17)

and the absolute estimand is $\widehat{D} = (\mathrm{SS}_\Delta - df_1 \cdot \mathrm{MSE})/\sum_i w_i$, so that estimator and test measure the same object, interaction included. The $df_1 \cdot$ MSE term removes the sum of squares that fitting $df_1$ parameters produces under the null. The correction is unbiased for balanced, unweighted analysis of variance [60] and is not known to be unbiased under importance weights and clustering, so its residual bias is measured by simulation on each log's own weights and clusters and reported against the interval half-width (Table S17).

*(ii) Negative estimates.* Because the correction subtracts estimated noise, the numerator can be negative. The point estimate is reported as $\max(\hat{\omega}^2, 0)$, but intervals are built on the untruncated scale, and an interval that touches or crosses zero is reported as such.

*(iii) Non-positive denominators.* $\widehat{\mathrm{Var}}_\rho(Y) = \widehat{\mathbb{E}}[wY^2] - (\widehat{\mathbb{E}}[wY])^2$ can be non-positive in small samples with extreme weights. Such cases are marked unestimable, routed to 0 (I) and counted, never treated as large coupling.

*(iv) Resampling.* Resampling is at the cluster level. With $G \geq 50$ clusters the interval is a pairs cluster bootstrap that refits $\hat{\pi}_b$ in each resample (Algorithm 1). With fewer clusters the pairs bootstrap is unreliable [58], and the interval is a wild cluster bootstrap with Rademacher weights (Webb weights when $G < 12$), with propensities held at their fitted values; a pairs bootstrap with refitted propensities is then reported as a sensitivity row. Randomised logs with known propensities use the randomisation distribution instead (Algorithm 2).

*(v) Boundary tests.* For an $R^2$-type statistic with few clusters, the percentile lower limit is biased upward, because the bootstrap world treats the fitted action block, noise included, as the truth. The *robustly detectable* branch of (12) is therefore decided by inverting a test at the boundary: rescale the fitted action block so that the implied coupling equals $\eta^2_{\mathrm{min,hi}}$, resample around this null on the footing of (iv), and take the branch if the observed $\hat{\omega}^2$ lies in the upper $\alpha$ tail. The mirror test at $\eta^2_{\mathrm{min,lo}}$, lower tail, confirms the *below* branch. The size and power of the boundary test are reported in the self-check table on each log's design.

*(vi) Calibrated threshold.* The calibrated threshold (9) is estimated by Algorithm 2. For randomised logs it replaces (8); for observational logs both are reported, and a disagreement larger than the Monte-Carlo error is itself a finding about the information count.

**Algorithm 1** Cluster bootstrap for the coupling interval

```
input: rows (s_i, a_i, y_i) with cluster labels c_i; reference ρ;
  R replicates; level α; the specification fixed in the plan
for r = 1 to R do
  resample clusters with replacement (pairs) or draw
  cluster-level weights (wild), per Appendix A.5(iv)
  refit π̂_b^(r) when the pairs scheme is used; form w^(r) =
  ρ(a)/π̂_b^(r)(a | s) and apply the trimming rule
  fit M_0 and M_1; compute ω̂^2(r) by (17)
end for
return max(ω̂^2, 0), (df_1, ν), the percentile interval of
  the untruncated {ω̂^2(r)}, and δ̄ with its jackknife interval
```

**Algorithm 2** Calibrated detectability threshold

```
input: the log; the contrast; level α; power 1 − β; a grid
  0 < η_1^2 < ⋯ < η_J^2; R replicates per grid point
for j = 1 to J do
  for r = 1 to R do
    generate a null dataset: randomised log, permute
    actions within strata; observational log, fit M_0
    and resample residuals on the footing of Appendix A.5(iv)
    add an action effect, drawn as a rescaled copy of the
    fitted action block of M_1, whose implied partial
    coupling is η_j^2
    run the audit's test exactly as specified and record
    whether it rejects
  end for
  estimate the power pŵ_j with its binomial interval
end for
return the smallest η_j^2 with pŵ_j ≥ 1 − β, interpolated
  between grid points, and its Monte-Carlo interval
```

## Appendix A.6. The design effect equals the normalised effective overlap

Let $\rho$ be uniform. Then $\mathbb{E}[w] = \mathbb{E}_s \sum_a \pi_b(a \mid s) \cdot \frac{1/K}{\pi_b(a|s)} = 1$ and

$$\begin{aligned}\mathbb{E}[w^2] &= \mathbb{E}_s \sum_a \pi_b(a \mid s) \frac{1/K^2}{\pi_b(a \mid s)^2} \\ &= \frac{1}{K^2}\, \mathbb{E}_s \sum_a \frac{1}{\pi_b(a \mid s)} = \frac{H}{K},\end{aligned} \tag{18}$$

with $H$ as in Definition 2. The Kish design effect [43, 44] therefore gives

$$n_{\mathrm{eff}} = \frac{\left(\sum_i w_i\right)^2}{\sum_i w_i^2} \longrightarrow \frac{n\,(\mathbb{E}[w])^2}{\mathbb{E}[w^2]} = \frac{nK}{H} = n\, o_{\mathrm{eff}}. \tag{19}$$

Under a uniform reference design the effective sample size of the coupling estimator and the effective overlap coincide, so Stage 0 reports one number. For a non-uniform $\rho$, $\mathbb{E}[w^2] = \mathbb{E}_s \sum_a \rho(a)^2/\pi_b(a \mid s)$, which is not $H/K$, and the two are reported separately. For a candidate policy $\pi$ the value-estimation functional is $H(\pi)$, which depends on $\pi$ and is recomputed at Stage 2.

## Appendix A.7. Sensitivity analysis for a negative verdict

The body attaches a sensitivity figure to each negative verdict. It is derived here under the latent-ignorability device of Appendix A.1.

*A checkable bound.* Fix $s$ and treat $Q(s,\cdot)$, $Q_{\text{obs}}(s,\cdot)$ and $\beta(s,\cdot) = Q_{\text{obs}} - Q$ as elements of $L^2(\rho)$ with seminorm $\|f\|^2 = \text{Var}_{a\sim\rho}(f)$. Let $\mathcal{D}_{\text{true}} = \mathbb{E}_s\|Q(s,\cdot)\|^2$, $\mathcal{D}_{\text{obs}} = \mathbb{E}_s\|Q_{\text{obs}}(s,\cdot)\|^2$ and $\mathcal{D}_{\text{bias}} = \mathbb{E}_s\|\beta(s,\cdot)\|^2$. Minkowski's inequality gives

$$\sqrt{\mathcal{D}_{\text{true}}} \;\le\; \sqrt{\mathcal{D}_{\text{obs}}} + \sqrt{\mathcal{D}_{\text{bias}}}. \tag{20}$$

A true coupling of size $\mathcal{D}^\star$ is therefore compatible with an observed coupling whose one-sided upper limit is $U$ only if

$$\sqrt{\mathcal{D}_{\text{bias}}} \;\ge\; \sqrt{\mathcal{D}^\star} - \sqrt{U}. \tag{21}$$

The bound is tight: the masking construction of Appendix A.7.1, with $\beta = -Q$, gives $\mathcal{D}_{\text{obs}} = 0$ and equality.

*From required bias to confounding strength.* Under latent ignorability,

$$|\beta(s,a)| \;\le\; \sigma_Q(s,a)\cdot\sqrt{\chi^2\big(p(\cdot\mid s,a)\,\|\,p(\cdot\mid s)\big)}, \tag{22}$$

where $\sigma_Q^2(s,a) = \text{Var}_{U|s}\big(Q(s,U,a)\big)$ is the outcome variation due to the unrecorded variable and the $\chi^2$ divergence measures how strongly action assignment depends on it. Parameterising the two by $R^2_{Y\sim U|s,a}$ and $R^2_{a\sim U|s}$ connects the bound to omitted-variable sensitivity analysis [25]; combining (21) and (22) gives the frontier of $(R^2_{Y\sim U}, R^2_{a\sim U})$ pairs that can mask a coupling of size $\mathcal{D}^\star$.

*What the audit reports.* With $\mathcal{D}^\star$ the break-even coupling of Appendix C.1, the robustness value RV is the smallest $r$ such that a confounder with $R^2_{Y\sim U|s,a} = R^2_{a\sim U|s} = r$ can mask a coupling worth building. It is reported with every negative verdict, alongside the same quantity for a benchmark confounder. When no cost data exist, break-even is undefined and so is RV; the audit then reports the frontier at the detectability threshold instead and says so.

*Why this motivates randomisation.* Randomising a fraction of decisions sets $r_a(u \mid s) \equiv 1$ on that subsample, so $\beta \equiv 0$ there and masking is broken by design. A low RV therefore points to the exploration pilot of §5.2, and (21) sizes it. Because randomisation removes masking but not interaction, the pilot is sized with the omnibus calculation at its own $df_1$, not with a one-way calculation.

### Appendix A.7.1. What a negative verdict establishes

Let $a$ be binary with uniform $\rho$, $U \sim \text{Bern}(1/2)$ unobserved, $Y(1) = U$ and $Y(2) = 1 - U$, so the action fully determines the outcome. A dispatcher chooses action 1 with probability $p$ when $U = 1$ and $1-p$ when $U = 0$, for some $p \in (0,1)$, so overlap holds. Then $\Pr(U = 1 \mid a = 1) = \Pr(U = 0 \mid a = 2) = p$ and $\mathbb{E}[Y \mid a = 1] = \mathbb{E}[Y \mid a = 2] = p$: the measured coupling is zero for every $p$, while with $U$ observed $\mathcal{D} = 1/4 = \text{Var}_\rho(Y)$ and $\eta_a^2 = 1$. Dispatchers who act on undocumented knowledge produce exactly this structure. A Tier 0 verdict therefore means no detectable coupling under the recorded state, qualified by its robustness value.

The same bound addresses the mirror concern after a positive verdict whose ladder has not stabilised. If $\beta(s,\cdot)$ varies with $a$ in the direction of $Q(s,\cdot)$, the measured coupling is inflated and a certified gain may be only observational. Since $\sqrt{\mathcal{D}_{\text{obs}}} \le \sqrt{\mathcal{D}_{\text{true}}} + \sqrt{\mathcal{D}_{\text{bias}}}$, the confounding needed to manufacture the certified coupling follows from the same calculation with the certification level in place of break-even. The audit reports this positive-side robustness value with every observational Tier 2 (C?) verdict, on the same scale and with the same benchmark confounder, before any shadow deployment is authorised.

## Appendix A.8. Coarsening weakly reduces measured coupling

*Proof of Remark 4.* Let $B$ be a partition of the action set, $\rho_B(b) = \sum_{a\in b}\rho(a)$ the induced reference design and $Q_B(s,b) = \mathbb{E}_{a\sim\rho}[Q(s,a) \mid a \in b]$. Fix $s$ and apply the law of total variance under $\rho$, conditioning on the bin:

$$\begin{aligned}\text{Var}_{a\sim\rho}\left[Q(s,a)\right] &= \text{Var}_{b\sim\rho_B}\big(Q_B(s,b)\big)\\ &\quad+\mathbb{E}_{b\sim\rho_B}\big[\text{Var}_{a\sim\rho}(Q(s,a)\mid a\in b)\big].\end{aligned} \tag{23}$$

The second term is non-negative, so $\text{Var}_b(Q_B(s,b)) \le \text{Var}_a(Q(s,a))$ at every $s$, and taking $\mathbb{E}_s$ gives $\mathcal{D}_B \le \mathcal{D}$. □

Each bin is a union of actions, so $\rho_{B,\min} \ge \rho_{\min}$; since $c_\rho$ decreases in $\rho_{\min}$, $c_{\rho_B} \le c_\rho$ and

$$c_{\rho_B}\sqrt{\mathcal{D}_B} \;\le\; c_\rho\sqrt{\mathcal{D}}. \tag{24}$$

No binning can manufacture an optimistic ceiling. The effective sample size moves the other way. For each bin, Cauchy–Schwarz in Engel form gives

$$\sum_{a\in b}\frac{\rho(a)^2}{\pi_b(a\mid s)} \;\ge\; \frac{\big(\sum_{a\in b}\rho(a)\big)^2}{\sum_{a\in b}\pi_b(a\mid s)} = \frac{\rho_B(b)^2}{\pi_B(b\mid s)}, \tag{25}$$

so $\mathbb{E}[w_B^2] \le \mathbb{E}[w^2]$ and, by (19), $n_{\text{eff}}$ weakly increases under coarsening. Together with the smaller $df_1$ this lowers the threshold while (23) lowers the measured coupling, so the net effect on the verdict is indeterminate. Too coarse a partition sinks the measured coupling; too fine a partition inflates $df_1$ and depresses $n_{\text{eff}}$. Either can push a log below detectability, which is why the binning rule is fixed in the plan and its sensitivity reported.

## Appendix A.9. Extended discussion of the feasibility conditions

*C1, non-definitional target.* The failure occurs when the outcome can be reconstructed from the inputs, an accounting identity, so that optimisation targets the definition rather than the world. The screen fits a flexible model of $y$ on $(s, a)$ and flags near-perfect out-of-sample $R^2$ or exact functional recovery. A flag triggers human review, not an automatic verdict, because high predictability need not mean definitionality.

*C2, post-decision leakage.* Manual column tagging is fallible, so an automatic rule marks every column timestamped after the decision as post-decision, and a training-time assertion blocks any model that uses one. The guarantee is then a property of the pipeline rather than of the analyst.

*C4a and C4b.* C4a reports the share of state mass at which some action with $\rho(a) > 0$ has $\hat{\pi}_b(a \mid s) < \delta_{\text{trim}}$; a non-negligible share routes the log to 0 (I). C4b reports $\widehat{H}$, $o_{\text{eff}}$ and $n_{\text{info}}$, computed for the uniform policy at Stage 0 and recomputed for the proposed policy at Stage 2.

*C5a and C5b.* All reweighting depends on $\hat{\pi}_b$, so Stage 0 requires expected calibration error below 0.05 and reports the calibration slope. C5b compares the held-out log-loss of $\hat{\pi}_b$ with that of the marginal action shares. A log at the marginal baseline is near-randomly assigned given the recorded state: admissible, often favourable, for estimation, but it must be identified, because the reuse error of Theorem 2 is largest near uniform logging.

*C6, downstream responsiveness.* For sequential objectives, $\mathcal{D}_{\text{cont}}$ of (6) is estimated by fitted $Q$-evaluation of $\rho$ [49] as the dispersion over actions of $\hat{Q}^{\rho} - \hat{R}$, with an interval from a bootstrap over trajectories and a threshold calibrated by Algorithm 2 applied to the continuation term. C6 fires when the lower limit clears the threshold. Its reliability is bounded by that of the fitted $Q$-evaluation, which requires the concentrability condition of Theorem 1(c). The diagnostic $\chi_{\text{seq}}$ of (7) is reported alongside with both signs interpreted (Proposition 1).

### Appendix A.10. Scope notes

*Remark* 5 (The marginal floor on $H$, and what attaining it means). Let $p_a = \mathbb{E}_s[\pi_b(a \mid s)]$ be the marginal action shares. Jensen's inequality gives $\mathbb{E}_s[\pi_b(a \mid s)^{-1}] \geq p_a^{-1}$, so

$$H \;\geq\; \frac{1}{K}\sum_a \frac{1}{p_a}, \qquad \text{equivalently} \qquad o_{\text{eff}} \;\leq\; \frac{K^2}{\sum_a p_a^{-1}},$$

with equality if and only if $\pi_b(\cdot \mid s)$ is almost everywhere constant in $s$. The bound is a free check on any reported $o_{\text{eff}}$. It is also a warning: a fitted $\hat{\pi}_b$ with $\widehat{H}$ at the floor has no state dependence, C5b fails, reweighting only adjusts the action mix, and the log should be treated as near-randomly assigned. For this reason $\widehat{H}$ is computed from the fitted model, $n^{-1}\sum_i K^{-1}\sum_a \hat{\pi}_b(a \mid s_i)^{-1}$, never from the marginal shares, and the audit reports the fitted value, the floor and their gap.

*Remark* 6 (The three standing scope conditions). Clustering: $n_{\text{eff}}$ corrects uneven action assignment, not within-cluster dependence, so the threshold uses the information count of §3.4. Reward-dependence: $\mathcal{D}$ depends on $Y$, so every verdict applies to a (dataset, reward, reference design, policy class) combination. Binning: a coarser partition lowers both the measured coupling and the threshold, with ambiguous net effect (Appendix A.8), so the rule is fixed in advance.

## Appendix B. Proofs for Section 4

Proofs are given in the order the results are stated.

### Appendix B.1. Proof of Theorem 1 and its corollaries

*Part (a).* Fix $s$, set $d_a = Q(s,a) - \bar{Q}_\rho(s)$, so that $\sum_a \rho(a) d_a = 0$ and $\sigma_\rho^2(s) = \sum_a \rho(a) d_a^2 = \operatorname{Var}_{a\sim\rho} Q(s,a)$, and let $m = \arg\max_a d_a$. Since $\rho_m d_m = -\sum_{a\neq m}\rho_a d_a$, Cauchy–Schwarz gives

$$(\rho_m d_m)^2 \leq \Big(\sum_{a\neq m}\rho_a\Big)\Big(\sum_{a\neq m}\rho_a d_a^2\Big) = (1-\rho_m)\big(\sigma_\rho^2(s) - \rho_m d_m^2\big),$$

hence $d_m \leq \sigma_\rho(s)\sqrt{(1-\rho_m)/\rho_m} \leq c_\rho\,\sigma_\rho(s)$. Taking $\mathbb{E}_s$ and applying Jensen's inequality to the concave square root gives

$$\mathbb{E}_s[\max_a Q - \bar{Q}_\rho] \leq c_\rho\,\mathbb{E}_s\sigma_\rho(s) \leq c_\rho\sqrt{\mathcal{D}}.$$

For uniform $\rho$, $\rho_m = 1/K$ gives $\sqrt{K-1}$. Equality holds when, at every state, the action with the smallest $\rho$-mass carries the whole advantage and the other actions share one value, with $\sigma_\rho(s)$ constant in $s$; the constant is therefore sharp.

*Part (b).* The same argument applied to $-Q(s,\cdot)$ gives $\bar{Q}_\rho(s) - \min_a Q(s,a) \leq c_\rho\sigma_\rho(s)$, so $\max_a Q - \min_a Q \leq 2c_\rho\sigma_\rho(s)$. Since $V(\pi_b) \geq \mathbb{E}_s[\min_a Q]$ and $V(\pi^\star) = \mathbb{E}_s[\max_a Q]$, taking $\mathbb{E}_s$ and Jensen's inequality yields the claim.

*Part (c).* The performance difference lemma [47] gives

$$V^{\pi^\star}(\mu) - V^{\rho}(\mu) = \frac{1}{1-\gamma}\,\mathbb{E}_{s\sim d_\mu^{\pi^\star}}\Big[\textstyle\sum_a \pi^\star(a \mid s) A^{\rho}(s,a)\Big], \qquad A^\rho := Q^\rho - V^\rho.$$

The inner sum is at most $\max_a A^\rho(s,a) = \max_a Q^\rho(s,a) - \mathbb{E}_{a\sim\rho}Q^\rho(s,a)$, which part (a) bounds by $c_\rho$ times the $\rho$-standard deviation of $Q^\rho(s,\cdot)$. Taking $\mathbb{E}_{s\sim d_\mu^{\pi^\star}}$ and applying Jensen's inequality gives the first inequality. For the second, $\mathbb{E}_{d_\mu^{\pi^\star}}[f] = \mathbb{E}_{d_\mu^\rho}[(d_\mu^{\pi^\star}/d_\mu^\rho) f] \leq C\,\mathbb{E}_{d_\mu^\rho}[f]$ for $f \geq 0$. At $\gamma = 0$, $d_\mu^\pi = \mu$ for every $\pi$, $Q^\rho = R$ and $C = 1$, and both bounds reduce to (a). We checked part (c) on 3,000 random MDPs (2–8 states, 2–4 actions, Dirichlet(0.5) transitions, $\gamma \in \{0.5, 0.8, 0.9, 0.95\}$, uniform or Dirichlet $\rho$): neither inequality was violated and the first was attained up to a ratio of 0.9999. □

*Corollary 1.* $\mathcal{D} = 0$ if and only if $\mathrm{Var}_{a\sim\rho} Q(s,a) = 0$ for almost every $s$, that is, $Q(s,\cdot)$ is constant on $\mathrm{supp}(\rho)$. Then $V(\pi) = \mathbb{E}_s \bar{Q}_\rho(s)$ for every $\pi$ supported in $\mathrm{supp}(\rho)$, including $\pi_b$, so $\Delta(\pi) = 0$. □

*Corollary 2.* For $\pi_a \equiv a$, $V(\pi_a) - V(\rho) = \mathbb{E}_s[Q(s,a) - \bar{Q}_\rho(s)] = M(a) - \bar{M}$. Applying the argument of part (a) to the single vector $(M(a))_a$ under $\rho$ gives $\max_a M(a) - \bar{M} \le c_\rho \sqrt{\mathrm{Var}_{a\sim\rho} M(a)} = c_\rho \sqrt{\mathcal{D}_{\mathrm{main}}}$, and $\mathcal{D}_{\mathrm{main}} \le \mathcal{D}$ by (3). The decomposition itself follows from $Q(s,a) - \bar{Q}_\rho(s) = (M(a) - \bar{M}) + I(s,a)$ and $\mathbb{E}_s I(s,a) = M(a) - \mathbb{E}_s \bar{Q}_\rho(s) - M(a) + \bar{M} = 0$, because $\mathbb{E}_s \bar{Q}_\rho(s) = \bar{M}$. □

*Proposition 1.* On functions of $(s,a)$ define the seminorm $\|f\|^2 = \mathbb{E}_{s\sim d_\mu^\rho} \mathrm{Var}_{a\sim\rho} f(s,a)$. It is the $L^2(d_\mu^\rho \otimes \rho)$ norm of $f$ after subtracting its $\rho$-mean at each state, so it satisfies the triangle inequality. With $Q^\rho = R + (Q^\rho - R)$ we have $\mathcal{D}_{\mathrm{seq}} = \|Q^\rho\|^2$, $\mathcal{D}_1 = \|R\|^2$ and $\mathcal{D}_{\mathrm{cont}} = \|Q^\rho - R\|^2$, which gives both inequalities, and expanding the square gives the identity, since $Q^\rho - R = \gamma P V^\rho$ by the Bellman equation of $\rho$. Finally $\mathcal{D}_{\mathrm{cont}} = 0$ if and only if $PV^\rho(s,\cdot)$ is constant on $\mathrm{supp}(\rho)$ for $d_\mu^\rho$-almost every $s$. □

*Remark* 7 (Why the constant depends on the reference design). The Laguerre–Samuelson constant $\sqrt{K-1}$ belongs to an unweighted mean and variance. For $K = 2$ and $\rho = (0.1, 0.9)$ the sharp constant is $c_\rho = 3$, so $\sqrt{K-1} = 1$ would understate the attainable improvement threefold. Because §3.1 allows a business-chosen reference design, the general constant is needed; the uniform design remains the default and gives the smallest constant.

*Remark* 8 (On the constant, and on reporting $\mathcal{D}$). The ceiling depends on the absolute $\mathcal{D}$, not on $\mathrm{Var}_\rho(Y)$, while the normalised couplings depend on $\mathrm{Var}_\rho(Y)$, which under state-dependent logging differs from the observed $\mathrm{Var}(Y)$. The audit therefore reports $\hat{\mathcal{D}}$, $\widehat{\mathrm{Var}}_\rho(Y)$ and the noise term, from which $\eta_a^2$, $\eta_{a|s}^2$ and the money ceiling follow, together with the empirical tightness ratio $\hat{c}_{\mathrm{emp}} \in (0,1]$ of any policy value that can be estimated (Remark 1).

*Remark* 9 (Heterogeneous coupling). The ceiling controls the average improvement through the average coupling. Strong coupling in a segment (a lane, a product family) can be diluted in the aggregate, so the audit is applied by pre-specified segment as well as overall, under the multiplicity control of Appendix C.2.

## Appendix B.2. The overlap floor does not determine $H$

Take $K = 2$ and $\pi_b$ constant within state groups. A log playing $(0.1, 0.9)$ everywhere and one playing $(0.1, 0.9)$ on half the states and $(0.5, 0.5)$ on the rest share the floor $\epsilon_0 = 0.1$ but have $H = 5.56$ and $3.78$, so the floor is not sufficient. A log playing $(0.2, 0.8)$ everywhere and a mixture with weight 0.3 on $(0.1, 0.9)$ and 0.7 on $(0.4, 0.6)$ have floors 0.2 and 0.1 but the same $H = 3.125$, so the floor is not necessary.

*Remark* 10 (Correlated and unequal fitting errors). Proposition 2 assumes errors independent across actions with a common variance. Shared representations induce positive correlation, which for equal variances lowers $\mathbb{E}[\max_a \epsilon]$ [61]; the independent case is then an upper bound on the spurious gain. With unequal variances $\sigma_a$ no single $\sigma_\epsilon$ summarises the gain; for $K = 2$ it equals $\sqrt{(\sigma_1^2 + \sigma_2^2)/(2\pi)}$ and is driven by the noisier estimate, which belongs to the rarer action. The greedy rule then selects rare actions disproportionately. Fulfilment action shares are highly unequal, so per-action fit quality is unequal by construction, and the audit reports per-action residual variances.

## Appendix B.3. Proof of Proposition 3

Following Cohen [51], in the Gaussian linear model with $n_{\mathrm{info}}$ independent rows the $F$ statistic for $\mathcal{M}_0$ against $\mathcal{M}_1$ is noncentral $F_{df_1,\nu}$ with noncentrality $\lambda = n_{\mathrm{info}} \eta_{a|s}^2 / (1 - \eta_{a|s}^2)$. Power is strictly increasing in $\lambda$, so the noncentrality $\lambda_{\alpha,\beta,df_1}$ at which power equals $1 - \beta$ is unique, and solving for $\eta_{a|s}^2$ gives (8). For a one-way contrast, $df_1 = K - 1$ and $\lambda_{0.05,0.20,df_1} = 7.85$, $10.90$ and $14.35$ for $K = 2$, 4 and 8. For large $df_1$,

$$\lambda_{\alpha,\beta,df_1} \approx (z_{1-\alpha} + z_{1-\beta})\sqrt{2\,df_1},$$
$$\eta_{\min}^2 \approx \frac{(z_{1-\alpha} + z_{1-\beta})\sqrt{2\,df_1}}{n_{\mathrm{info}}}, \tag{26}$$

so the threshold scales as $\sqrt{df_1}/n_{\mathrm{info}}$: the omnibus test gains power against interactive alternatives and loses it against a pure main effect. When a main effect is the pre-specified target, the main-effect contrast is reported alongside the omnibus contrast as one multiplicity-controlled family (Appendix C.2).

*Which $\lambda$.* Every $\lambda_{\alpha,\beta,df_1}$ used here is the exact solution of $\Pr(F_{df_1,\nu,\lambda} > F_{1-\alpha;df_1,\nu}) = 1 - \beta$. Equation (26) is illustrative only: over the range used here it understates $\lambda$ (Table S1), and since $\eta_{\min}^2$ increases in $\lambda$ the approximation makes detection look easier, the unsafe direction.

## Appendix B.4. The in-sample estimator and its standard error

**Lemma 1** (State-level limits under reuse)**.** *In the setting of Theorem 2, fix a state s, let $\hat{p}_s$ be the share of rows with $a = 1$ and $X_s = \hat{Q}(s,1) - \hat{Q}(s,0)$, and let $\hat{\Delta}_s$ and $\hat{v}_s$ be the mean and the variance of $\phi_i$ over the rows of s. Then:*

(a) *$\hat{\Delta}_s = (1 - \hat{p}_s) X_s$ if $X_s > 0$ and $\hat{\Delta}_s = -\hat{p}_s X_s$ otherwise; in particular $\hat{\Delta}_s \ge 0$ and the inverse-propensity term contributes nothing to it.*

(b) *As $n_s \to \infty$, $\sqrt{n_s}\,\hat{\Delta}_s / \sigma \Rightarrow c_s |Z_s|$ and $\hat{v}_s \to \sigma^2 c_s^2$ in probability, with $Z_s$ standard normal and $c_s$ as in (11).*

(c) *Consequently $n\,\mathbb{E}[\widehat{\mathrm{SE}}^2] \to \sigma^2 \mathbb{E}[c^2] = \sigma^2 (H - 1)$ and $n\,\mathrm{Var}(\hat{\Delta}) \to \sigma^2 \mathrm{Var}(c|Z|) = \sigma^2 (H - 1 - H/\pi)$.*

*Proof.* (a) Within each cell $(s,a)$ the residuals $y_i - \hat{Q}(s,a)$ sum to zero because $\hat{Q}$ is the cell mean, so the inverse-propensity term, which sums residuals of the selected cell only, vanishes. What remains is $\hat{Q}(s,\hat{\pi}(s)) - \bar{y}_s$ with $\bar{y}_s = \hat{p}_s\hat{Q}(s,1) + (1-\hat{p}_s)\hat{Q}(s,0)$, which equals the stated expressions.

**Table S1**
Exact noncentrality $\lambda_{\alpha,\beta,df_1}$ at $\alpha = 0.05$, $1-\beta = 0.80$ and large $\nu$, against the large-$df_1$ approximation (26). The first three rows are the one-way values for $K = 2, 4, 8$. The approximation is shown for its scaling and is never used to set a threshold.

| $df_1$ | exact $\lambda$ | Eq. (26) | shortfall |
|---|---|---|---|
| 1 | 7.85 | 3.52 | 55% |
| 3 | 10.90 | 6.09 | 44% |
| 7 | 14.35 | 9.30 | 35% |
| 14 | 18.34 | 13.16 | 28% |
| 21 | 21.36 | 16.11 | 25% |
| 42 | 28.12 | 22.79 | 19% |

(b) $\hat{p}_s \to p$ and, by the central limit theorem for the two cell means, $\sqrt{n_s}X_s/\sigma \Rightarrow \mathcal{N}(0, 1/(p(1-p)))$. Write $Z_s = \sqrt{p(1-p)}\sqrt{n_s}X_s/\sigma$. If $Z_s > 0$ the selected action is 1 and $\sqrt{n_s}\hat{\Delta}_s/\sigma \to (1-p)|Z_s|/\sqrt{p(1-p)} = \sqrt{(1-p)/p}\,|Z_s|$; otherwise it is $p|Z_s|/\sqrt{p(1-p)} = \sqrt{p/(1-p)}\,|Z_s|$. For the variance, suppose action 1 is selected. Rows with $a = 1$, a share $p$, have $\phi_i = (1/p - 1)(y_i - \hat{Q}(s,1))$ and contribute $p \cdot ((1-p)/p)^2\sigma^2$; rows with $a = 0$ have $\phi_i = X_s - (y_i - \hat{Q}(s,0))$ and contribute $(1-p)\sigma^2$, since $X_s \to 0$. The sum is $\sigma^2(1-p)/p = \sigma^2 c_s^2$. The case of action 0 is symmetric and gives $\sigma^2 p/(1-p)$. Since $\hat{\Delta}_s \to 0$, centring does not change the limit.

(c) With $a = \sqrt{(1-p)/p}$, $c$ takes the values $a$ and $1/a$ with probability $\frac{1}{2}$ each, and $a + 1/a = 1/\sqrt{p(1-p)} = \sqrt{2H}$. Hence $\mathbb{E}c = \sqrt{H/2}$, $\mathbb{E}c^2 = \big((a+1/a)^2 - 2\big)/2 = H-1$, $\mathbb{E}[c|Z|] = \sqrt{H/2}\sqrt{2/\pi} = \sqrt{H/\pi}$ and $\mathrm{Var}(c|Z|) = H - 1 - H/\pi$. The statements follow from (b), from $\sqrt{n}\,\hat{\Delta} = S^{-1/2}\sum_s \sqrt{n_s}\,\hat{\Delta}_s$ with independent states, and from $n\widehat{\mathrm{SE}}^2 = S^{-1}\sum_s \hat{v}_s + o_p(1)$. □

At $p = \frac{1}{2}$ the influence-function variance is $\sigma^2$ while the sampling variance of $\sqrt{n}\,\hat{\Delta}$ is $0.363\sigma^2$; at $p = 0.05$ they are $9.53\sigma^2$ and $6.18\sigma^2$. Simulation with $S = 10$, $n_s = 500$ and 1,500 replications gives $0.999\sigma^2$ and $0.364\sigma^2$ at $p = \frac{1}{2}$, and $9.19\sigma^2$ and $6.74\sigma^2$ at $p = 0.05$. The ratio $\kappa = \mathrm{sd}(\hat{\Delta})/\mathbb{E}[\widehat{\mathrm{SE}}]$ is therefore well below one under reuse, 0.60 at uniform logging and 0.81 at $p = 0.05$; this is the overstatement behind Remark 2.

## Appendix B.5. Transporting a calibrated learnability threshold

Because Conjecture 4.3 makes the threshold depend on $n_{\mathrm{eff}}$, a certification threshold measured on one design does not apply unchanged to another. If $\eta^2_{\mathrm{cert}}$ is the coupling at which a learner first certifies an honest-split improvement on a calibration design of effective size $n^{\mathrm{cal}}_{\mathrm{eff}}$, the fitted constant is $\hat{\kappa}' = \eta^2_{\mathrm{cert}}\, n^{\mathrm{cal}}_{\mathrm{eff}}$ and the transported threshold is

$$\eta^2_{\mathrm{learn}}(n_{\mathrm{eff}}) = \eta^2_{\mathrm{cert}} \cdot \frac{n^{\mathrm{cal}}_{\mathrm{eff}}}{n_{\mathrm{eff}}}.$$

Two rules follow: $n^{\mathrm{cal}}_{\mathrm{eff}}$ is reported with every $\eta^2_{\mathrm{cert}}$, and the rescaling is applied before any comparison. Whether $\hat{\kappa}'$ is constant in $n_{\mathrm{eff}}$ is the conjecture's one testable content. The quantity to plot is $\eta^2_{\mathrm{cert}}\, n_{\mathrm{eff}}$ across at least five sample sizes covering the target logs; if it is not flat, the rescaling is not licensed and the threshold must be measured at the target's own size.

## Appendix B.6. Proof of Theorem 2

*The limit* (11). States are independent, $n = Sn_s$, and $\hat{\Delta} = S^{-1}\sum_s \hat{\Delta}_s$. By Lemma 1,

$$\frac{\hat{\Delta}}{\widehat{\mathrm{SE}}} = \frac{S^{-1/2}\sum_s \sqrt{n_s}\,\hat{\Delta}_s/\sigma}{\left(S^{-1}\sum_s \hat{v}_s/\sigma^2\right)^{1/2} + o_p(1)} \Rightarrow \frac{\sum_s c_s|Z_s|}{\left(\sum_s c_s^2\right)^{1/2}} = T_S,$$

jointly over the $S$ independent states and by the continuous mapping theorem, the denominator being bounded away from zero. Every $\hat{\Delta}_s \geq 0$, so the two-sided Wald test rejects exactly when $\hat{\Delta}/\widehat{\mathrm{SE}} > z$, and since $T_S$ has a continuous distribution the rejection probability converges to $\Pr(T_S > z)$.

*Part (i).* For $S = 1$, $T_1 = c_1|Z_1|/c_1 = |Z_1|$ and $\Pr(|Z_1| > z) = \alpha$. For a pooled learner $\hat{Q}(s,a) = \hat{g}(s) + \hat{\beta}\,\mathbf{1}[a = 1]$ fitted by least squares with state effects, the residuals are orthogonal to the action indicator and sum to zero, so the inverse-propensity term again vanishes and $\hat{\Delta} = \hat{\beta}(1-\bar{a})$ if $\hat{\beta} > 0$ and $|\hat{\beta}|\bar{a}$ otherwise, with $\bar{a} \to p$ and $\sqrt{n}\hat{\beta}/\sigma \Rightarrow \mathcal{N}(0, 1/(p(1-p)))$. This is the $S = 1$ structure with $n$ in place of $n_s$, and the same computation gives the limit $|Z|$.

*Part (ii).* Write $A_s = c_s|Z_s|$ and $B_s = c_s^2$. These are independent across states with $\mathbb{E}A = \sqrt{H/\pi}$, $\mathrm{Var}\,A = H - 1 - H/\pi$, $\mathbb{E}B = H - 1$, $\mathrm{Var}\,B = H(H-2)$ and $\mathrm{Cov}(A,B) = \sqrt{H/\pi}\,(H-2)$, using $\mathbb{E}c^3 = \sqrt{2H}(2H-3)/2$ and $\mathbb{E}c^4 = 2(H-1)^2 - 1$. Since $T_S = \sqrt{S}\,\bar{A}/\sqrt{\bar{B}}$, the delta method gives $(T_S - \mu_S)/\varsigma \Rightarrow \mathcal{N}(0,1)$ with $\mu_S = \sqrt{S}\,\mathbb{E}A/\sqrt{\mathbb{E}B}$ and

$$\varsigma^2(H) = 1 - \frac{H}{\pi(H-1)} - \frac{H(H-2)}{\pi(H-1)^2} + \frac{H^2(H-2)}{4\pi(H-1)^3}. \tag{27}$$

At $H = 2$ the last two terms vanish and $\varsigma^2 = 1 - 2/\pi$. Because $\mu_S \to \infty$, the rejection probability tends to one. Against the exact limit, the approximation $\Phi((\mu_S - z)/\varsigma)$ errs by at most 0.03 for $S \geq 10$ on the grid of Table 3; it is poor for $S \leq 3$, where $T_S$ is far from normal.

*Part (iii).* $|Z_s|$ is independent of the sign of $Z_s$ and hence of $c_s$, so $\mathbb{E}[T_S \mid c] = \sqrt{2/\pi}\sum_s c_s/\|c\|$. By Cauchy–Schwarz, $\sum_s c_s \leq \sqrt{S}\,\|c\|$, with equality if and only if all $c_s$ are equal. For $p = \frac{1}{2}$ every $c_s = 1$; for $p \neq \frac{1}{2}$ and $S \geq 2$ unequal values occur with positive probability, so $\mathbb{E}[T_S] < \sqrt{2S/\pi}$. Finally $H/(H-1)$ is decreasing in $H$ on $[2,\infty)$, so $\mu_S$ decreases from $\sqrt{2S/\pi}$ at $H = 2$ to its limit $\sqrt{S/\pi}$.

*Part (iv).* If $\hat{Q}$ and $\hat{\pi}$ are fitted on an independent fold, then on the evaluation fold

$$\begin{aligned}\mathbb{E}[\phi_i \mid \hat{Q},\hat{\pi}] &= \mathbb{E}_s\big[\hat{Q}(s,\hat{\pi}(s)) + Q(s,\hat{\pi}(s)) \\ &\quad - \hat{Q}(s,\hat{\pi}(s))\big] - \mathbb{E}[Y] \\ &= \mathbb{E}_s Q(s,\hat{\pi}(s)) - \mathbb{E}_s\mathbb{E}_{a\sim\pi_b}Q(s,a) = 0\end{aligned}$$

under $\mathcal{D} = 0$. The $\phi_i$ are independent given the training fold, the influence-function variance is consistent, and $\hat{\Delta}/\widehat{\mathrm{SE}} \Rightarrow \mathcal{N}(0,1)$, so the rate converges to $\alpha$. □

*Monotonicity in overlap.* Part (iii) controls the conditional mean and the centring; the exact rate $\Pr(T_S > z)$ was evaluated on $p \in \{0.5, 0.35, 0.2, 0.1, 0.05, 0.02\}$ for $S \in \{1,2,3,5,8,13,20\}$ with $3\times10^5$ draws per point. It is constant at 0.050 for $S = 1$ and non-increasing as $p$ moves away from $\frac{1}{2}$ for every $S \geq 2$. We state this as a computation, not as a theorem.

*Remark* 11 (Finite samples and the cap). At finite $n$ the rate is below one, but its supremum over designs is one, approached through the number of selections $S$ (part (ii)) rather than through $n$ alone. Uniform logging is the worst case for state-wise learners (part (iii)) and irrelevant for learners that make a single selection (part (i)).

*Remark* 12 ($\tau$ is not monotone in model complexity). One might expect $\tau$ of Definition 3 to increase with the flexibility of the outcome model, as optimism grows with effective degrees of freedom [41]. It need not. The doubly robust correction is an inverse-propensity average of in-sample residuals. If the outcome fit memorises the data, the residuals and the correction vanish, the estimator reduces to the direct method, and the full optimism remains; the saturated learner of Theorem 2 is the extreme case, with $\tau = 0$ exactly. So $\tau \to 0$ at both ends, with little optimism to remove at low complexity and no residual to remove it with under memorisation, and it can peak at intermediate flexibility. The riskiest regime is memorisation, even though the outcome model looks better by the usual criteria.

*Remark* 13 (The correction can amplify the optimism it was introduced to remove). Because the correction is computed on the reused fold, nothing forces $\tau \geq 0$, and Definition 3 does not assume it. When $\tau < 0$ the corrected estimate is more optimistic than the direct method, so doubly robust machinery worsens the error that Theorem 2 describes. Using a doubly robust estimator is therefore not a mitigation for reuse; the fold discipline of §5.1 is.

## Appendix B.7. The optimism constant, measured

Definition 3 belongs to Remark 3, which covers learners between the two ends that Theorem 2 describes. The theorem itself contains no $\tau$: for the saturated learner the doubly robust correction vanishes identically ($\tau = 0$, Lemma 1(a)), and for the pooled learner the residuals are orthogonal to the action indicator, so part (i) applies without reference to $\tau$. What $\tau$ measures is what happens in between, and Table S13b reports it.

Three readings follow. First, $\tau$ is not monotone in flexibility: it peaks at intermediate complexity (a depth-4 tree, $\hat{\tau} = 0.544$) and returns to zero under memorisation, as Remark 12 argues and the deliberately overfit learner confirms exactly. Second, $\hat{\tau}$ is negative for gradient boosting with 200 trees, $-0.282$ $[-0.29, -0.27]$, so the correction there *adds* optimism rather than removing it (Remark 13). The sign reversal is an empirical finding and we do not derive it; no verdict in this paper depends on it. Third, its practical consequence is the one the audit acts on: a doubly robust estimator is not a mitigation for fold reuse, because the correction can run either way, and only a disjoint split returns the rate to $\alpha$ (Theorem 2(iv)).

*Remark* 14 (Validity floor in effective overlap). Part (iv) and the interval estimates of the audit are central-limit statements, and Assumption A5 can fail at extreme overlap. Under a uniform reference the second moment of the importance weight is $H/K = o_{\mathrm{eff}}^{-1}$, so as $o_{\mathrm{eff}} \to 0$ the weighted summands become heavy-tailed and the Wald approximation degrades at fixed $n$, even with a proper split. Results are therefore asserted only for $o_{\mathrm{eff}} \geq o^{\dagger}$, a design-dependent floor that the audit estimates by simulation with the log's own $n$ and weight distribution. Below it, the audit flags the log and reports a self-normalised or trimmed-weight interval (Appendix A.4).

# Appendix C. Supporting material for Section 5

This appendix specifies the instrument: the scales its thresholds live on, the gates and their evidence, the routing rule, and the operating characteristics it is held to. Everything measured on a corpus is in Appendix D.

## Appendix C.1. Scale conventions and the economic gate

The thresholds and the ceiling live on different scales, and mixing them can misstate ratios or required sample sizes by a factor of two or more. Break-even is defined on the absolute scale,

$$\mathcal{D}_{\mathrm{be}} = \left(c^{\mathrm{be}}/c_{\rho}\right)^{2},$$

where $c^{\mathrm{be}}$ is the fully loaded cost per decision. Its images on the normalised scales are

$$\eta^{2}_{a,\mathrm{be}} = \frac{\mathcal{D}_{\mathrm{be}}}{\mathrm{Var}_{\rho}(Y)}, \qquad \eta^{2}_{a|s,\mathrm{be}} = \frac{\mathcal{D}_{\mathrm{be}}}{\mathcal{D}_{\mathrm{be}} + \mathbb{E}_{\rho}[\mathrm{Var}(Y \mid s, a)]},$$

and the audit reports all three. A measured coupling is compared with break-even only on its own scale. Threshold comparisons, inversions of Proposition 3 and the exploration budget use the partial scale, so a pilot aimed at break-even targets $\eta^{2}_{a|s,\mathrm{be}}$; $\eta^{2}_{a,\mathrm{be}}$ reads directly against $\mathcal{D}$ and the money ceiling.

The per-decision cost $c^{\mathrm{be}}$ is the annualised form of (14): setting $\mathrm{NPV}_{\max} = 0$ with constant volume $N$ and operating cost $o$ gives $c^{\mathrm{be}} = \left(b\, a_{T,i}^{-1} + o\right)/N$, with $a_{T,i} = \sum_{t=1}^{T}(1+i)^{-t}$ the annuity factor. The horizon $T$, the discount rate $i$, and the sources of $b$ and $o$ are reported with every economic statement. The levels $(\alpha, 1-\beta)$ are configurable rather than fixed at $(0.05, 0.8)$: they are conventions, and an organisation should set them from the relative costs of a false kill and a false licence.

## Appendix C.2. The audit and the gate in detail

*Stage 0: a design claim is not a statistic.* Marginally uniform actions do not imply overlap. The Jensen bound $o_{\mathrm{eff}} \leq K^{2}/\sum_{a} p_{a}^{-1}$ is reached only when $\pi_b$ does not depend on the state (Remark 5), so actions can be uniform in aggregate yet predictable from the state. The Stage-0 statistics separate two cases that design descriptions blur. An uninformative propensity (C5b at or below zero, $\widehat{H}$ at its floor) means near-random assignment given the recorded state. A degenerate propensity ($\hat{\pi}_b(a \mid s)$ near 0 or 1) means the action is recoverable from the state, C4a fails and $\mathcal{D}$ is not identified. A deterministic business rule on recorded fields is the second case even when marginal shares are uniform. A design-based information count (route (i) of §3.4) replaces the estimated one only when genuine randomisation occurred and the propensities are known.

*The saturation ladder.* A measured coupling may reflect confounding rather than causation, so the audit re-estimates $\hat{\omega}^{2}$ over a sequence of richer state models fixed in the plan, together with a noise control in which each added covariate group is replaced by noise of the same degrees of freedom. Three constraints apply. (i) The covariates, their order and each rung's complexity are fixed in advance, since different orders give different paths. (ii) A decline is evidence of confounding only if the added covariates also predict the action; a decline without that is equally consistent with variance absorption, and a covariate measured after the decision can produce one through collider bias. The collapse branch therefore requires both the noise-adjusted decline and $\Delta_{\mathrm{prop}} > 0$, the held-out log-loss improvement of a propensity model when the enrichment covariates are added. (iii) Both verdicts are threshold tests stated with uncertainty, with tolerances $\delta_{\mathrm{pl}}$ and $\delta_{\mathrm{col}}$ fixed in the plan; a verdict resting on a point value that misses its threshold by less than its own interval width is reported as indeterminate. Because richer rungs change $df_1$, the detectability threshold is recomputed at each rung. (iv) The ratio is read on the *absolute* coupling $\hat{\mathcal{D}}$, not on the normalised $\hat{\omega}^{2}$. Simulation motivates this choice (Appendix D.11): a covariate that is both a confounder and a strong predictor of the outcome shrinks the numerator and the denominator of $\hat{\omega}^{2}$ together, and can shrink the denominator faster, so a ladder can rise on the normalised scale while the coupling it measures falls. On the absolute scale the same rungs decline as the construction intends.

On a randomised log the ladder answers a different question. The enrichment covariates cannot predict the action, so collapse cannot fire, and a rising ladder reveals effect modification: the coupling, and hence the ceiling for state-dependent policies, grows with the state resolution. The relevant ladder for a non-contextual licence is then the main-effect ladder, which is stable by construction under randomisation. The audit reports both and states which one bounds which policy class.

*Multiplicity.* The audit tests segments, the aggregate, and possibly several contrasts and rewards, so it controls multiplicity. Segmentation (definition, count, minimum size) is fixed in the plan. Segment tests are hierarchical, descending only when the aggregate test justifies it, and $p$-values are adjusted with Benjamini–Hochberg; raw and adjusted values are both reported. The family includes the main-effect contrast reported alongside the omnibus contrast and any additional reward audited on the same log. Without this control, auditing 100 lanes at $\alpha = 0.05$ would yield about five spurious coupled segments on a decoupled log.

*Properties of the three-branch rule.* (i) The branches are exhaustive and mutually exclusive. (ii) Each decisive branch is a one-sided comparison; with a two-sided $1-\alpha$ coupling interval, its one-sided error is at most $\alpha/2$ before accounting for the uncertainty of the threshold interval, and the self-check table reports the realised error of each branch on each log's design. (iii) The rule is conservative: it asserts neither side when the data cannot separate coupling from its threshold, which is the right default when a false refusal ends a valuable programme and a false licence sustains a dead one. (iv) It is monotone in information. The audit reports which interval binds when it abstains.

*The certification protocol.* Certification has four requirements. First, $(S, q)$ is fixed in advance with $S \geq 30$; with $S = 8$, a $7/8$ result has a one-sided $p$ of 0.035 against a coin and cannot separate a learner that certifies three times in four from one that almost always does. Second, every reported learner uses the same evaluation protocol (folds, seeds, estimator, hyperparameter search). Third, the deployed learner is fixed in advance, since choosing a certifying learner among several is selection. Fourth, the seed runs share one dataset and are strongly dependent, so the seed proportion is not a calibrated error rate; the null is calibrated by permuting the action within state strata and rerunning the whole pipeline $B \geq 999$ times, and $q$ is set to the upper 95% point of the resulting null distribution. For time-ordered logs the folds are rolling-origin splits, so that seeds differ in more than the learner's own randomness. Results are reported as a seeds-by-learners table of doubly robust estimates and lower limits, read by margin as well as verdict, together with the baselines that share the learner's policy class (for a non-contextual licence, the best single action chosen on the training fold).

*The coherence constraint.* Certification below the detectability threshold is inadmissible. If a learner certifies at a coupling whose interval does not clear $\eta^2_{\text{min,hi}}$, the audit reports the certificate as unsupported, routes the log by its detectability branch (0 (P) or 0 (P?)), and investigates the discrepancy rather than resolving it in the learner's favour. Theorem 2 is the reason: a learner that selects state by state produces confident improvements on a decoupled log at a rate set by the number of selections.

## Appendix C.3. Routing detail

*Refusal and abstention.* 0 (P) and 0 (P?) share a tier but not a response. P says that the action does not matter at the resolution the data provide, and permits using the log as a negative control. P? says that the audit cannot tell, and permits nothing beyond a pilot sized to resolve the frontier.

*Tier 2 with C?.* A Tier 2 verdict arises when the deployed learner certifies on an honest split while the ladder has not plateaued. On an observational log the certificate inherits ignorability from identification, and an unplateaued ladder means ignorability is not yet corroborated (Table 1, row A1); the licence covers optimisation on the logged distribution and measured shadow deployment, not rollout, and the positive-side robustness value of Appendix A.7 is reported. On a randomised log the causal reading is secured by design and C? records that the state specification is still unsettled.

*The band between the frontiers.* If Conjecture 4.3 holds and $\kappa' \leq \lambda_{\alpha,\beta,df_1}$ in (10), the band between the detectability and learnability frontiers is empty and the rule reduces to detectability plus certification. The audit prints $\hat{\kappa}'$ and $\lambda_{\alpha,\beta,df_1}$ whenever it uses the conjecture for planning.

*Reconciliation, and what a gate failure is not.* Aggregate and segment verdicts may disagree. The headline verdict is the aggregate; a segment verdict is issued only if the segment was specified in advance, meets the minimum size and survives the hierarchical multiplicity control, and it is reported as a scoped verdict, never as an upgrade of the dataset. Policy classes are reconciled the same way. A Tier 3 verdict for $\Pi_{\text{const}}$ says nothing about $\Pi$, whose contrast has more degrees of freedom and whose learner must estimate more; conversely, a purely interactive log ($\mathcal{D}_{\text{main}} = 0 < \mathcal{D}$) can support $\Pi$ while refusing $\Pi_{\text{const}}$. A gate failure is not a tier: C1 and C2 test whether $Q(s, a)$ is meaningful for the chosen reward, so the audit returns G for that (dataset, reward) pair and may still return a tier for another reward on the same log.

*Terminology.* [prov] is reserved for single-step verdicts issued for a sequential objective when C6 is detectable or could not be run. The qualification carried by C? is different: the objective may be single-step and the verdict still uncorroborated. A sequential objective with an unplateaued ladder is reported as 2 (C?) [prov].

## Appendix C.4. Operating characteristics and governance

*Why the two errors are reported separately.* A false kill sends a truly prescriptive log below Tier 3 and destroys value; it is made more likely by low power (small information counts, coarse binning, a ladder that absorbs signal). A false licence certifies a decoupled log, the failure this paper is about. A single accuracy figure would let one error hide the other.

*Three reporting rules under abstention.* Because (12) may abstain, directional error rates are reported conditional on a verdict, and the abstention rate is reported as its own operating characteristic, by distance from the frontier and by abstaining branch (detectability, certification, ladder); otherwise the instrument could achieve accuracy through silence. The share of truly prescriptive cells that received any verdict is reported, since "routes true Tier 3 cases correctly" otherwise describes only the cases the rule chose to answer. Adjacent-tier misroutings are reported by pair, because each tier carries a different response. Zero observed errors on a finite grid of $m$ verdicts are reported as $0/m$ with the exact one-sided 95% bound, about $3/m$.

*Governance.* The instrument is meant to stop an organisation from deploying the error it detects, so controls are built in. Condition C2 is enforced in code by a column schema, a training-time assertion and an automatic timestamp screen (Appendix A.9). A written plan fixes the analysis and thresholds before the target data are analysed and, where possible, is lodged with a public registry before the confirmatory analysis. It covers six choices that would otherwise be silent: the binning rule for continuous actions (Appendix A.8); the contrasts and the policy classes audited; the saturation ladder, its covariates, order, per-rung complexity, $\delta_{\text{pl}}$, $\delta_{\text{col}}$ and noise control; segmentation and its multiplicity control; the candidate rewards; and the certification protocol, including the learner, $(S, q)$, folds and estimator. A deviation from the plan is reported as a deviation, with its reason and the point at which it was decided.

# Appendix D. Supporting material for Section 6

This appendix holds the full audits and validation studies that Section 6 summarises. Each subsection is referenced from the paragraph of Section 6 it supports. It runs corpus by corpus (OBD, then the two logistics logs side by side, then each in full with its ladder, row accounting and measured design effect) and closes with the studies that test the instrument against a known truth.

## Appendix D.1. The OBD audit in full

The Open Bandit Dataset [15] logs recommendations on a fashion e-commerce platform under a genuinely uniform policy, $\pi_b \equiv 1/K$ for every impression, so overlap is perfect by construction ($o_{\text{eff}} = 1.0000$) and A1 holds by design. It is not a logistics log; we use it to validate the instrument. The reward is the click, the action the recommended item ($K = 80$ pooled), the state the four user-feature strata, and the cluster unit the 404 strata they form. The 80 affinity columns are excluded: they are computed from pre-week click history (which we verified), so C2 passes, but no verdict uses them. Gates C1, C2 and C4a pass; per-item CTR runs from 0.115% to 0.784%, so the click is not a function of the item.

**Two policy classes, two verdicts.** The rank condition decides the contrast. The main-effect contrast needs $df_1 = K-1 = 79$ against $\dim \mathcal{M}_0 = 406$; the interacted one would need $df_1 = (K-1)(1+d) = 32{,}074$ against $G = 404$ clusters and is not estimable. So the audit measures a non-contextual coupling, and §5.2 requires the two classes to be routed separately (Table S2).

Detectability is decided against the randomisation itself, not against (5): the cluster unit is the user stratum but items are randomised independently of it, so flipping residual signs within clusters does not break the item–click link and a wild cluster bootstrap is not a valid null. The null permutes item labels within cluster. Applying Algorithm 2 to that null (injecting a rescaled copy of the fitted action block and finding the smallest coupling detected with probability 0.8) gives $\eta^2_{\text{min,cal}} = 0.027 \times 10^{-3}$ with Monte-Carlo band $[0.025, 0.029]$. The three-branch rule (12) compares the coupling's lower limit with the threshold's upper limit, and here $\hat{L} = 0.513 \times 10^{-3}$ exceeds $\eta^2_{\text{min,hi}} = 0.029 \times 10^{-3}$ by a factor of 18 at $p \le 0.001$; the women's and men's campaigns clear theirs by 9.7× and 4.5×. Comparing the two point estimates instead would give 26×, 12× and 5.0×, and we report the branch-relevant ratio. The 95th percentile of the null is the test's critical value at size $\alpha$, not a detectability threshold at power $1-\beta$; a ratio against it would be 44× and would overstate the margin by a factor of 1.7.

The two ladders then separate. The main-effect ladder, which is the one that bounds $\Pi_{\text{const}}$ (Appendix C.2), is flat: enriching $\mathcal{M}_0$ with the four user features one at a time leaves the coupling at $0.70 \times 10^{-3}$ throughout, and the paired cluster-bootstrap upper limits of the last two relative changes are 0.0024 and 0.0051 against $\delta_{\text{pl}} = 0.15$. It plateaus, as randomisation guarantees it should. The covariate ladder, which bounds $\Pi$, rises monotonically from 0.70 to $10.31 \times 10^{-3}$ with upper limits far above $\delta_{\text{pl}}$ (Appendix D.2). On a randomised log a rising ladder is effect modification, not confounding: $\pi_b$ is constant, so $\Delta_{\text{prop}} = 0$ by construction and the *collapse* branch cannot fire. This corpus therefore cannot test whether the ladder separates confounding from variance absorption, and we make no such claim.

One alternative explanation of the rise has to be excluded before it can be read as effect modification. The covariate ladder's degrees of freedom grow from 79 at L0 to 19,456 at L4, by which point the 404 × 80 cells hold 0.15 clicks each, and the degrees-of-freedom correction in $\hat{\omega}^2$ is derived for balanced unweighted analysis of variance, not for sparse weighted Bernoulli outcomes (Appendix A.5(i)). A ladder could therefore rise simply because the correction under-corrects as $df_1$ grows. We test this directly by running a permutation null at each rung, so that the null carries that rung's own degrees of freedom. It does not rise: the null mean is $0.004 \times 10^{-3}$ at L0 and $0.002 \times 10^{-3}$ at L4, and subtracting it moves the $L0 \to L4$ rise from 14.68× to 14.76× (Appendix D.2). The rise is in the data, not in the estimator.

**Certification.** The deployed learner is a shrunk tabular estimator, $\hat{Q}(a) = (\sum_{i \in \text{train}} y_i \mathbf{1}[a_i{=}a] + \alpha \bar{y})/(n_a + \alpha)$ with $\alpha = 500$, acting greedily; it is the $\gamma = 0$ case of a conservative value-based learner. Seeds vary the 80% training subsample, not the split, which is fixed by date (train days 1–4, evaluate 5–7). Against a null that permutes clicks within stratum and reruns the pipeline 999 times (Appendix C.2 requires $B \ge 999$), the 95th percentile of the certified-seed proportion is 0.167, so $q = 0.5$ is calibrated. The non-contextual learner certifies in 26/30 seeds ($p = 0.003$). A contextual learner certifies in 23/30 against the same logging baseline, but that is not the quantity the $\Pi$ verdict is decided on: $\Pi_{\text{const}} \subset \Pi$, so a contextual policy beating the logging policy is implied by the fixed rule already certified (§5.2). What has to be certified for $\Pi$ is the increment over that fixed rule, and it is not.

**Verdict, by class.** $\Pi_{\text{const}}$ clears detectability, plateaus and certifies, so the routing rule returns **Tier 3** for non-contextual rules. For $\Pi$ the audit asks what personalisation *adds* (§5.2), and the two halves of that question separate. The increment is detectable: at the one estimable interacted contrast, item by the four-level `user_feature_0` ($df_1 = (K-1)(L-1) = 237$), the interaction coupling is $0.601 \times 10^{-3}$ against a within-cluster permutation null whose 95th percentile is $0.0001 \times 10^{-3}$ ($p = 0.005$). But it does not certify: the paired difference between the contextual policy and the certified fixed rule, on the same evaluation fold, is −0.00145 [−0.00357, +0.00066] click per impression, and 0/30 seeds certify it. The point estimate is negative but the interval contains zero, so the log rules out a gain above +0.00066 rather than establishing a loss. The verdict for $\Pi$ is therefore **1 (L)** (the effect is there, but the log cannot turn it into value), and the covariate ladder's C? is recorded alongside. Both verdicts carry [econ] not evaluated, the corpus having no cost data.

Reporting $\Pi$ at Tier 2 (C?) on the same coupling and the same certificate as $\Pi_{\text{const}}$ would state nothing that the $\Pi_{\text{const}}$ verdict does not already imply, since $\Pi_{\text{const}} \subset \Pi$, and a reader would reasonably take "Tier 2 for contextual rules" to mean that personalisation may go to shadow deployment, which is the opposite of what these data show.

**What the ceiling is worth knowing.** The ceiling is $c_\rho\sqrt{\mathcal{D}_{\text{main}}} = 0.0139$ click per impression at $c_\rho = \sqrt{79}$, and the certified policy reaches +0.00411, 30% of it. That 30% is almost entirely the looseness of the bound and not a shortfall of the learner. Choosing the best item on the training days and measuring it on the evaluation days, an honest estimate of the best attainable non-contextual gain, gives +0.00394, itself 28.4% [14.3%, 42.5%] of the ceiling (Table S4). The learner is slightly but detectably better: the paired difference over the same evaluation rows is +0.000175 [+0.000021, +0.000329], excluding zero, so shrinkage adds a little over picking the single best item. Both percentages carry wide intervals (the learner's 29.7% spans [15.7%, 43.7%]) because the ceiling is divided into an uplift estimated from 2,221 clicks; the point of the comparison is the order of magnitude, not the second digit. Theorem 1(a) is attained only when one action carries the whole advantage at every state, and at $K = 80$ the Cauchy–Schwarz constant is loose by a factor of 3.4 here. For a manager the reading is that the dollar ceiling is conservative, and increasingly so as the action set grows; Remark 1 is the formal statement and this is its magnitude.

Three further checks guard the verdict. The estimator: run on the random log, our doubly robust code recovers the value of the Bernoulli Thompson policy that ran concurrently as 0.004874 against an on-policy ground truth of 0.004953, an error of −1.6% with the interval covering truth. The baselines: on the same split the contextual rule reaches +0.0023 (17% of the ceiling) and the platform's own Thompson policy +0.0014 (10%), so the non-contextual rule is not beaten by either (Appendix D.2); the paired difference against the fixed rule is what the $\Pi$ verdict is decided on, and its interval contains zero. The folds: the fixed split is not load-bearing. Rolling-origin splits (train days $1, \dots, k$, evaluate day $k+1$, for $k = 3, \dots, 6$) evaluate on one day each and carry about 700 clicks apiece, so individually only two of four exclude zero. Pooling the four evaluation days gives 824,429 impressions and 2,856 clicks, more than the fixed split's 614,691 and 2,221, and an uplift of +0.00338 with a day-block bootstrap interval of [+0.00205, +0.00476] that excludes zero. The fixed-split and pooled-rolling estimates agree, so the verdict does not depend on which honest split is used.

**This audit is exploratory.** The written plan of Appendix C.4 covers Olist only. Every choice made for this corpus (the state, the cluster unit, the day-1–4/5–7 split, the learner and its shrinkage) was made with the data in hand, so the positive verdict shows that the instrument *can* license a log, not that it does so under a plan fixed in advance. Limitation (i) records this.

## Appendix D.2. The two OBD ladders, and the baselines

Table S3 gives both ladders rung by rung. The main-effect ladder, which bounds $\Pi_{\text{const}}$, is flat to within 0.5% across four enrichments; the covariate ladder, which bounds $\Pi$, roughly doubles at each of the last two rungs. On a uniformly randomised log the second is effect modification by construction, since $\Delta_{\text{prop}} = 0$ and the collapse branch cannot fire.

Both are read from one set of cluster resamples, so that each relative change is a paired quantity. Resampling each rung independently and taking percentile intervals would put rung L3 at [3.873, 16.10] around a point estimate of 3.835, an interval that excludes its own estimate: this is the upward bias of the percentile bootstrap for an $R^2$-type statistic that Appendix A.5(v) describes. With shared resamples every interval contains its estimate, and the ladder verdict is the same under either scheme.

A sharper interval is available on this corpus and we report it for the base rung. Because the logging policy is known and uniform, the permutation distribution is exact, so the interval can be obtained by inverting the permutation test: a rescaled copy of the fitted action block is injected at each candidate coupling, and the interval collects the levels at which the observed $\hat{\omega}^2$ falls between the 2.5th and 97.5th percentiles of the injected distribution. This gives $[0.621, 0.789] \times 10^{-3}$ for the base rung against the cluster bootstrap's [0.513, 1.596], a fourfold narrower interval, centred on the estimate rather than below it. The three-branch comparison uses the bootstrap limit, which is the conservative one for a positive verdict; the inverted interval is reported so the magnitude of the percentile bootstrap's skew is visible.

## Appendix D.3. The two logistics audits, side by side

Table S5 carries the full audit of both logistics logs in one template, so that the two refusals can be read stage by stage. Threshold rows marked "iid $v$" use the independent-rows convention and are shown for comparison; verdicts use the calibrated footing of Table S8.

## Appendix D.4. The DataCo audit in full

We audit the widely used public supply-chain decision log [DataCo; 17], treating the committed shipping mode as the action. It is the de facto benchmark for the class of programmes our proposed instrument is meant to gate.

The gate applies before estimation. The main service target violates C1, and no sample size can repair that. The late-delivery flag equals $\mathbf{1}$[realised > scheduled] in 97.55% of rows, and within each mode the scheduled promise takes a single value. The target is therefore effectively an accounting identity of the action. Consistently, the leakage protocol's AUC drops from 0.980 to 0.642 when post-decision variables are removed. No sampling choice, overlap restriction, or model flexibility can fix a definitional target, so we record verdict 0 (G). Figure S1 shows the consequences of optimising such a target. The same inspection also speaks to what the corpus is. Across 180,519 rows, realised transit time takes at most five distinct integer values within a mode; for *First Class* it is always 2.000 days (SD = 0). Realised transit for *Second* vs. *Standard Class* differs by 0.005 days ($p = 0.56$, Mann–Whitney), matching the degeneracy in Figure S1. Zero within-mode variance in physical delivery time is implausible for a real carrier network, so we describe DataCo as a public benchmark of uncertain provenance, not field evidence, and do not use it as evidence of the instrument's licensing behaviour. We nonetheless retain the C1 finding: a target that is an accounting identity of the action in 97.55% of rows is a community-relevant defect regardless of the file's origin, and its prominence in the literature makes reporting it especially important.

**Table S2**
The randomised positive control, audited on the conventions of Table S5. All three campaigns clear their calibrated detectability threshold. On the pooled campaign the two classes route differently: non-contextual rules reach Tier 3, while the increment personalisation would add is detectable but does not certify, so contextual rules return 1 (L).

| | all | women | men |
|---|---|---|---|
| $n$; $K$; clusters $G$ | 1,374,327; 80; 404 | 864,585; 46; 351 | 452,949; 34; 345 |
| CTR | 0.347% | 0.482% | 0.512% |
| $\pi_b$; $o_{\text{eff}}$ | 1/80 exact; 1.000 | 1/46 exact; 1.000 | 1/34 exact; 1.000 |
| $df_1$ (main effect) | 79 | 45 | 33 |
| *Detectability, against the corpus's own randomisation* | | | |
| coupling $\hat{\omega}^2$ | 0.703 | 0.497 | 0.272 |
| $\hat{L}$ (lower limit, cluster bootstrap) | 0.513 | 0.440 | 0.265 |
| by inverting the permutation test | 0.621 | — | — |
| $\eta^2_{\text{min,cal}}$ [MC band] | 0.027 [0.025, 0.029] | 0.041 [0.038, 0.045] | 0.054 [0.048, 0.060] |
| $\hat{L}/\eta^2_{\text{min,hi}}$ (branch ratio) | 17.9× | 9.7× | 4.5× |
| permutation $p$ ($B = 999$) | $\leq 0.001$ | $\leq 0.001$ | $\leq 0.001$ |
| critical value (null 95th pct) | 0.016 | 0.020 | 0.032 |
| *Ladder, by policy class* | | | |
| main-effect ladder ($\Pi_{\text{const}}$) | **plateau** | — | — |
| covariate ladder ($\Pi$) | **no** (C?) | — | — |
| $\Pi$*: what personalisation adds* | | | |
| $\mathcal{D}_{\text{int}}$ at item×uf$_0$ ($df_1$=237) | 0.601 | — | — |
| null 95th pct; $p$ | 0.0001; 0.005 | — | — |
| paired uplift over fixed rule | −0.00145 [−0.00357, +0.00066] | — | — |
| seeds certifying the increment | 0/30 | — | — |
| *Certification, $B = 999$ null, $q = 0.5$ calibrated at 0.167* | | | |
| $\Pi_{\text{const}}$: seeds; $p$ | 26/30; 0.003 | — | — |
| **Verdict** $\Pi_{\text{const}}$ | **Tier 3** [econ] | detectable | detectable |
| **Verdict** $\Pi$ (increment) | **1 (L)** [econ], C? | — | — |

**Table S3**
The two OBD saturation ladders ($\eta^2_{a|s}$, $\times 10^{-3}$), from one set of 199 cluster resamples. The main-effect ladder plateaus; the covariate ladder does not. The covariate ladder also carries a per-rung permutation null: items are permuted within cluster 199 times at each rung, so the null inherits that rung's degrees of freedom. Every rung has $p = 0.005$, the smallest 199 permutations resolve.

| | main-effect ladder ($\Pi_{\text{const}}$, $df_1 = 79$) | | | covariate ladder ($\Pi$) | | | |
|---|---|---|---|---|---|---|---|
| Rung | $\hat{\omega}^2$ | 95% CI | $\lvert\Delta_{\text{rel}}\rvert$ up. | $df_1$ | $\hat{\omega}^2$ | null mean | $\hat{\omega}^2$ adj. |
| L0 base | 0.703 | [0.513, 1.596] | — | 79 | 0.703 | 0.0040 | 0.699 |
| L1 +uf$_0$ | 0.702 | [0.512, 1.596] | 0.0009 | 316 | 1.303 | 0.0011 | 1.302 |
| L2 +uf$_1$ | 0.702 | [0.513, 1.594] | 0.0031 | 1,140 | 1.899 | 0.0000 | 1.899 |
| L3 +uf$_2$ | 0.702 | [0.513, 1.595] | 0.0024 | 4,443 | 3.835 | 0.0000 | 3.835 |
| L4 +uf$_3$ | 0.703 | [0.514, 1.594] | 0.0051 | 19,456 | 10.315 | 0.0024 | 10.312 |
| **Branch** | **plateau** ($\delta_{\text{pl}} = 0.15$) | | | **C?**: last two $\lvert\Delta_{\text{rel}}\rvert$ upper limits 1.02, 1.71 | | | |

**Table S4**
Policies on the OBD evaluation split (train days 1–4, evaluate 5–7), all evaluated with the doubly robust estimator against the uniform logging policy. The ceiling is $c_\rho\sqrt{\mathcal{D}_{\text{main}}} = 0.0139$ click per impression.

| Policy | uplift | 95% CI | % ceiling |
|---|---|---|---|
| *Non-contextual ($\Pi_{\text{const}}$)* | | | |
| best single item, chosen on train | +0.00394 | [+0.00198, +0.00590] | 28.4% [14.3, 42.5] |
| deployed learner, $\alpha = 500$ | +0.00411 | [+0.00218, +0.00605] | 29.7% [15.7, 43.7] |
| paired difference | +0.000175 | [+0.000021, +0.000329] | — |
| pooled rolling-origin (4 folds) | +0.00338 | [+0.00205, +0.00476] | 24.4% |
| *Contextual ($\Pi$), 404 strata* | | | |
| by stratum, $\alpha = 50$ | +0.00233 | [+0.00056, +0.00411] | 17% |
| $\alpha = 500$ | +0.00238 | [+0.00067, +0.00409] | 17% |
| $\alpha = 5000$ | +0.00232 | [+0.00061, +0.00403] | 17% |
| Bernoulli TS (on-policy, same days) | +0.00136 | [+0.00131, +0.00142] | 10% |

The audit is underpowered on the margin reward, and the two policy classes stop at different branches of (12): 0 (P?) for fixed rules, whose coupling interval overlaps the threshold, and 0 (P) for contextual ones, whose interval lies below it (Table S8). Stage 0 (Table S5) explains the weakness: the fitted propensity model adds little beyond marginal action rates (C5b = −0.009 nats), and $\widehat{H} = 7.973$ is near its Jensen lower bound 7.958, so logging is almost random with respect to recorded state. We do not claim more than that. A near-uniform $\pi_b$ implies only that recorded covariates do not drive actions; it says nothing about unrecorded $U$, and ignorability for $U$ is untested. The masking construction in Appendix A.7 gives the counterexample: if a dispatcher acts on $U$, then $\pi_b(a \mid s) = 1/K$ for all $s$ and $\widehat{H}$ reaches the Jensen floor, even though the true coupling is maximal.

Thus, the binding constraint is lack of power, not measured confounding. Overlap is only moderate ($o_{\text{eff}} = 0.502$), because action frequencies are skewed, not because actions depend on the state; the identity $1/o_{\text{eff}} = \mathbb{E}[w^2]/\mathbb{E}[w]^2 = 1.993$ also holds. Near-random logging is the most exposed regime under reuse, but only within the factor of $\sqrt{2}$ in drift that Theorem 2(iii) allows, a bounded effect on the rate; the magnitude is set by the number of state cells the learner selects over. We do not attach the warning to the learner deployed here: measured at the Olist configuration, that learner's effective number of selections is 1.4, so on decoupled data it behaves like a pooled one (§6.5). What makes reuse unsafe is not a property of any particular learner but the absence of a split, and Theorem 2(iv) is what the audit relies on (Table S11). The intervals are wide because the log has only 23 lanes. A residual intraclass correlation estimated at zero would suggest the clustered footing changes nothing, but that is the design effect of a mean, not the quantity (5) requires, and the correct quantity is not one: the measured design effect of the action block is $\bar{\delta} = 3.35$ [2.91, 3.50] at $df_1 = 3$ and 3.60 [3.31, 3.70] at $df_1 = 21$ (Appendix D.7). The threshold band at $df_1 = 21$ therefore rises from $[0.174, 0.356]$ to $[0.779, 0.869] \times 10^{-3}$, which leaves the coupling interval $[0.066, 1.05] \times 10^{-3}$ straddling it. Calibrating the threshold on the bootstrap that produced the interval (Algorithm 2) settles the branch, and it settles it differently for the two policy classes. For $\Pi_{\text{const}}$, which carries the headline because its verdict is the least sensitive to the uncertainty in $\bar{\delta}$, the threshold is $[0.235, 0.354] \times 10^{-3}$ against a coupling of $[0.000, 0.295]$: the two overlap, so the verdict is 0 (P?) and the audit cannot separate this corpus's coupling from what its design can detect. For $\Pi$ the threshold is $[1.09, 1.73]$ against $[0.066, 1.05]$, which lies below it, giving 0 (P). Both are Tier 0 and neither supports a policy; the distinction is what the firm should do next, since P licenses use as a negative control while P? licenses only a pilot sized to resolve the frontier (Appendix C.3). On the independent-rows footing the two codes swap, which is why we report the footing alongside every code (Table S8).

**Economic gate.** The estimated cap is \$1.92 per order over the reference design (3.85× break-even), an upper bound of \$347,000 in total value against \$90,000 implementation cost; both sides of (13) (Table S5) equal 14.7. This is not a realised payout: the cap is an upper bound, and (14) forbids transferring a realisation rate $r$ from one log to another, since $r$ depends on the learner, the reward and the state resolution. We therefore report $\text{NPV}_{\max}$ and a sensitivity map over $(b, r)$ instead of a point payout: the programme repays its \$90,000 build cost only for $r > 0.26$, and no measurement on this log licenses a value of $r$. Because the economic and statistical gates agree, the instrument yields a single recommendation, not "large prize, unresolvable evidence". Solving (8) at 80% power gives the pilot, and the two policy classes price differently. For contextual rules ($df_1 = 21$) it takes $n_{\text{eff}} \approx 187{,}000$ orders to verify the measured coupling and $\sim 2.7$M to test break-even; for non-contextual rules ($df_1 = 3$, $\lambda = 10.90$) the first figure falls to $\approx 95{,}000$. At $\sim 67{,}000$ orders per year that is 2.8 years of uniform randomisation against 1.4, while the break-even pilot is $\sim 21$ years even in the cheaper class and so is infeasible either way. A firm that only needs to choose one default mode can therefore obtain its answer in half the time, which is the practical content of routing by policy class. Because a pilot is randomised it falls under route (i) of §3.4,

**Table S5**
Unified audit of the two public logs, one column each, so the refusals can be read side by side.

| | **DataCo** (margin) | **Olist** (review) |
|---|---|---|
| *Gates* | | |
| C1 non-definitional target | **FAIL** (on-time, 97.55%) | pass |
| C2 leakage, AUC pre/post | $0.980 \to 0.642$ | pass |
| *Stage 0: estimability, overlap, dependence* | | |
| C4a state mass below $\delta$ | — | 0.16% (pass) |
| C5a ECE; C5b nats vs marginal | —; −0.009 (**fail**) | 0.037; +0.27 |
| $\widehat{H}$; Jensen floor; gap | 7.973; 7.958; 0.015 | 13.6; 4; — |
| *Row flow* | | |
| raw orders; delivered; analysed | 180,729; —; 180,519 | 99,441; 96,478; 95,306 |
| development window; held-out | — | 56,383; 38,923 |
| retained after trim $\pi_b \geq \delta$ | — | 56,295 (99.84%) |
| $o_{\text{eff}}$; identity check | 0.502; = 1/1.993 ✓ | 0.304; = 1/3.29 ✓ |
| $\max_i w_i$; validity floor $o^{\dagger}$ | 7.4; — | 24.5 trimmed; 524 untrimmed; $\lesssim 0.012$ |
| $\bar{m}$ (arithmetic; Kish); clusters $G$ | 7,849; 14,231; 23 | 2,088; 11,433; 27 |
| $n_{\text{eff}}$ | 90,738 | 17,133 |
| held-out: $n_{\text{eff}}$; $\bar{m}$; $G$ | — | 11,882; 1,442; 27 |
| *Stage 1a: specification of the omnibus contrast* | | |
| intercept; state covariates $d$ | 1; 6 | 1; 11 |
| action main effect $K-1$ | 3 | 3 |
| action×state $d(K-1)$ | 18 | 33 |
| $\dim \mathcal{M}_0$; $\dim \mathcal{M}_1$; $df_1$ | 7; 28; 21 | 12; 48; 36 |
| rank check $df_1 < G-1$ | 21 < 22 ✓ | 36 > 26 **FAILS** |
| interacted spec. carried; $df_1$ | as planned; 21 | reduced; 9 |
| exact $\lambda_{\alpha,\beta,df_1}$; $\nu$ | 21.37; 25,198 | 15.70; 2,518 |
| *Stage 1b: coupling against detectability* ($\eta^2_{a\|s}$, $\times 10^{-3}$) | | |
| $\Pi$*: interacted contrast* ($df_1 = 21$; 9) | | |
| $\bar{\delta}$ of action block [jackknife] | 3.60 [3.31, 3.70] | 6.75 [5.07, 7.00] |
| $n_{\text{info}} = n_{\text{eff}}/\bar{\delta}$ | 25,226 | 2,539 |
| coupling $[\hat{L}, \hat{U}]$ (wild cluster bootstrap) | [0.066, 1.05] | [0.30, 4.08] |
| threshold $[\eta^2_{\text{min,lo}}, \eta^2_{\text{min,hi}}]$ (iid $\nu$) | [0.779, 0.869] | [4.62, 6.37] |
| branch of (12) | **indeterminate** | **below** |
| held-out: $\bar{\delta}$; coupling; threshold; branch | — | 3.25; [1.46, 7.22]; [3.79, 4.51]; **indet.** |
| $\Pi_{\text{const}}$*: main-effect contrast* ($df_1 = 3$)*, the headline* | | |
| $\bar{\delta}$ [jackknife]; $n_{\text{info}}$ | 3.35 [2.91, 3.50]; 27,124 | 4.15 [3.62, 4.88]; 4,133 |
| coupling $[\hat{L}, \hat{U}]$; threshold (iid $\nu$); branch | [0.000, 0.295]; [0.350, 0.420]; **below** | [0.00, 0.96]; [2.30, 3.10]; **below** |
| held-out: coupling; threshold; branch | — | [0.46, 4.67]; [2.66, 3.86]; **indet.** |
| *Threshold calibrated on the same bootstrap (Alg. 2)* | | |
| $\Pi_{\text{const}}$: threshold; branch | [0.235, 0.354]; **indet.** | [1.55, 2.33]; **below** |
| $\Pi$: threshold; branch | [1.09, 1.73]; **below** | [9.40, 16.9]; **below** |
| held-out: $\Pi_{\text{const}}$; $\Pi$ | — | [2.66, 4.46] **indet.**; [9.03, 17.1] **below** |
| *Stage 1c: saturation ladder* $L4/L0$ *on* $\hat{\mathcal{D}}$ [CI] (Table S9) | | |
| development, $df_1 = 9$; noise control | — | 1.17 [0.73, 2.57]; 1.02 |
| held-out, $df_1 = 9$; $df_1 = 3$ | — | 1.29 [0.88, 2.03]; 1.45 [0.92, 2.32] |
| propensity test $\Delta_{\text{prop}}$ [CI] | — | −0.191 [−0.358, +0.033] nats, dev |
| held-out $\Delta_{\text{prop}}$ [CI] | — | −0.135 [−0.281, +0.067] nats |
| branch | — | no collapse; C? |
| *Stage 2: certification* | | |
| seeds at $(S, q)$=(30, 0.5); null | — | 29/30; $p < 0.01$ |
| *Segment tests* (hierarchical, Benjamini–Hochberg) | | |
| segments tested; surviving; expected | — | 37; 6; 1.9 |
| smallest raw $p$; smallest adjusted $p$ | — | $3.7 \times 10^{-46}$; $1.1 \times 10^{-44}$ |
| *Economic gate* (informational below Tier 1; binding at Tier ≥ 1) | | |
| ceiling over reference; incumbent | \$1.92; \$3.83 | out of routing |
| break-even $\eta^2_{\text{be}}$ | 0.0078 ($\times 10^{-3}$, i.e. $7.8 \times 10^{-6}$) | out of routing |
| square identity (13), both sides | 14.7 = 14.7 ✓ | out of routing |
| **Verdict** $\Pi_{\text{const}}$ ($df_1$=3), headline | **0 (G)** on-time; **0 (P?)** margin | **0 (P)** dev; **0 (P?)** held-out |
| **Verdict** $\Pi$ ($df_1$=21; 9) | **0 (G)**; **0 (P)** margin | **0 (P)** both windows |

*Notes.* "—" indicates a check that is not binding for that corpus.

**Table S6**
What an offline agent does on a decoupled benchmark. Panel (a) is on a disjoint fold.

| (a) Off-policy evaluation, margin reward (logging value 0.0211) | | | |
|---|---|---|---|
| Estimator | Value | 95% CI | Excl. logging? |
| IPS | 0.0174 | [0.0051, 0.0307] | no |
| SNIPS | 0.0172 | [0.0051, 0.0303] | no |
| DR | 0.0168 | [0.0046, 0.0301] | no |
| FQE | −0.0008 | [−0.0017, 0.0002] | below |
| (b) Reward-weighting grid (all learn 100% a single mode) | | | |
| Weighting | DR uplift | CI excl. 0 | Reading |
| margin only | −20.0% | no | no gain |
| margin + on-time | +34.1% | **yes** | definitional |
| on-time heavy | +36.3% | **yes** | definitional |
| margin − fraud | −41.3% | no | no gain |
| (c) What the certified policy delivers | | | |
| Certified metric gain (on-time) | | | +23.3 pp |
| Change in realised delivery speed | | | +0.000 days |
| Change in promised lead time | | | +1.07 days |
| Net change in customer outcome | | | −1.07 days |

so its threshold comes from the randomisation distribution rather than from (8); the figures here are planning estimates and the audit recomputes them on the realised design.

**Implications for a production agent.** We nonetheless trained a conservative offline agent (discrete CQL; 11) on order-time features, with leakage checks and a stricter temporal split (126,363/54,156), evaluating on the later hold-out (the benign split case of Theorem 2(iv)). The learned policy is degenerate, putting all mass on one shipping mode regardless of business weighting, consistent with decoupling or with the unequal-variance account in §4.2. Table S6 shows that no estimator certifies a margin improvement. Interval widths differ by a factor of 13 across estimators, so tight importance-weighting intervals are not evidence, while the low-variance model-based estimator indicates a loss. The only "certified" gains arise under weightings that overemphasise the definitional on-time component, so the certificate is a certificate of an artefact. Figure S1 shows why: the policy increases promised delivery by 1.07 days, does not speed realised delivery, yet yields a reported 23.3-point metric increase.

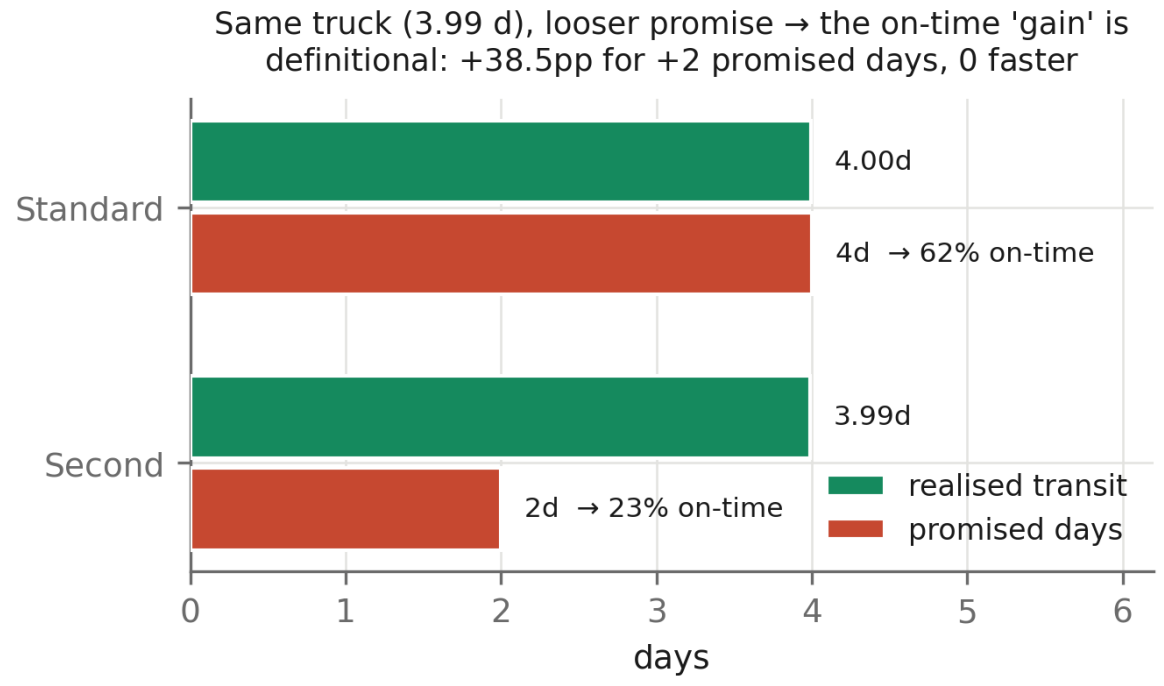

**Figure S1:** The Goodhart mechanism. Two service levels with the same realised delivery time (3.99 vs 4.00 days) differ by 38.5 on-time percentage points, entirely through the promised lead time.

## Appendix D.5. The Olist audit in full

In the second corpus [Olist; 23], the delivery promise is treated as the decision variable and the customer review as the reward. The degrees of freedom are tied to the actions, not to geography (Table S5, Stage 1a): with $K = 4$ and $d = 11$, $df_1 = (K-1)(1+d) = 36$ (three action indicators plus 33 state interactions), while distance and same-state enter as continuous covariates instead of using a 37-level lane fixed effect. Even so, lane geography is strongly (though not perfectly) predictive of the promise: a held-out propensity model reaches 0.49 accuracy versus a 0.25 baseline (C5b = +0.27 nats) and is well calibrated (C5a ECE = 0.037), implying that the extremely small propensities are genuine.

If we do not trim, observations with $\pi_b \approx 4.8 \times 10^{-4}$ produce $\max_i w_i \approx 524$ and push $n_{\text{eff}}$ down to 2,905, i.e. $o_{\text{eff}} = 0.05$. On the pre-specified trimmed support ($\pi_b \geq \delta = 0.01$, covering 99.84% of rows), the diagnostics agree and a validity-floor simulation yields a decoupled-null Wald rate of $0.040 \approx \alpha$, so value claims are permissible. Adjusting $\delta$ within its admissible range leaves the routing takeaway unchanged, though not the branch: on the Rao–Scott footing of (5) the coupling lies below its threshold at the registered trim and at every tighter one, and is indeterminate against it at looser trims, so no admissible trim reaches a positive tier (Table S10). This corpus also clarifies why C4a and C4b cannot be conflated: although only 0.16% of rows are below $\delta$, that same 0.16% drives the untrimmed $n_{\text{eff}}$ from 17,133 to 2,905.

**The certificate.** Only the deployed conservative learner certifies (Table S7). Under the pre-specified ($S$=30, $q$=0.5)

configuration with CQL held fixed, the doubly robust interval excludes zero in 29/30 honest-split seeds, with an average uplift of +0.023 review points. A permutation null (shuffling the promise within geographic strata and rerunning the full pipeline) gives a 95th-percentile seed proportion of 0.25; hence $q = 0.5$ is calibrated and observing 29/30 implies $p < 0.01$.

The written plan named one omnibus contrast with $df_1 = 36$, but at $G = 27$ a cluster-robust Wald test of 36 restrictions is rank-deficient, so we report three contrasts: the main-effect contrast $df_1 = 3$, which carries the headline $\Pi_{\text{const}}$ verdict, and the richest contrast that satisfies $df_1 < G - 1$ while retaining action-by-state interaction, $df_1 = 9$, which carries the $\Pi$ verdict. Measured coupling falls monotonically as the contrast is restricted: $3.20 \times 10^{-3}$ at $df_1 = 36$, $0.80 \times 10^{-3}$ at $df_1 = 9$, and essentially zero at $df_1 = 3$. Each is compared with a threshold calibrated on the wild cluster bootstrap (Table S8), and both carried contrasts are 0 (P): [0.00, 0.96] against [1.55, 2.33] at $df_1 = 3$, and [0.30, 4.08] against [9.40, 16.9] at $df_1 = 9$. The rank-deficient contrast would be 0 (P?) on the independent-rows footing ([2.23, 11.93] against [6.88, 7.88]), which is a third reason not to carry it, beyond the rank failure and the loading of coupling onto terms 27 clusters cannot identify. The ordering is itself the finding: the promise has almost no average effect on the review, and nearly all of what the saturated contrast attributes to the action sits in the 33 interaction terms. The larger number is not a better measurement of the smaller one; it is the same signal read through a specification the data cannot carry.

The audit's coherence constraint (Appendix C.2) forbids certification at a coupling that does not clear $\eta^2_{\text{min,hi}}$, and here it does not: [0.00, 0.96] against [1.55, 2.33] at $df_1 = 3$ and [0.30, 4.08] against [9.40, 16.9] at $df_1 = 9$. By Proposition 2 a positive certificate at such a coupling is more plausibly selection on fitting error than signal, so the audit reports the certification as unsupported and routes by the detectability branch. This is the paper's own rule applied to its own corpus, and the clearest illustration of the central claim: a conservative learner produced a clean 29/30 certificate on a coupling that 27 clusters cannot resolve.

The ladder is run on the audit's own specification (the registered trim, the importance weights and the 27 customer-state clusters) with 199 paired cluster resamples shared across rungs, and on both scales, since §5.2 reads the ratio on the absolute $\hat{D}$ (Table S9). Enriching order-time state with geography, category, zip and expanding seller- and lane-level prior transit time does not reduce the coupling on this specification. At the interacted contrast the ratio is 1.17 [0.73, 2.57] on the development window and 1.29 [0.88, 2.03] on the held-out one; at the main-effect contrast the held-out window gives 1.45 [0.92, 2.32]. Every cell sits far above $\delta_{\text{col}} = 0.5$, and the pure-noise control stays within 4% of one throughout, so the movement is not an artefact of the growing state model. The branch is C? and the verdict is set by detectability.

The ladder is sensitive to these choices. Read on the normalised scale, with a weighting and clustering that differ from the audit's, the same enrichment gives ratios of 0.47 on development and 0.40 on held-out, close to $\delta_{\text{col}}$ and with intervals that cross it. On the audit's own specification, and on the absolute scale that step 2 requires, the ladder does not decline. Neither reading supports a collapse verdict or a confounding claim.

The propensity condition points the same way. Whether an enrichment-driven movement is confounding or variance absorption turns on whether those features predict the *action*, and they do not: adding the history terms to a held-out propensity model for the promise bin raises multinomial log-loss by 0.191 nats on development ($\Delta_{\text{prop}} = -0.191$, cluster-bootstrap interval $[-0.358, +0.033]$) and by 0.135 on the held-out window ($[-0.281, +0.067]$). An in-sample calculation suggests otherwise, giving a 0.032-nat improvement, but it does not survive a held-out evaluation with a cluster bootstrap on either window. Both conditions of step 2 therefore fail, and 0 (C) is not available on either window.

The development analysis uses Olist data to 2018-02; the pipeline was then corrected for the row-accounting and weight-convention deviations of Appendix D.6, after the first held-out results had been seen, so the later window is a second observation window and not a confirmatory test. We ran a single analysis on the final six months (after 2018-02-28 15:00:37, $n = 38{,}923$). What recurs there is the refusal, not a causal diagnosis: the coupling does not clear its detectability threshold, and the ladder gives no collapse.

The measured coupling rises between windows, from [0.30, 4.08] to $[1.46, 7.22] \times 10^{-3}$ at the interacted contrast, and whether that changes the branch depends on the reference distribution. Calibrated on the bootstrap that produced the interval, the later window's threshold at the interacted contrast is $[9.03, 17.1] \times 10^{-3}$, well above the coupling, so that cell is 0 (P). At the headline main-effect contrast the same calibration gives [2.66, 4.46] against a coupling of [0.46, 4.67]: the two overlap and the cell is 0 (P?). The development window is 0 (P) on both contrasts. What recurs across windows is the refusal, not its code: on neither window can 27 clusters resolve the coupling well enough to license a policy, but only on the development window can the audit say the coupling lies *below* what the design can detect. Nor is the held-out coupling robustly detectable on either footing: that branch needs the lower limit above $\eta^2_{\text{min,hi}}$, and 1.46 lies below both 4.51 and 17.1. The rise coincides with an operational shift, in which commitments tighten (median 24.2 → 21.3 days) and transit times fall (11.3 → 8.6 days), and with the ladder's propensity condition failing for the same reason, commitments becoming uncoupled from historical lane speed.

A Tier 0 refusal is not a dismissal of the dataset. The association is spread across lanes: under hierarchical Benjamini–Hochberg control, among 37 lanes with $\geq 300$ orders, 6 remain significant (vs 1.9 expected false positives), with raw and adjusted $p$-values reported in Table S5. Since

**Table S7**
Certification on the pre-specified trimmed support ($S$=30, $q$=0.5).

| Learner | seeds cert. | mean uplift | $\hat{\kappa}'$ vs $\lambda$ |
|---|---|---|---|
| CQL **(deployed)** | **29/30** | +0.023 | 53 > 15.7 (band non-empty) |
| greedy-DR reference | 8/30 | +0.029 | 13 < 15.7 (band empty) |
| GBM (unregularised) | 3/30 | +0.024 | < 1 (no cert.) |

*Notes.* $\lambda = \lambda_{0.05,0.20,9} = 15.70$ is the exact non-centrality at the interacted contrast $df_1 = 9$ and $\nu = 2{,}518$ (Table S8).

predictive value does not depend on action–outcome coupling, Table 4 advises calibrated quoting below Tier 3.

## Appendix D.6. Row accounting and deviations from the plan

Table S5 carries the row flow. Two points of row accounting are recorded here because they bear on how the Olist results should be read.

*The two cut-offs.* The written plan fixed one boundary between the development and the held-out window. A first pass of the pipeline used `2018-02-28 15:00:37` for the held-out window but `2018-03-01` for the development window, so 164 orders fell in both. The results reported here use the unified boundary, which gives 56,383 development and 38,923 held-out rows, summing to the 95,306 analysed rows, with no row counted twice. Because the first pass used rows that the plan had reserved, the development analysis is reported as exploratory.

*The weight convention.* The audit prints $w = \rho(a)/\pi_b(a \mid s)$ with uniform $\rho$, so $\rho(a) = 1/4$ at $K = 4$. On this convention $\max_i w_i$ is 24.5 on the trimmed support and 524 untrimmed; on the $1/\pi_b$ convention both would be four times larger. $n_{\text{eff}}$, $o_{\text{eff}}$ and every coupling use the same $\rho/\pi_b$ convention.

*Cluster size.* Equation (5) deflates $n_{\text{eff}}$, not the raw row count, and the cluster sizes are extremely uneven: the largest of the 27 customer states holds 22,490 development rows and the smallest 22. The arithmetic mean cluster size is 2,088 against a Kish mean of 11,433. This unevenness is one reason the mean-based Kish factor is the wrong deflator for the action block, as §3.4 and Appendix D.7 discuss.

## Appendix D.7. The design effect of the action block, measured

Section 3.4 defines the information count through the generalised design effect $\bar{\delta} = \text{tr}(V_0^{-1} V_{\text{CR}})/df_1$ of the action block and warns that the familiar Kish factor $1+(\bar{m}-1)\hat{\rho}_{\text{ICC}}$, being the design effect of a *mean*, overstates it for a regressor that varies within clusters. Table S8 reports the measured $\bar{\delta}$, with a leave-one-cluster-out jackknife band, for every specification and both windows.

The gap is large and it runs in the direction §3.4 predicts. At $df_1 = 9$ the Kish factor is 27.7 and the measured $\bar{\delta}$ is 6.75, so a threshold computed with the Kish factor would be inflated about fourfold. Two features of the measurement are worth stating. First, $\bar{\delta}$ *falls* as the contrast is enriched, from 6.75 at $df_1 = 9$ to 5.06 at $df_1 = 36$: the interaction coefficients are identified by within-cluster contrasts, which cluster dependence degrades less than it degrades a between-cluster mean. This is the Moulton mechanism [46] and it is why the audit reports $\bar{\delta}$ per contrast rather than one factor per log. Second, a smaller deflator raises $n_{\text{info}}$ and so *lowers* the threshold, which is the permissive direction for a refusal; using $\bar{\delta}$ rather than the Kish factor therefore works against the paper's own refusals.

## Appendix D.8. The Olist ladder and trim sweep

Table S9 gives the saturation ladder rung by rung for both windows and both covariate paths, and Table S10 the verdict over the admissible trim. Both are referenced from §6.4; they are placed here because the rung-level detail supports the verdict without carrying it.

## Appendix D.9. Theorem 2 against the finite procedure and real outcomes

The degeneracy constant is the exact $m_K$, not its asymptotic approximation $\sqrt{2 \log K}$, which overstates the spurious gain by a factor of 2.09 at $K = 2$, a binary shipping mode (Table S13a).

The reuse theorem is verified against the procedure itself. Theorem 2 describes a limit, and what an analyst runs is a finite procedure: fit cell means, take the greedy policy, evaluate it doubly robustly on the same rows. Table S11 puts the two side by side at $S = 10$ states and $n_s = 500$ rows per state. For the saturated learner, which is the learner the theorem assumes, the measured rate matches $\Pr(T_S > z)$ within three Monte-Carlo standard errors at four of five logging probabilities. The exception is $p = 0.05$, where the rare action receives 25 rows per state and the cell-mean normal approximation has not taken hold; we report it as a finite-sample departure rather than as evidence against the limit. Two features of the departures are worth stating. All four are positive, so the finite-sample rate sits *above* the limit, which is the conservative direction for a warning about reuse. And they grow as $p$ falls, consistent with the cell-mean approximation degrading as the rare action thins: the theorem is a limit in $n_s$, and these cells hold $n_s = 500$.

The number of selections, not overlap, is the operative parameter, and the measurement separates them cleanly. Holding $S = 10$ and sweeping the logging probability from uniform to $p = 0.05$ moves the saturated learner's rate from 0.818 to 0.468, a factor of 1.7, in the direction Theorem 2(iii) gives. That part bounds the drift $\mu_S$ within a factor of $\sqrt{2}$, not the rate itself, which is a nonlinear function of the drift:

**Table S8**
Measured design effect of the action block, against the mean-based Kish factor. Jackknife band is leave-one-cluster-out. Threshold band is Eq. (8) at $n_{\mathrm{eff}}/\bar{\delta}$ over that band. The last two columns calibrate the threshold on the wild cluster bootstrap that produces the interval, by Algorithm 2; this is the footing §4.3 requires, since interval and threshold then come from one test. The CR2 Satterthwaite degrees of freedom for the action block are 3.3 and 9.0 on Olist and 11.7 and 21.0 on DataCo, which diagnoses how little independent information the clusters supply but cannot itself set a threshold (§6.5).

| Corpus, window | $df_1$ | $\bar{\delta}$ | $n_{\mathrm{info}}$ | coupling | thr. (iid $\nu$) | thr. (Alg. 2) | branch |
|---|---|---|---|---|---|---|---|
| Olist, dev | $3^{\ddagger}$ | 4.15 | 4,133 | $[0.00, 0.96]$ | $[2.30, 3.10]$ | $[1.55, 2.33]$ | 0 (P) |
| Olist, dev | 9 | 6.75 | 2,539 | $[0.30, 4.08]$ | $[4.62, 6.37]$ | $[9.40, 16.9]$ | 0 (P) |
| Olist, dev | $36^{\S}$ | 5.06 | 3,387 | $[2.23, 11.93]$ | $[6.88, 7.88]$ | — | — |
| Olist, held-out | $3^{\ddagger}$ | 3.75 | 3,172 | $[0.46, 4.67]$ | $[2.66, 3.86]$ | $[2.66, 4.46]$ | 0 (P?) |
| Olist, held-out | 9 | 3.25 | 3,651 | $[1.46, 7.22]$ | $[3.79, 4.51]$ | $[9.03, 17.1]$ | 0 (P) |
| Olist, held-out | $36^{\S}$ | 3.12 | 3,807 | $[4.96, 15.99]$ | $[6.01, 7.16]$ | — | — |
| DataCo | $3^{\ddagger}$ | 3.35 | 27,124 | $[0.00, 0.29]$ | $[0.350, 0.420]$ | $[0.235, 0.354]$ | 0 (P?) |
| DataCo | 21 | 3.60 | 25,226 | $[0.07, 1.05]$ | $[0.779, 0.869]$ | $[1.09, 1.73]$ | 0 (P) |
| *Mean-based Kish factor, for comparison* | | | | | | | |
| Olist dev; held-out | | | | | Kish 27.7; 19.4 ($\hat{\rho}_{\mathrm{ICC}} = 0.0128$) | | |
| DataCo | | | | | $\hat{\rho}_{\mathrm{ICC}} = 0$, read as $\bar{\delta} = 1$ | | |

**Table S9**
Olist saturation ladder, recomputed on the audit's own specification: the registered trim $\pi_b \geq 0.01$, importance weights $w = \rho/\pi_b$, clusters the 27 customer states, and 199 paired cluster bootstrap resamples shared across rungs. Both scales are reported, the normalised $\hat{\omega}^2$ and the absolute $\hat{D}$ that the routing rule reads (§5.2, step 2).

| | development, $df_1$=3 | | development, $df_1$=9 | | held-out, $df_1$=9 | |
|---|---|---|---|---|---|---|
| Rung | $\hat{\omega}^2$ | $\hat{D}$ | $\hat{\omega}^2$ | $\hat{D}$ | $\hat{\omega}^2$ | $\hat{D}$ |
| L0 base | 0.00 | 0.00 | 0.80 | 1.33 | 3.06 | 4.67 |
| L1 +geo | 0.00 | 0.00 | 0.88 | 1.47 | 2.92 | 4.45 |
| L2 +cat/zip | 0.00 | 0.00 | 0.80 | 1.32 | 2.69 | 4.10 |
| L3 +seller | 0.30 | 0.49 | 0.93 | 1.53 | 3.35 | 5.08 |
| L4 +lane | 0.88 | 1.44 | 1.24 | 2.04 | 4.32 | 6.50 |
| $L4/L0$ on $\hat{D}$ | 7.11 $[0.29, 234]$ | | 1.17 $[0.73, 2.57]$ | | 1.29 $[0.88, 2.03]$ | |
| noise control | 1.04 | | 1.02 | | 0.97 | |
| **Branch** | **no collapse** in any cell ($\delta_{\mathrm{col}} = 0.5$); ladder is C? | | | | | |

over $p \in [0.02, 0.5]$ the exact limit moves the rate by up to 2.33× (at $S = 8$), so the envelope on the rate is looser than the envelope on the drift and we state the two separately. Holding $p = \frac{1}{2}$ and sweeping the number of state cells from 1 to 32 moves it from 0.057 to 1.000 (Figure 3). One parameter changes the rate by about twofold at most; the other takes it from the nominal level to near-certainty. This is the practical content of part (ii), and it is why the audit's fold discipline is stated per learner and not per overlap regime.

Where a learner is not the saturated one, Theorem 2 is read through the effective number of selections $S_{\mathrm{eff}}$, the value of $S$ at which $\Pr(T_S > z)$ equals the learner's measured rate (Remark 3). The reading is informative in both directions. A shrunk tabular learner with $\alpha = 50$ pseudo-observations, the single-step analogue of the conservative learner deployed in §6.4, has $S_{\text{eff}}$ between 9.1 and 10.5 at $S = 10$: its shrinkage changes the rate very little, because the selection is still made state by state. A pooled learner has $S_{\text{eff}} = 1.0$ at every overlap level and at every number of cells up to 32, sitting in $[0.047, 0.057]$ against $\alpha = 0.05$. That is Theorem 2(i) measured rather than assumed, and it is why the audit attaches its reuse warning to the absence of a split, not to near-random logging or to a particular learner. A disjoint-fold split returns every learner to $[0.043, 0.064]$, part (iv).

**Table S10**
Olist verdict swept over the admissible trim $\delta$, on the Rao–Scott footing of (5) at the interacted contrast $df_1 = 9$ (the headline is the main-effect contrast $df_1 = 3$; Table 6). With the Kish factor and the rank-deficient $df_1 = 36$ contrast, every row would read 0 (P).

| $\delta$ | $n_{\mathrm{eff}}$ | $\bar{\delta}$ | $n_{\mathrm{info}}$ | coupling $[\hat{L}, \hat{U}]$ | threshold band | verdict |
|---|---|---|---|---|---|---|
| 0.002 | 8,789 | 8.30 | 1,059 | $[2.08, 10.75]$ | $[12.01, 15.37]$ | 0 (P) |
| 0.005 | 11,939 | 6.73 | 1,775 | $[1.31, 6.19]$ | $[6.74, 9.08]$ | 0 (P) |
| $0.010^{\ddagger}$ | 17,133 | 6.75 | 2,539 | $[0.30, 4.08]$ | $[4.62, 6.37]$ | 0 (P) |
| 0.020 | 21,993 | 3.99 | 5,506 | $[0.36, 3.89]$ | $[2.22, 2.99]$ | 0 (P?) |
| 0.050 | 30,269 | 3.28 | 9,241 | $[0.29, 3.22]$ | $[1.38, 1.75]$ | 0 (P?) |

The conservative learner deployed on Olist is the case where this reading matters most, and it does not fall where one would guess. Run on decoupled data at that corpus's configuration ($K = 4$, eleven covariates, $n = 17{,}000$, state-dependent logging), the conservative learner's reuse rate is 0.067 at $\alpha_{\text{CQL}} = 1$ and 0.008 at $\alpha_{\text{CQL}} = 5$, giving $S_{\text{eff}} = 1.4$ and 1.0; a tabular learner over four cells on the same data sits at 0.367, or $S_{\text{eff}} = 6.2$. The conservative learner behaves like a pooled one under reuse, not like a state-wise one, and the reason is not that it collapses onto a single action (it uses all four) but that its value function is flat enough for the doubly robust correction to cancel almost the whole direct-method gain: +0.150 against −0.143 at $\alpha_{\text{CQL}} = 1$, and +0.172 against −0.171 at $\alpha_{\text{CQL}} = 5$. That is $\tau = 0.95$ in Definition 3: the correction removes almost all of the direct-method optimism, which is what Remark 3 describes for a learner between the two ends of Theorem 2. This is not the value in Table S13b, which is −0.022 [−0.38, 0.30]; the two are measured on different designs (that one on the simulation grid of this subsection, this one at the Olist configuration), and 0.95 lies outside that interval, so they are separate measurements and not a check on one another. The practical reading is the opposite of a reassurance: this learner's low reuse rate on decoupled data is a property of its flatness, not evidence that reuse is safe for it, and the 29/30 certificate on Olist is still run on disjoint folds because part (iv) is what guarantees validity for every learner, whatever its $S_{\text{eff}}$. For $o_{\text{eff}} < o^{\dagger} \approx 0.05$ the split alone exceeds $\alpha$ (up to 0.15 at $o_{\text{eff}} \approx 0.01$); below that floor the interval fails, not the ordering (Remark 14).

The optimism constant is reported for completeness in Appendix B.7, with Table S13b. It belongs to Remark 3, not to Theorem 2, which contains no $\tau$: the sign of $\tau$ says whether the doubly robust correction removes or adds optimism for a flexible learner, and its measured negativity for gradient boosting is the reason the audit treats a doubly robust estimator as no mitigation for fold reuse.

**The limit against real outcomes.** Theorem 2 assumes equal $n_s$ and $p$ across states and approximately Gaussian cell means. Field outcomes satisfy none of that, so we run the shortcut on real outcomes with the coupling removed by construction: actions are permuted within (stratum × cluster), which leaves the outcome distribution, the action shares and the cell imbalance exactly as the log has them (Table S12). Both parts of the theorem hold. Splitting keeps the rate at 0.030–0.065 on both corpora and at every stratification, which is part (iv) measured on real data. Reuse rises with the number of strata as part (ii) requires, and it tracks the limit generalised to the empirical $n_s$, $p_s$ and $\sigma_s$: the ratio of measured rate to generalised limit runs from 0.84 to 1.18 across the seven cells, with no systematic direction. The limit therefore describes these field outcomes (a margin distribution with skew −5.0 and excess kurtosis 80, and a click rate of 0.35%), not only the Gaussian equal-cell design it was derived under. We report the generalisation as an empirical finding, since substituting empirical $n_s$, $p_s$ and $\sigma_s$ into (11) is not part of the proved result.

**The learnability constant.** $\hat{\kappa}'$ is calibrated at $n_{\text{eff}}^{\text{cal}} \approx 2{,}640$ and is stable across $n = 4{,}000$ to 8,000. Applying it to Olist would require extrapolating by a factor of 6.1 in $n_{\text{eff}}$ (0.8 in $n_{\text{eff}}^{\text{clu}}$), beyond the twofold range swept, which is why the Olist verdict is set by direct certification.

## Appendix D.10. Few-cluster tests and the calibrated threshold

The interval and the threshold must come from the same test, and at 23–27 clusters that test has to be chosen rather than assumed. We compare four on designs matched to each corpus, under the null (Table S14). A cluster-robust Wald statistic referred to $\chi^2_{df_1}$ rejects far too often, and increasingly so as the contrast is enriched. The Hotelling reference $F(df_1, G - df_1)$ is adequate on the more even DataCo-like design and not on the Olist-like one, whose largest cluster holds 67 times the rows of its smallest. The CR2 correction of [57] with Satterthwaite degrees of freedom holds its size exactly at $df_1 = 3$ (0.050, with power 0.985 against a main effect of $\beta = 0.05$), but at $df_1 \geq 9$ on this design it rejects essentially never, so its threshold would not be interpretable. The constrained wild cluster bootstrap with Webb weights is the only test whose size stays within $[0.056, 0.068]$ across every contrast we can run, and it is the one the field intervals use.

Its degrees of freedom are the binding constraint, and the CR2 fit makes that visible. On Olist's real design the Satterthwaite degrees of freedom for the action block are 3.3 at $df_1 = 3$ and 9.0 at $df_1 = 9$, at or barely above $df_1$ itself, against the 2,500 that treating the information count as independent rows would imply. That diagnosis is useful, but it does not yield a usable threshold. CR2's own reference distribution is $F(df_1, \eta - df_1 + 1)$, which at $\eta = 3.3$ and $df_1 = 3$ leaves 1.3 denominator degrees of freedom and a noncentrality of 336, so the implied threshold would be two orders of magnitude above any coupling these data could carry. We therefore do not read a threshold off (8) at those degrees of freedom.

The threshold is instead calibrated on the test that produces the interval, as §4.3 and Appendix A.5(vi) require. Algorithm 2 is run with the wild cluster bootstrap itself: residuals are resampled around the fitted $\mathcal{M}_0$ with Rademacher cluster weights, a rescaled copy of the fitted action block is injected at each candidate coupling, and $\eta^2_{\text{min,cal}}$ is the

**Table S11**
Theorem 2 against the finite procedure, on decoupled data (true improvement = 0), $K = 2$, $n_s = 500$ rows per state, 1,500 replications per cell (Monte-Carlo SE $\sqrt{r(1-r)/1500}$ evaluated at the limiting rate $r$, between 0.010 and 0.013 across these cells). Upper panel fixes $S = 10$ and sweeps the logging probability; lower panel fixes $p = \frac{1}{2}$ and sweeps the number of state cells. $S_{\mathrm{eff}}$ is the number of selections at which the limit reproduces the measured rate.

| | | saturated (the theorem's learner) | | | shrunk, $\alpha = 50$ | | pooled | | split |
|---|---|---|---|---|---|---|---|---|---|
| $p$ | $o_{\mathrm{eff}}$ | $\Pr(T_S > z)$ | measured | $\Delta$/SE | measured | $S_{\mathrm{eff}}$ | measured | $S_{\mathrm{eff}}$ | measured |
| *Sweeping overlap at $S = 10$ cells* | | | | | | | | | |
| 0.50 | 1.000 | 0.822 | 0.818 | −0.4 | 0.814 | 9.9 | 0.050 | 1.0 | 0.043 |
| 0.35 | 0.910 | 0.771 | 0.799 | +2.6 | 0.802 | 10.5 | 0.049 | 1.0 | 0.064 |
| 0.20 | 0.640 | 0.611 | 0.620 | +0.7 | 0.633 | 10.3 | 0.050 | 1.0 | 0.053 |
| 0.10 | 0.360 | 0.478 | 0.501 | +1.8 | 0.490 | 10.2 | 0.050 | 1.0 | 0.050 |
| 0.05 | 0.190 | 0.419 | 0.468 | +3.8† | 0.367 | 9.1 | 0.048 | 1.0 | 0.046 |
| *Sweeping the number of state cells at $p = \frac{1}{2}$* | | | | | | | | | |
| $S$=1 | — | 0.050 | 0.057 | +0.5 | 0.057 | 1.1 | 0.057 | 1.1 | — |
| $S$=2 | — | 0.098 | 0.091 | −0.5 | 0.091 | 1.9 | 0.047 | 1.0 | — |
| $S$=4 | — | 0.257 | 0.247 | −0.8 | 0.247 | 3.9 | 0.048 | 1.0 | — |
| $S$=8 | — | 0.670 | 0.668 | −0.1 | 0.669 | 8.0 | 0.049 | 1.0 | — |
| $S$=16 | — | 0.986 | 0.981 | −0.4 | 0.981 | 15.4 | 0.049 | 1.0 | — |
| $S$=32 | — | 1.000 | 1.000 | 0.0 | 1.000 | > 200 | 0.049 | 1.0 | — |

*Notes.* † at $p = 0.05$ the rare action has 25 rows per state, below the $n_s \min(p, 1-p) \geq 100$ regime in which the cell-mean normal approximation holds.

**Table S12**
The reuse shortcut on real outcomes with coupling removed by permutation within (stratum × cluster). "Generalised limit" is $\Pr(T_S > z)$ evaluated at the empirical $n_s$, $p_s$ and $\sigma_s$ rather than at the theorem's equal-cell values; that generalisation is not part of the proved result.

| Corpus (reward) | strata | $\bar{n}_s$ | reuse | split | gen. limit |
|---|---|---|---|---|---|
| OBD (click), 1,374,327 rows | 1 | 1,374,327 | 0.045 | 0.030 | 0.049 |
| 4,768 clicks, CTR 0.35% | 2 | 687,164 | 0.100 | 0.045 | 0.097 |
| | 4 | 343,582 | 0.185 | 0.030 | 0.221 |
| DataCo (margin), 142,968 rows | 1 | 142,968 | 0.058 | 0.052 | 0.049 |
| skew −5.0, kurtosis 80 | 3 | 47,656 | 0.150 | 0.065 | 0.128 |
| | 9 | 15,885 | 0.445 | 0.062 | 0.445 |
| | 14 | 10,212 | 0.690 | 0.037 | 0.678 |

smallest coupling the bootstrap test rejects with probability 0.8 (60 replications per grid point). Table S8 sets the result beside the independent-rows convention. The two footings agree on three of six cells and differ only in the reason code of the other three, never in the tier, and Olist's headline 0 (P) at $df_1 = 3$ holds under both, with the calibrated threshold $[1.55, 2.33] \times 10^{-3}$ against a coupling of $[0.00, 0.96] \times 10^{-3}$. Where they differ the calibrated threshold governs, because it is the only one whose reference distribution matches the interval's.

**What twenty-five clusters can and cannot test.** Size alone overstates what the wild cluster bootstrap delivers, so the same grid reports its power at each design's own detectability threshold rather than at an arbitrary effect. The bootstrap resolves a main-effect contrast with probability 0.55–0.65 at 1× and 0.82–0.93 at 2×, but only 0.19–0.21 and 0.38–0.43 once $df_1$ reaches 18 (Table S14). A log with a few tens of clusters can therefore license a fixed rule and not a contextual one, however many rows it has: the binding resource is clusters, and no amount of row-level data substitutes for it. This is why §5.2 routes the two policy classes separately and why the headline verdicts are read at the main-effect contrast.

## Appendix D.11. What the saturation ladder can and cannot separate

The ladder's collapse branch is the audit's only causal diagnostic, so the rule itself needs testing on data where the answer is known. Table S15 runs four scenarios at $n =$ 20,000 independent rows, each with and without a genuine action effect, and compares a rule based on the point ratio with the tested rule of step 2.

**The scale of the ratio.** In the confounding scenario an unobserved $U$ drives both the action and the outcome, and a proxy $z$ for $U$ enters at rung L3. Adjusting for $z$ should remove the spurious coupling, and on the absolute scale it does: $\hat{\mathcal{D}}$ falls to 0.06 of its base value. On the normalised scale the same rungs give 0.21, and in the variant with a true effect the normalised ladder *rises* to 2.16 while the absolute one falls to 0.58. The reason is mechanical. Writing $\hat{\omega}^2$ as $\mathrm{SS}_\Delta/(\mathrm{SS}_T+\mathrm{MSE})$, a covariate that both confounds the action and explains the outcome shrinks $\mathrm{SS}_\Delta$, as intended, and $\mathrm{SS}_T$ at the same time. In our confounding design the numerator falls by 57% and the denominator by 72%, so the ratio climbs even though the coupling it is meant to track has more than halved. The normalised scale is right for comparing a coupling with a detectability threshold, which is also normalised; it is wrong for a ladder, whose rungs have different denominators by construction.

**Table S13**
Simulator constants. Panel (b) describes learners between the two ends of Theorem 2 (Remark 3); the theorem itself contains no $\tau$. Routing operating characteristics are in Table 7.

| (a) Degeneracy constant | | | |
|---|---|---|---|
| $K$ | $m_K$ exact | $\sqrt{2\log K}$ | ratio |
| 2 | 0.5642 | 1.1774 | 2.09 |
| 4 | 1.0294 | 1.6651 | 1.62 |
| 8 | 1.4236 | 2.0393 | 1.43 |
| (b) Optimism constant (Appendix B.7) | | | |
| Learner | eff. d.f. | $\hat{\tau}$ [95% CI] | correction |
| Linear | 6 | −0.038 [−0.10, 0.02] | neither |
| Tree, depth 4 | 16 | 0.544 [0.52, 0.57] | removes |
| GBM, 200 trees | 200 | −0.282 [−0.29, −0.27] | **adds** |
| Deliberately overfit | 1000 | 0.000 (exact) | vanishes |
| CQL (deployed) | — | −0.022 [−0.38, 0.30] | neither |

**Table S14**
Size and power of cluster-robust tests on designs matched to the two logistics corpora, at $\alpha = 0.05$ with 1,000 replications for size (Monte-Carlo SE 0.007) and 250 for power. Size is measured under $\mathcal{D} = 0$. Power is measured at the design's own detectability threshold and at twice it, rather than at an arbitrary effect, so the columns say what the test can resolve relative to what it claims to resolve.

| | | | size under $\mathcal{D} = 0$ | | | power, wild boot. | |
|---|---|---|---|---|---|---|---|
| Design | $df_1$ | $\eta^2_{\min}$ | Wald $\chi^2$ | Hotelling $F$ | wild boot. | at 1× | at 2× |
| Olist-like | 3‡ | 0.55 | 0.148 | 0.093 | **0.068** | 0.548 | 0.824 |
| $G = 27$, sizes | 9 | 0.78 | 0.505 | 0.176 | **0.068** | 0.440 | 0.692 |
| uneven 67× | 18 | 1.01 | 0.983 | 0.286 | **0.064** | 0.212 | 0.432 |
| DataCo-like | 3‡ | 0.55 | 0.100 | 0.050 | **0.056** | 0.648 | 0.928 |
| $G = 23$, sizes | 9 | 0.78 | 0.391 | 0.056 | **0.063** | 0.544 | 0.796 |
| uneven 4× | 18 | 1.01 | 0.963 | 0.059 | **0.065** | 0.188 | 0.384 |

**What the rule gets right, and what it costs.** The cell that tests detection is confounding with no true effect, where every unit of measured coupling is spurious: the tested rule collapses in 94% of replications. The cell that measures its cost is confounding with a true effect, where 58% of the coupling is genuine ($L4/L0 = 0.58$ on the absolute scale, above $\delta_{\mathrm{col}}$) and collapse would therefore be wrong: the rule still fires in 40% of replications. That is the rule's main failure mode and we report it as such, rather than as the detection rate it superficially resembles. Against it, the rule never fires in any of the six cells where collapse would be an error: variance absorption, a post-decision collider, and effect modification on a randomised log, each with and without a true effect.

**The propensity condition has to be a test.** Those six zeros depend on it. With $\Delta_{\mathrm{prop}} > 0$ implemented as a point estimate, the rule collapses in 55% of variance-absorption replications and 26% of randomised ones, because a point estimate of a quantity whose true value is zero exceeds zero about half the time. A one-sided test (the lower 95% limit of the held-out log-loss improvement above zero, which is what §5.2 means by "significantly improve") brings both to zero. The randomised cell matters particularly: Appendix C.2 states that collapse cannot fire on a randomised log because $\Delta_{\mathrm{prop}} = 0$ by construction, and only the tested version of the condition actually delivers that.

One limit of this evidence remains. The scenarios use 20,000 independent rows, so they test the rule's logic and not its behaviour at 23–27 clusters, where the intervals are wider and the power lower still. The 40% false-kill rate in the mixed cell is therefore likely to understate what a few-cluster log would show.

**Table S15**
The saturation ladder on four known-truth scenarios ($n$ = 20,000 independent rows, $R$ = 400 replications, $\delta_{col}$ = 0.5). "rel" is the ratio on the normalised $\hat{\omega}^2$ and "abs" on the absolute $\hat{\mathcal{D}}$. "point" collapses when the point ratio falls below $\delta_{col}$; "tested" requires a noise-adjusted decline *and* a one-sided test that the enrichment covariates improve action prediction ($\Delta_{prop}$ significantly positive), and is read on either scale.

| | | $L4/L0$ | | | | collapse rate | | |
|---|---|---|---|---|---|---|---|---|
| Scenario | effect | rel | abs | noise | $\Delta_{prop}$ | point | tested, rel | tested, abs |
| confounding | none | 0.20 | 0.05 | 1.00 | +0.257 | 0.73 | 0.72 | **0.94** |
| | yes | 2.13 | 0.58 | 1.00 | +0.256 | 0.08 | 0.07 | 0.40 |
| variance absorption | none | 0.34 | 0.16 | 1.00 | −0.000 | 0.55 | 0.00 | **0.00** |
| | yes | 2.07 | 1.00 | 1.00 | −0.000 | 0.00 | 0.00 | **0.00** |
| collider | none | 4125 | 1787 | 1.00 | +0.143 | 0.00 | 0.00 | **0.00** |
| | yes | 94 | 41 | 1.00 | +0.146 | 0.00 | 0.00 | **0.00** |
| randomised, effect modification | none | 0.92 | 0.84 | 1.00 | −0.000 | 0.26 | 0.00 | **0.00** |
| | yes | 1.09 | 1.00 | 1.00 | −0.000 | 0.00 | 0.00 | **0.00** |

## Appendix D.12. Plasmode on the two logistics designs

Both field logistics logs are refused, so the instrument's licensing behaviour is never exercised on a logistics design. A plasmode study supplies that. We keep each corpus's real weights, clusters, overlap, contrast and row count, and replace only the outcome, as $y^* = \hat{f}_0(s) + \theta\, h(s,a) + r \cdot v_g$, where $\hat{f}_0$ is the fitted $\mathcal{M}_0$ prediction, $r$ are the real residuals, $v_g$ are cluster-level Webb weights that preserve the log's own heteroscedasticity and within-cluster dependence, and $h$ is a known action block scaled so the implanted coupling is a stated multiple of that design's own calibrated threshold. Thresholds are the ones of Table S8, so the study runs on the footing the field audits use rather than on a more permissive one (Table S16).

Three things follow. First, the instrument licenses when it should: at four times the calibrated threshold, $\Pi_{const}$ reaches Tier 3 in 39 of 40 replications on the Olist design and 37 of 40 on the DataCo design. The refusals of §6.3 and §6.4 are therefore not an artefact of designs that could never yield a licence. Second, it does not license when it should not: across the 80 replications with no implanted coupling no verdict above Tier 1 was issued on either design, bounding the false-licence rate at 3.7% by the rule of three, and the same holds for the 160 replications in which the only association between action and outcome runs through a confounder. Third, the grid is informative about where the audit abstains rather than decides: at the threshold itself, 90% of cells return 0 (P?) and almost none return Tier 3, which is the intended behaviour of a rule that abstains when the data cannot separate a coupling from what the design detects.

The confounding rows show both the value of the ladder and its limit. With a proxy for the confounder in the ladder, the Olist design returns 0 (C) in 30 of 40 replications: the coupling is spurious and the audit says so rather than refusing for power. The DataCo design returns none, and that is correct rather than a failure. Step 2 requires the enrichment covariates to predict the action before a decline counts as confounding, and geography predicts the promise on Olist (C5b = +0.27 nats), whereas on DataCo no recorded covariate does: the full propensity model, geography included, scores −0.009 nats against marginal action rates. With the condition unmet the log is refused on detectability instead. When the proxy is absent from every rung, neither design ever licenses the confounded log, but neither can label it confounded. This is the blind spot that Appendix A.7 quantifies with a robustness value, and the reason a negative verdict is reported with one.

## Appendix D.13. Baselines and self-checks

An artefact that issues verdicts has to be measured against the alternatives it replaces, so the full rule is run end-to-end on a grid of 1,800 simulated logs per policy class (five coupling levels, three action-block structures, three designs), alongside three baselines on the same data (Table 7). B1 is the common shortcut: check overlap, then certify an off-policy estimate on reused folds. B2 splits the folds and certifies against the logging policy. B2′ splits the folds and certifies the increment over the best fixed rule; like B2, it runs no coupling test and no ladder. B3 is the audit of §5.1.

For non-contextual rules all four are defensible and the ordering is as expected: false-licence rates of 1.7%, 0.3%, 0.3% and 0.0%. B2 and B2′ coincide there, because for a fixed rule the increment is the policy itself. For contextual rules they separate sharply, and the separation is informative about where the protection comes from. B1 licenses 32.8% of decoupled cells. B2, which splits the folds, still licenses 18.7%, because splitting fixes the inference but not the question: it certifies a contextual policy against the logging policy, so a log whose value is entirely non-contextual passes. B2′, which splits and asks what personalisation adds over the best fixed rule, licenses 0.1%. B3 licenses none of 991 verdicts.

The decomposition is less favourable to the audit than a two-way comparison would be. The whole 18.6-point gap between B2 and B2′ comes from changing the question, a definitional change that any analyst could adopt without running this audit. The coupling test and the ladder that B3 adds remove the remaining 0.1 point. What they change is how the remaining logs are answered. B2′ has a false-kill rate of 22.7% and B3 one of 6.1%, but both rates are conditional on a verdict and B3 issues far fewer. Counted over the whole grid, B3 rejects 60 usable logs outright against 408 for B2′, but it abstains on 809 of 1,800 logs and licenses 32 of the 480 true Tier-3 cells, against 72 for B2′. The reduction in outright rejections is absorbed by abstention, not converted into licences. Its contribution is a third answer: an indeterminate verdict that comes with the pilot that would settle it, where B2′ can only license or reject.

**Table S16**
Plasmode on the two logistics designs: real weights, clusters, overlap and contrasts, synthetic outcomes with known coupling. Thresholds are the calibrated ones of Table S8, so the study is run on the footing the field audits use. Entries count verdicts out of 40 at the fixed-rule contrast.

| | Olist design ($G = 27$, $\eta^2_{\text{min,cal}} = 1.81 \times 10^{-3}$) | | | | | DataCo design ($G = 23$, $0.27 \times 10^{-3}$) | | | | |
|---|---|---|---|---|---|---|---|---|---|---|
| Scenario | P | P? | C | 1(L) | T3 | P | P? | C | 1(L) | T3 |
| no coupling | 40 | 0 | **0** | 0 | **0** | 38 | 2 | **0** | 0 | **0** |
| 1× | 4 | 36 | 0 | 0 | 0 | 2 | 36 | 0 | 1 | 1 |
| 2× | 0 | 25 | 0 | 3 | 12 | 0 | 26 | 0 | 1 | 13 |
| 4× | 0 | 1 | 0 | 0 | **39** | 0 | 3 | 0 | 0 | **37** |
| confounding, proxy in ladder | 4 | 6 | **30** | 0 | 0 | 22 | 18 | 0 | 0 | 0 |
| confounding, proxy absent | 12 | 28 | 0 | 0 | **0** | 22 | 18 | 0 | 0 | **0** |

The price is recall, and we state it on the whole grid rather than on the cells the rule chose to answer. On contextual cells B3 abstains on 45% of the grid and identifies 32 of the 480 true Tier-3 cells, against B1's 448. That comparison flatters B1 only until one notes how it is achieved: B1 licenses 57% of all cells in a grid where about a quarter deserve it. A rule that says yes to most logs will catch most good ones. The audit is built for the opposite error, since a false licence funds a programme that cannot work, and at these cluster counts it obtains that protection through abstention rather than through accuracy it does not have.

Table S17 collects the quantities a reader needs in order to trust every other number here, as measurements and not assurances. The first is the simulated bias of $\hat{\omega}^2$ under each corpus's *own* weight distribution and cluster structure, at the real $df_1$ and $\nu$, injected at three known coupling levels. The $\omega^2$ correction is derived for balanced unweighted analysis of variance and its transfer to importance-weighted clustered moments is not established, so its residual bias is quantified instead of assumed. Across both corpora it stays within 0.10 of the printed interval half-width, largest for Olist, whose heavier weights make the correction undercorrect. No verdict here turns on a bias comparable to its own uncertainty. The remaining rows are the tightness ratio $\hat{c}_{\text{emp}}(K)$ (Remark 8), the permutation null's seed proportion, and the robustness value of each refusal. Both robustness values are zero, which Appendix A.7 reads as underpowered at break-even: the coupling's upper limit already exceeds break-even, so no confounder is needed to reconcile the data with a coupling worth building. That is the applicable statement for a 0 (P) or 0 (P?) refusal, which turns on power. For 0 (C) or C? the question is instead what share of a ladder's movement could be mechanical, and the pure-noise control answers it: on Olist the control stays within 4% of one on every window (Table S9), and the ladder does not decline.

**Table S17**
The instrument's self-checks, reported as measurements and not assurances.

| Self-check | DataCo (margin) | Olist (review) |
|---|---|---|
| *$\hat{\omega}^2$ bias ÷ interval half-width, at true coupling* | | |
| $\eta^2_{\text{true}} = 0$ | 0.01 | 0.10 |
| $\eta^2_{\text{true}} = \eta^2_{\text{min}}$ | 0.02 | 0.06 |
| $\eta^2_{\text{true}}$ = measured | 0.07 | 0.04 |
| robustness value RV | 0 (underpowered) | 0 (underpowered) |
| $\hat{c}_{\text{emp}}(K)$, ceiling tightness | 0.30 on OBD at $K = 80$ (Appendix D.2); simulator elsewhere | |
| permutation-null seed prop. (95th pct) | — | 0.25 |